\documentclass[10pt,twocolumn,letterpaper]{article}

\usepackage{cvpr}              

\usepackage{graphicx}
\usepackage{amsmath}
\usepackage{amssymb}
\usepackage{booktabs}

\usepackage{algorithm}
\usepackage{algorithmic}

\usepackage{amsmath,amssymb}

\usepackage{color}
\usepackage{rotating}

\usepackage{dsfont}	
\usepackage{wrapfig}
\usepackage{footnote}
\usepackage{enumitem}

\usepackage{dirtytalk}
\usepackage{multirow}
\usepackage{boldline}
\usepackage{adjustbox}
\usepackage{balance}
\usepackage{caption}
\usepackage{subcaption}
 
\newcommand{\bfsection}[1]{\vspace*{0.1cm}\noindent\textbf{#1.}}
\usepackage[pagebackref,breaklinks,colorlinks]{hyperref}

\usepackage[capitalize]{cleveref}
\crefname{section}{Sec.}{Secs.}
\Crefname{section}{Section}{Sections}
\Crefname{table}{Table}{Tables}
\crefname{table}{Tab.}{Tabs.}

\def\cvprPaperID{8099} 
\def\confName{CVPR}
\def\confYear{2023}

\begin{document}

\title{Learning Dynamic Style Kernels for Artistic Style Transfer}

\author{Wenju Xu\\
OPPO US Research Center\\
Palo Alto, CA, USA\\
{\tt\small xuwenju123@gmail.com}
\and
Chengjiang Long\\
Meta Reality Lab\\
Burlingame, CA, USA\\
{\tt\small clong1@meta.com}
\and
Yongwei Nie\\
South China University of Technology\\
Guangzhou, Guangdong, China\\
{\tt\small nieyongwei@scut.edu.cn}
}

\twocolumn[{%
\renewcommand\twocolumn[1][]{#1}%
\maketitle
\captionsetup[subfigure]{justification=raggedright, singlelinecheck=false, labelformat=empty}  
\begin{center}
    \centering  
\vspace{-0.6cm}

    \includegraphics[height=0.088\textwidth,width=0.122\textwidth]{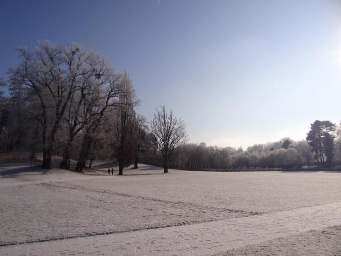}
      \hspace{-4pt} 
    \includegraphics[height=0.088\textwidth,width=0.122\textwidth]{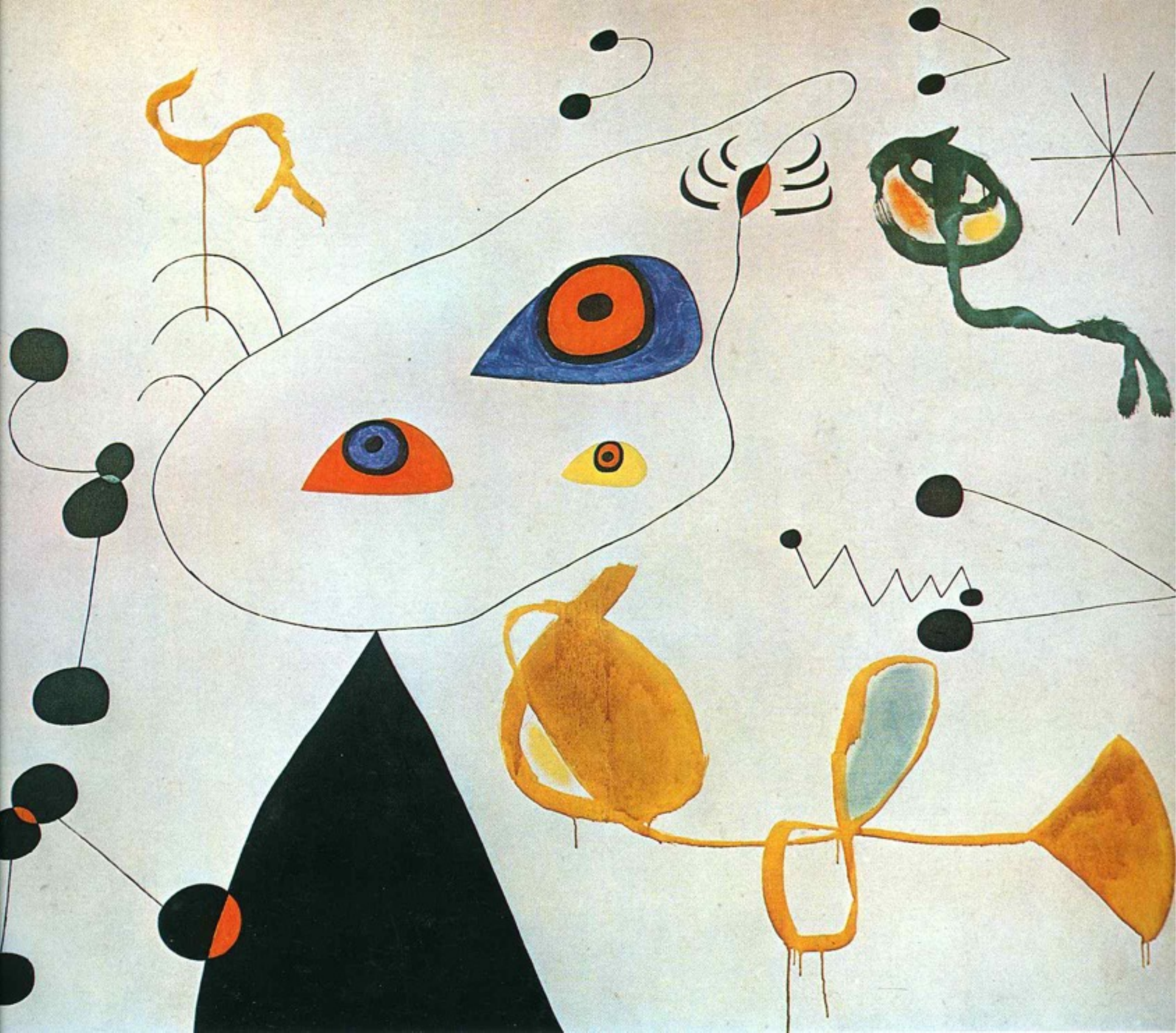}
      \hspace{-4pt}       
    \includegraphics[height=0.088\textwidth,width=0.122\textwidth]{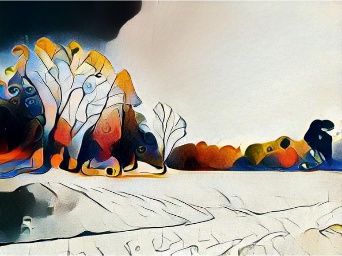}    
      \hspace{-4pt} 
    \includegraphics[height=0.088\textwidth,width=0.122\textwidth]{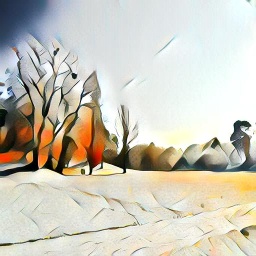}
      \hspace{-4pt}       
    \includegraphics[height=0.088\textwidth,width=0.122\textwidth]{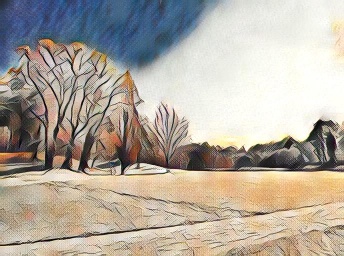}
      \hspace{-4pt}    
    \includegraphics[height=0.088\textwidth,width=0.122\textwidth]{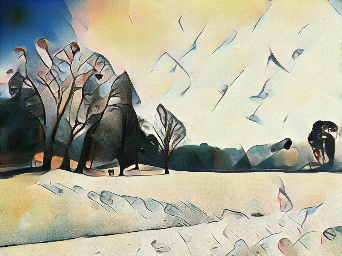}    
       \hspace{-4pt}  
     \includegraphics[height=0.088\textwidth,width=0.122\textwidth]{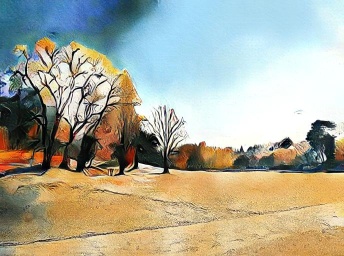}
       \hspace{-4pt}  
     \includegraphics[height=0.088\textwidth,width=0.122\textwidth]{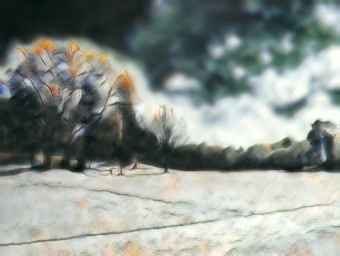}\\
    \includegraphics[height=0.088\textwidth,width=0.122\textwidth]{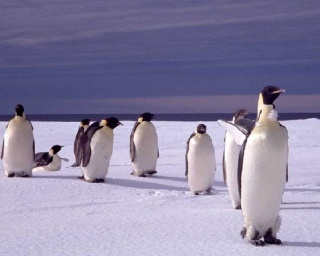}
      \hspace{-4pt} 
    \includegraphics[height=0.088\textwidth,width=0.122\textwidth]{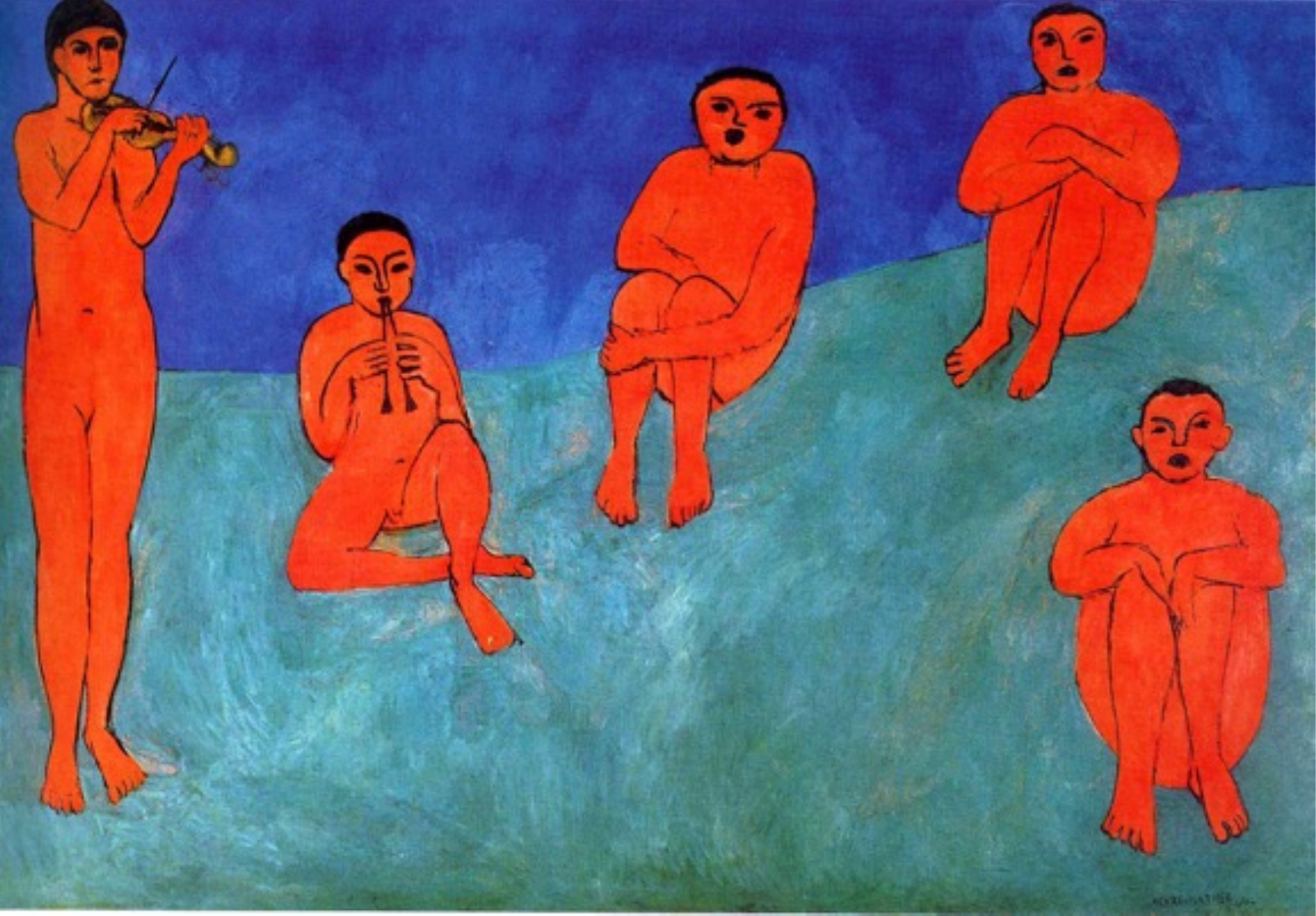}
      \hspace{-4pt}       
    \includegraphics[height=0.088\textwidth,width=0.122\textwidth]{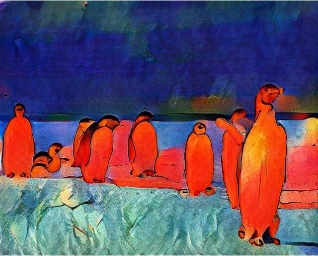}  
      \hspace{-4pt} 
    \includegraphics[height=0.088\textwidth,width=0.122\textwidth]{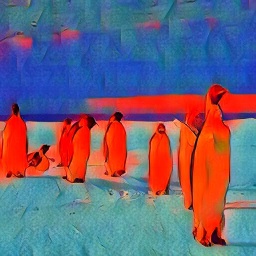}
      \hspace{-4pt}          
    \includegraphics[height=0.088\textwidth,width=0.122\textwidth]{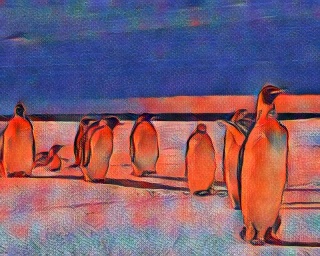}
      \hspace{-4pt}    
    \includegraphics[height=0.088\textwidth,width=0.122\textwidth]{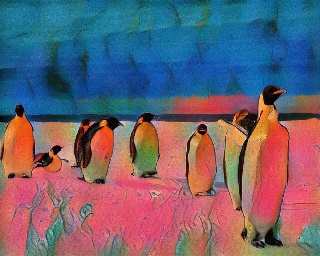}    
       \hspace{-4pt}  
     \includegraphics[height=0.088\textwidth,width=0.122\textwidth]{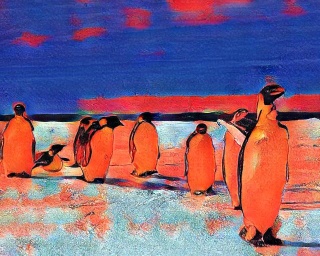}
       \hspace{-4pt}  
     \includegraphics[height=0.088\textwidth,width=0.122\textwidth]{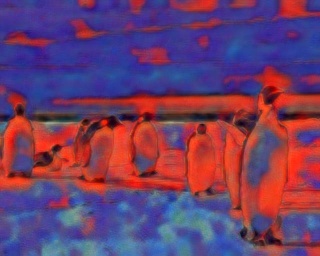}
\raisebox{0.005\textwidth }{\makebox[0.03\textwidth]{\parbox{0.92\linewidth}{\scriptsize \hspace{ 0pt} Content \hspace{ 40pt} Style  \hspace{ 44pt} Ours \hspace{ 36pt} StrTr$^2$'22 \hspace{ 32pt} AdaAttN'21 \hspace{ 18pt} DRG-GAN'21 \hspace{ 26pt} IEC'21 \hspace{ 32pt} MAST'21 }}}
 
    \includegraphics[height=0.12\textwidth,width=0.182\textwidth]{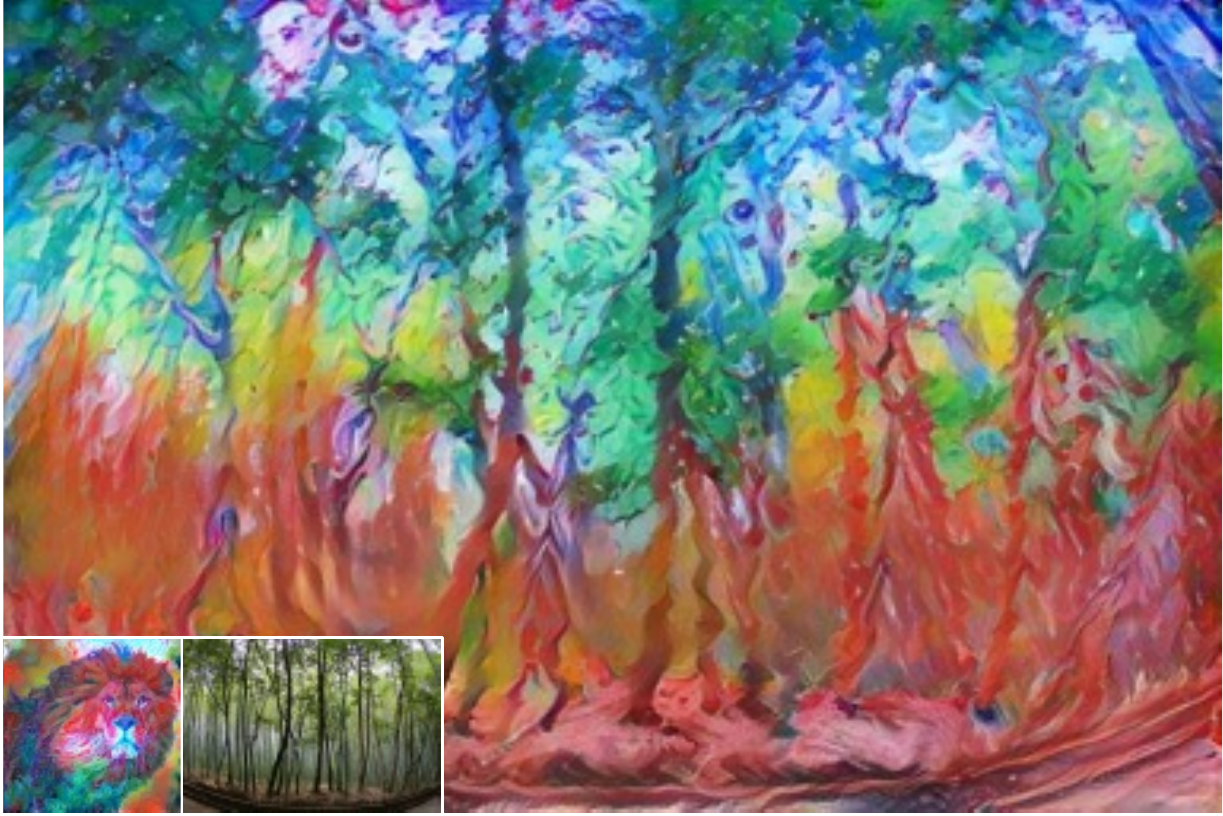}
      \hspace{-4pt}  
    \includegraphics[height=0.12\textwidth,width=0.162\textwidth]{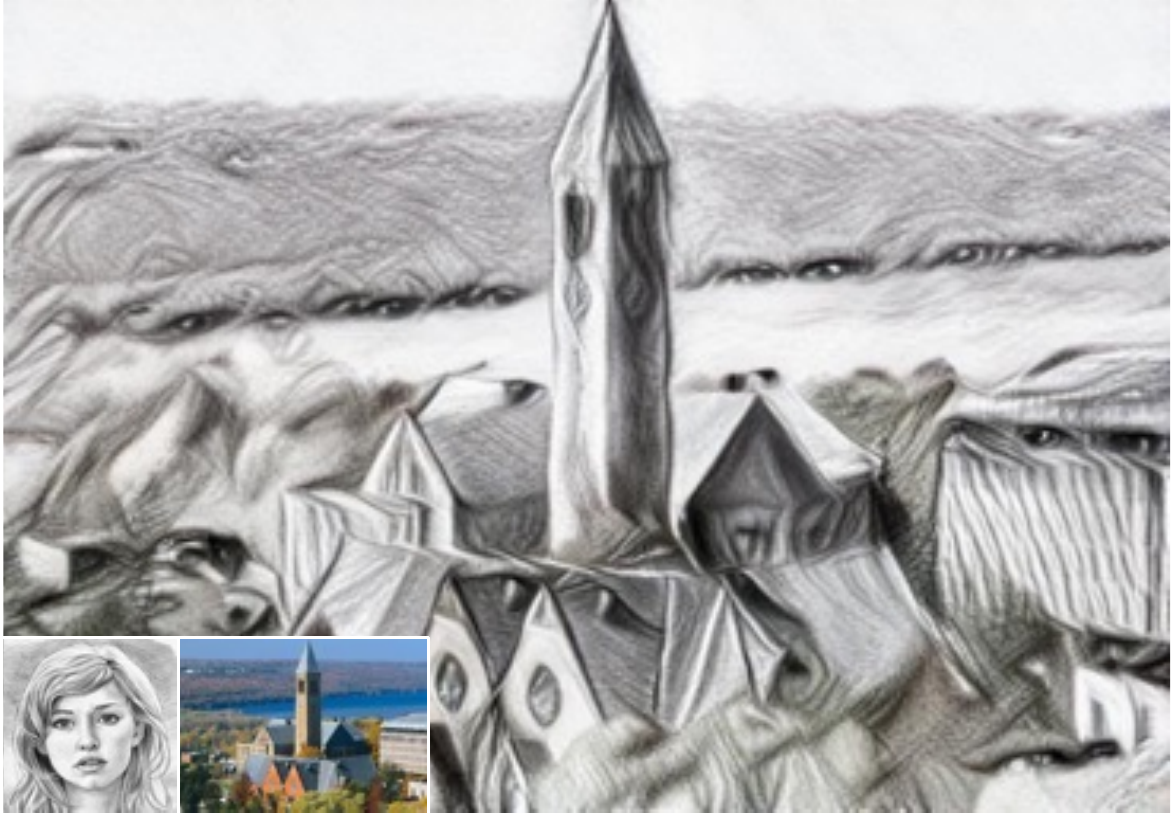}
      \hspace{-4pt}       
    \includegraphics[height=0.12\textwidth,width=0.182\textwidth]{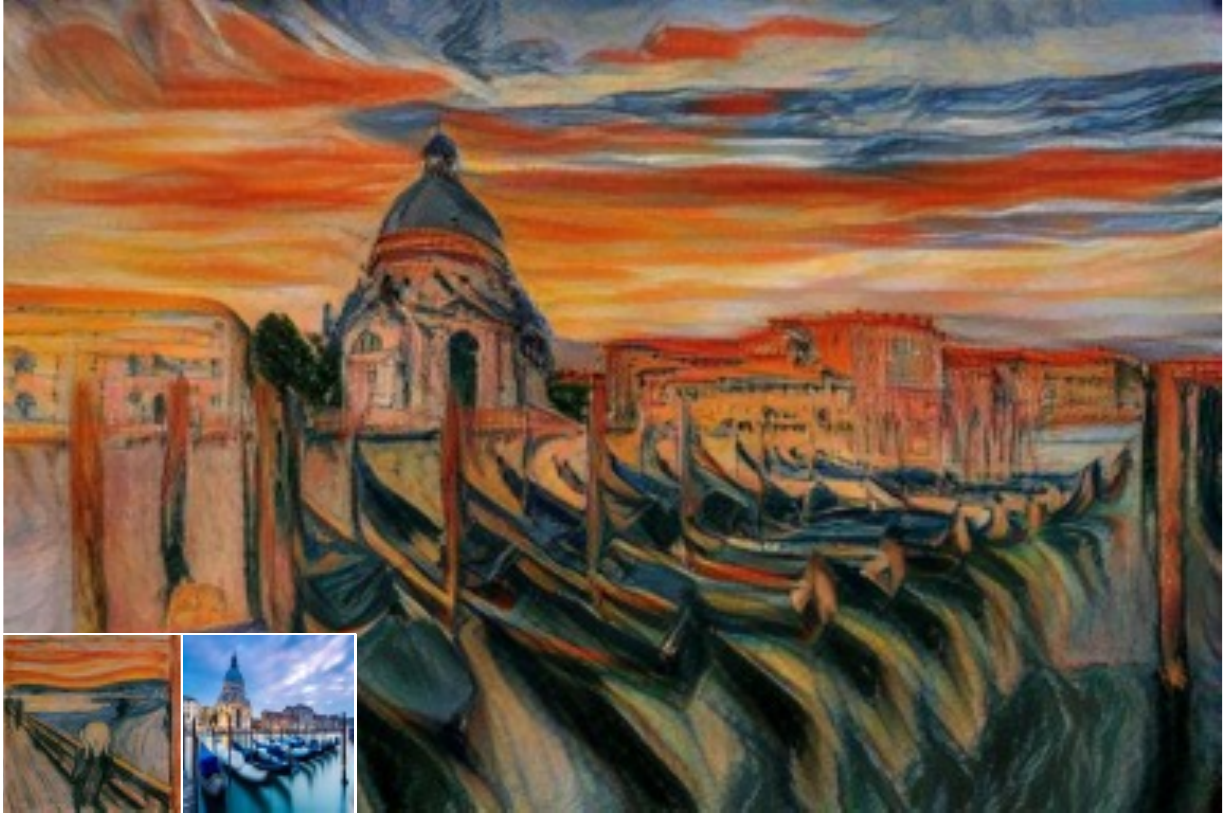}
      \hspace{-4pt}      
    \includegraphics[height=0.12\textwidth,width=0.152\textwidth]{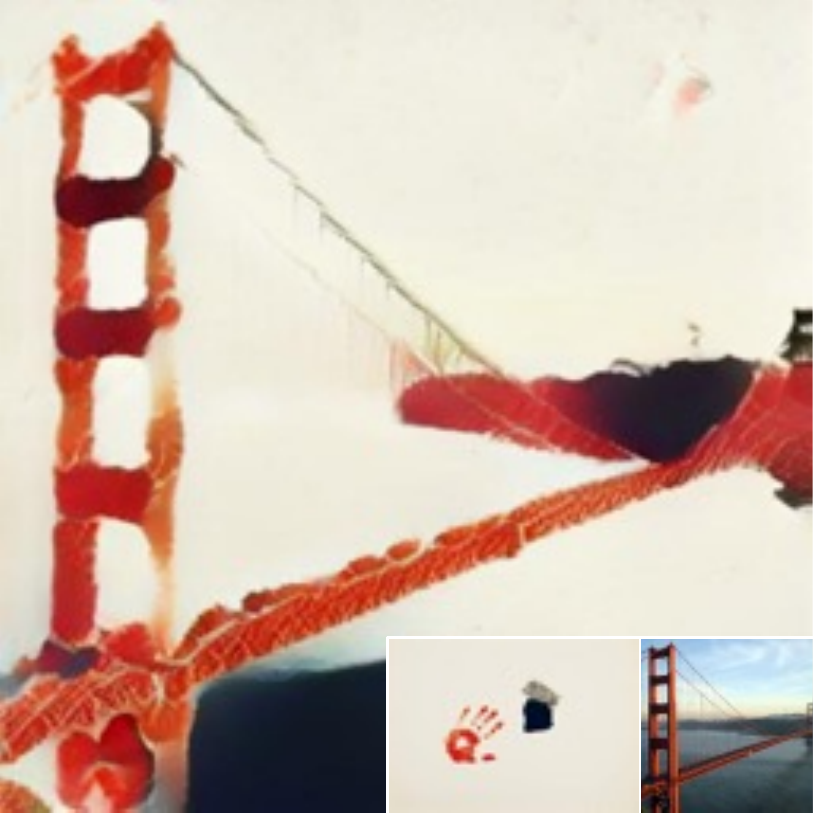}
      \hspace{-4pt}      
    \includegraphics[height=0.12\textwidth,width=0.182\textwidth]{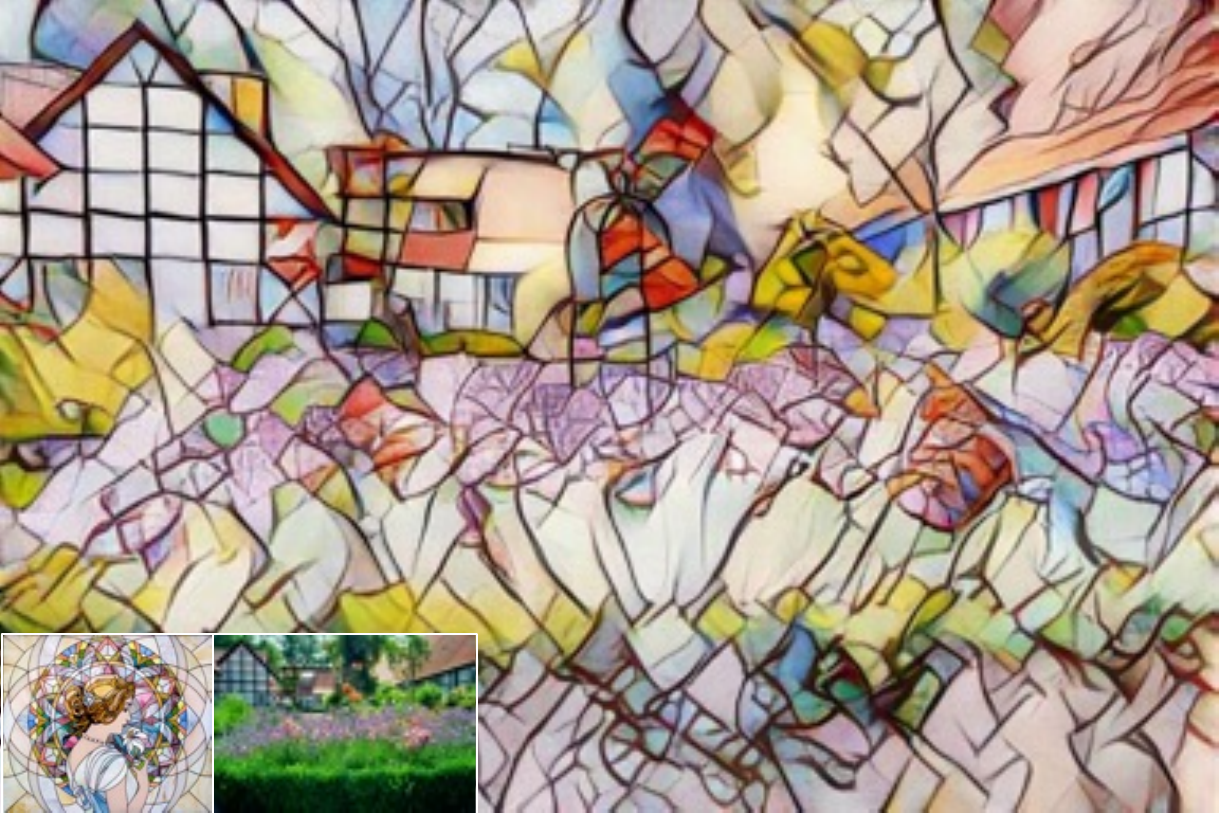}
      \hspace{-4pt}        
    \includegraphics[height=0.12\textwidth,width=0.122\textwidth]{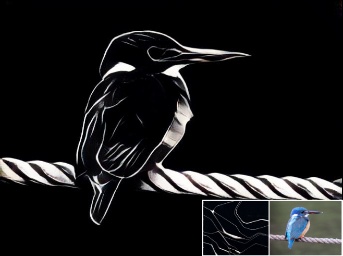} \\ 
\vspace{-7pt} 
    \captionof{figure}{The first two rows show comparisons with state-of-the-art methods. The last row shows results of our proposed method, which can faithfully transfer various styles and preserve the content structure.}\label{fig:Demo}
\end{center}%
}]

\maketitle

\begin{abstract}    \vspace{-10pt} 
Arbitrary style transfer has been demonstrated to be efficient in artistic image generation. Previous methods either globally modulate the content feature ignoring local details, or overly focus on the local structure details leading to style leakage. In contrast to the literature, we propose a new scheme \textit{``style kernel"} that learns {\em spatially adaptive kernels} for per-pixel stylization, where the convolutional kernels are dynamically generated from the global style-content aligned feature and then the learned kernels are applied to modulate the content feature at each spatial position. 
This new scheme allows flexible both global and local interactions between the content and style features such that the wanted styles can be easily transferred to the content image while at the same time the content structure can be easily preserved. To further enhance the flexibility of our style transfer method, we propose a Style Alignment Encoding (SAE) module complemented with a Content-based Gating Modulation (CGM) module for learning the dynamic style kernels in focusing regions. 
Extensive experiments strongly demonstrate that our proposed method outperforms state-of-the-art methods and exhibits superior performance in terms of visual quality and efficiency. 
\end{abstract}

\section{Introduction}\label{sec:intro}


Artistic style transfer~\cite{xu2021drb} refers to a hot computer vision technology that allows us to recompose the content of an image in the style of an artistic work. Figure~\ref{fig:Demo} shows several vivid examples. We might have ever imagined what a photo might look like if it were painted by a famous artist like Pablo Picasso or Van Gogh. Now style transfer is the computer vision technique that turns this into a reality. It has great potential values in various real-world applications and therefore attracts a lot of researchers to constantly put efforts to make progress towards both quality and efficiency.

Most of existing style transfer works~\cite{huang2017arbitrary,jing2020dynamic,park2019arbitrary,chen2022towards,kwon2022clipstyler} either globally modulate the content feature ignoring local details or overly focus on the local structure details leading to style leakage. In particular, \cite{huang2017arbitrary,jing2020dynamic} seeks to match global statistics between content and style images, resulting in inconsistent stylizations that either remain large parts of the content unchanged or contain local content with distorted style patterns. SANet \cite{park2019arbitrary} and AdaAttN \cite{liu2021adaattn} improve the content similarity by learning attention maps that match the semantics between the style and stylized images, but these methods tend to distort object instances when improper style patterns are involved. Recently, IEC \cite{chen2021artistic} adopts contrastive learning to pull close both the content and style representations between input and output images. MAST~\cite{huo2021manifold} seeks to balance the style and content via manifold alignment. Although both IEC and MAST have achieved some success in most cases, they are still far from satisfactory to well balance two important requirements, \ie, style consistency and structure similarity.

\begin{figure*}[th!]
\centering     
\includegraphics[width=0.99\linewidth]{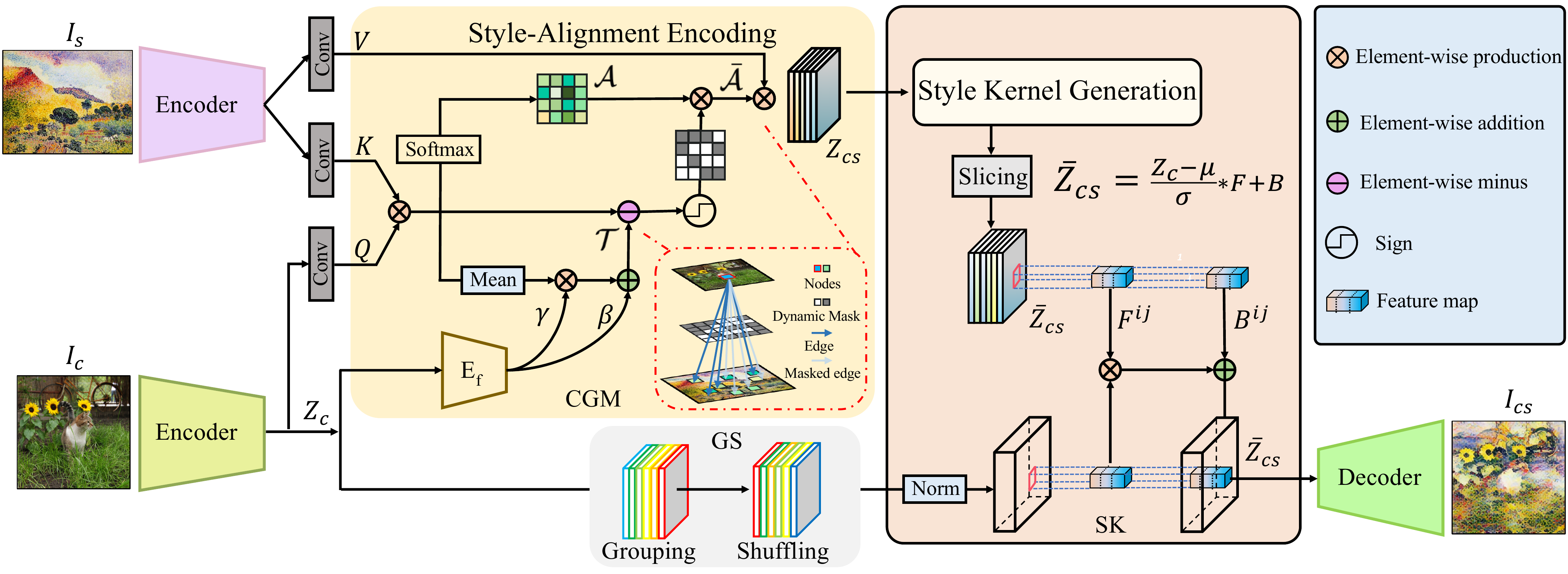}
\vspace{-0.2cm}
\caption{Overview of our proposed scheme \textit{``style kernel"}. A pretrained VGG encoder takes in the content and style image as input to produce the content feature $Z_c$ and style feature $Z_s$. The style-alignment encoding model employs an attention mechanism by taking the $Z_c$ as query and $Z_s$ as key and value to exploit their semantics matching. The content-based gating module (CGM) learns a dynamic threshold and adaptively adjusts attention weights for a globally content-style aligned feature $Z_{cs}$, which is transformed by style kernel generation module to dynamic style kernels. Finally, we conduct convolution operations on the channel-wisely grouped and shuffled (GS) content feature with the learned dynamic style kernels to get $\bar{Z}_{cs}$ fed into the decoder to produce the stylizations. }
\label{fig:overview}
\vspace{-0.4cm}
\end{figure*}

In this paper, we develop a novel style transfer framework in a manner of encoder-decoder structure, as illustrated in Figure~\ref{fig:overview}, which contains two important modules: (1) a Style Alignment Encoding (SAE) module enhanced by Content-based Gating Modulation (CGM), which is used to generate content-style aligned features, and (2) a Style Kernel Generation (SKG) module used to transform the output features of SAE to convolutional kernels. In such a framework, we propose a new style transfer scheme, \textit{``style kernel"}, which treats the features of the style images as dynamic convolutional kernels and then transfer the style information to the content image by convolving the content image by the learned dynamic style kernels. This allows fine-grained local interactions between the style and content features, the flexibility of which makes both style transferring and content structure preserving much easier. 



To enforce the global correlation between the content and style features, we simulate the self-attention mechanism of Transformer~\cite{vaswani2017attention} in the SAE module. This treats the content and style features as the queries and keys of the self-attention operator respectively and computes a large attention map the elements of which measure for each local content context the similarity to each local style context. With the attention map, we can aggregate for each pixel of the content image all the style features in a weighted manner to obtain content-style aligned features. 
We also design a Content-based Gating Modulation (CGM) operator that further thresholds the attention map and zeros the places where similarities are too small, seeking an adaptive number of correlated neighbors as a focusing region for each query point. Then, we use another set of convolutions that constitute the SKG module to further transform the content-style aligned features, and view (reshape) the output of SKG as dynamic style kernels. In order to improve the efficiency of the dynamic convolutions, SKG predicts separable local filters (two 1D filters and a bias) instead of a spatial 2D filter that has more parameters. The learned style kernels already integrate the information of the given content and style images. We further apply the learned dynamic style kernels to the content image feature for transferring the target style to the content image. 

We shall emphasize that unlike \textit{``style code"} in AdaIN~\cite{huang2017arbitrary}, DRB-GAN~\cite{xu2021drb}, and AdaConv~\cite{chandran2021adaptive} modeled as the dynamic parameters (\eg mean, variance, convolution kernel) which are shared over all spatial positions globally without distinguishing the semantic regions, the learned dynamic kernels via our \textit{``style kernel"} are point-wisely modeled upon the globally style-content aligned feature, and therefore is able to make full use of the point-wise semantic structure correlation to modulate content feature for better artistic style transfer. Our learned dynamic style kernels are good at transferring the globally aggregated and locally semantic-aligned style features to local regions. This ensures our style transfer model that works well on consistent stylization with well preserved structure, overcoming the shortages of existing style transfer methods which either globally modulate the content feature ignoring local details (\eg AdaIN, WCT \cite{li2017universal})  or overly focus on the local structure details leading to style leakage (\eg AdaAttN \cite{liu2021adaattn} and MAST~\cite{huo2021manifold}).

The main contributions of this work are 3-fold as follows:
\begin{itemize}
    \vspace{-0.2cm}\item We are the first to propose the \textit{``style kernel"} scheme for artistic style transfer, which converts the globally style-content alignment features into position point-wisely dynamic convolutional kernels to convolve the features of content images together for style transfer.
    \vspace{-0.2cm}\item We design a novel architecture, \ie SAE and CGM, to generate the dynamic style kernels by employing the correlation between the content and style image adaptively.
    \vspace{-0.2cm}\item Extensive experiments demonstrate that our method produces visually plausible stylization results with fine-grained style details and coherence content with the input content images.
\end{itemize}\vspace{-0.2cm}

\section{Related Works}
{\noindent\bf Artistic Style Transfer} was initially solved by Gatys \etal \cite{gatys2016image} a pre-trained neural network to synthesize stylizations in an iterative optimization manner,
and then a bunch of neural style transfer methods~\cite{chen2021artistic,sheng2018avatar,liu2021adaattn,chen2017stylebank,jing2020dynamic,shen2018neural,fu2022language} were developed. In particular, AdaIN~\cite{huang2017arbitrary} introduces the adaptive instance normalization to globally match the statistics between the content and style features. 
For the same purpose, WCT \cite{li2017universal} utilizes the whitening and coloring transforms, and LST \cite{li2018learning} introduces a linear transformation matrix predicted on content and style features. \cite{kalischek2021light,wang2021rethinking,zhang2022exact,cheng2021style} seek different ways to match the feature distribution. 
SANet~\cite{park2019arbitrary} seeks to align the semantics between the content and style features via a self-attention network. 
StrTr$^2$~\cite{deng2022stytr2} resorts to the transformers~\cite{vaswani2017attention,wu2021styleformer}. 
To further preserve the content details,
AdaAttN \cite{liu2021adaattn} perform attentive normalization on per-point basis, IEC \cite{chen2021artistic} employs stylization relations with contrastive learning and 
MAST~\cite{huo2021manifold} introduces manifold alignment to balance the style and content. However, emphasizing on content details prevents to transfer adequate style information to some degree, and results in style leakage. 
Recently, dynamic network based style transfer methods~\cite{chen2017stylebank,shen2018neural,huang2017arbitrary,jing2020dynamic}  extend the model generalization. As a generic extension of AdaIN, AdaConv~\cite{chandran2021adaptive} introduces a parameter network that predicts network parameters used by another network for style transfer. DRB-GAN~\cite{xu2021drb} models \textit{``style code"} as the global mean and variance for dynamic convolutions to conduct style transfer. 
Different from all the above methods, we propose a new scheme \textit{``style kernel"} to learn dynamic convolutional kernels from the style-content alignment feature to point-wisely modulate the content feature for artistic style transfer. 

{\noindent\bf Self-Attention Mechanism} is introduced in Transformer \cite{vaswani2017attention} as a new attention mechanism. Similar to non-local neural networks \cite{wang2018non,vaswani2017attention,han2020survey,jung2020hide}, Transformer directly works on sequences of image patches to aggregate information for long-range dependencies. The recent researches demonstrate it has been successfully applied in various vision tasks like image recognition \cite{dosovitskiy2020image,he2021transreid,chen2020pre,liang2022not}, object detection \cite{carion2020end,wang2022anchor,wang2021pyramid}, image captioning~\cite{Dong:MM2021}, image enhancement \cite{zamir2021restormer,Yu:TOG2021}, and text conditioned image generation \cite{rombach2022high,patashnik2021styleclip}. Our model takes the self-attention mechanism to align the semantics of the style feature to that of the content image.

{\noindent\bf Dynamic Network} consists of convolution layers for which the filter parameters are predicted by another sub-network. Recently, there is rapid progress in vision applications \cite{zhou2019spatio,xu2021drb,jiang2022fast,liu2022dynast,rao2021dynamicvit} that benefits from dynamic network for image/video enhancement. \cite{jo2018deep} employs dynamic upsampling filters for high-resolution image reconstruction. \cite{zhou2019spatio} predicts filters from spatio-temporal features for video deblurring. 
Differently, our StyleKernel learns spatially adaptive kernel for per-pixel stylization.

\begin{figure*}[!ht]
\captionsetup[subfigure]{justification=raggedright, singlelinecheck=false, labelformat=empty}    
\centering
 \includegraphics[width=0.98\textwidth]{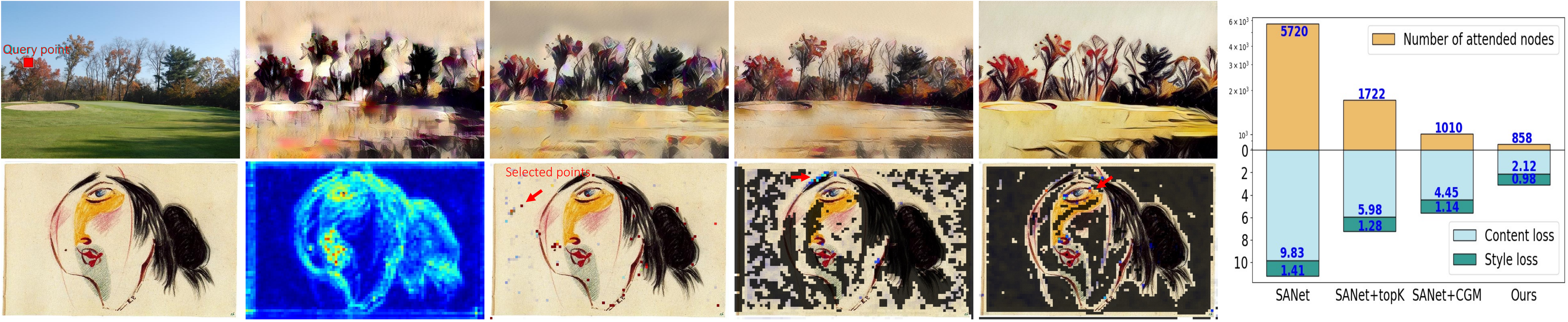}
\raisebox{0.005\textwidth }{\makebox[0.03\textwidth]{\parbox{0.92\linewidth}{\scriptsize \hspace{ 0pt} Content \& Style  \hspace{ 38pt} SANet \hspace{ 44pt} SANet + topK \hspace{ 36pt} SANet + CGM \hspace{ 38pt} Ours}}}
\vspace{-0.3cm}
\caption{The impact of the feature aggregation. Left: We compare the performances of models conducting different feature aggregation strategies. The SANet takes an attention module to aggregate information across all the nodes. The SANet + topK only acquires information from the nodes with top K attention scores. SANet + CGM takes our content-based gating modulation to aggregate information from a dynamic number of nodes. Our CGM is learned to masked out nodes with low attention score, leading to stylizations with well preserved structure content. The second row shows the distributions of selected nodes (highlighted in colors) and the focused areas (masked-out by black rectangles) learned by CGM.
Right: we demonstrate the relation between the content/style loss and the number of selected nodes.}   
\label{fig:attn}
\vspace{-0.3cm}
\end{figure*}

\section{Our Method}
While existing dynamic kernel based methods \cite{huang2017arbitrary,xu2021drb,chandran2021adaptive} can efficiently generate stylizations, we observe that their visual-quality is limited in two aspects. (1) As the dynamic kernels are predicted based on the style feature without distinguishing the content-style correlation, the transferred style only globally reflects the style characters of the style image, ruining local content details. (2) The completeness of semantic regions is not well preserved. This is because the dynamic kernels used by the layers are spatially shared. 

To alleviate these problems, we propose to align the semantics of the style image to those of the content image and transform the content features for per-pixel stylization in a separable convolution manner. The pipeline of our proposed model is shown in Figure \ref{fig:overview}. The content image $I_c$ and style image $I_s$ are first fed into a fixed \textit{VGG} encoder to produce the content and style feature maps $Z_c \in \mathbb{R}^{H_c\times W_c \times C_{in}}$ and $Z_s \in \mathbb{R}^{H_s\times W_s \times C_{in}}$, where $H$, $W$ and $C_{in}$ represent the height, width and channel size of the feature map, respectively. 
Then in SAE, we align the content and style features by using the content features $Z_c$ as the queries and the style features $Z_s$ as the keys in a self-attention mechanism. The CGM module is conducted on the attention map to seek a focusing region for each query point. 
Thereafter, we predict by SKG the dynamic style kernels $F$ from the content-style aligned features $Z_{cs}$ . Finally, we group and shuffle (GS) $Z_c$ to break down the correlation within the channel-wise feature and apply convolutions parameterized by the learned kernels to the grouped and shuffled $Z_c$ to obtain $\bar{Z}_{cs}$ which is further decoded to the target stylized image by a decoder.

\subsection{Style Alignment Encoding} \label{SAE}
This module takes the content and style features as input, aligns between them by learning content-style alignment attention, and finally attentively adjusts the attention map by a content-based gating modulation module.


\noindent{\bf Content-Style Alignment Attention.}
By taking the content feature $Z_c$ as the query, and the style feature $Z_s$ as the key and value, we compute the content-style alignment attention that seeks to aggregate information from the style feature map according to the matched content information. This warps the style feature to align with the content feature. To this end, we calculate the pairwise feature correlation $\mathcal{M}^{s \to c} \in \mathbb{R}^{H_cW_c \times H_sW_s}$. This results in an attention map $\mathcal{A} \in \mathbb{R}^{H_cW_c \times H_sW_s}$ by:
\begin{align}
\small
\label{eq:attention-matrix}
\begin{split}
\mathcal{A}(u,v) = \texttt{softmax}(\alpha \mathcal{M}^{s \to c}) =\texttt{softmax}(\alpha  \frac{Z_c(u)^TZ_s(v)}{||Z_c(u)||||Z_s(v)||}), 
\end{split}
\end{align}
where $\alpha$ is the coefficient that controls the sharpness of the $\texttt{softmax}$. $Z_c(u)$ and $Z_s(v)$ stands for the feature of $Z_c$ and $Z_s$ at position $u \in \mathbb{R}^{H_cW_c} $ and $v \in \mathbb{R}^{H_sW_s}$. The $u$-th row of the attention map $\mathcal{A}$ represents similarities between the $u$-th node of the content feature and all the nodes of the style feature.

To align the semantics of the style feature to that of the content feature, we aggregate information in $Z_s$ based on the attention score map $\mathcal{A}$ and obtain $Z_{cs} \in \mathcal{R}^{H_c W_c\times C}$ by calculating their weighted average:
\begin{align}
\small
\label{eq:content-style-aligned-feature}
\begin{split}
Z_{cs}(u) = \sum_{v} \mathcal{A}(u,v) \cdot Z_s(v).
\end{split}
\end{align}
where $Z_{cs}(u)$ represents the feature at position $u$ of $Z_{cs}$. In the following, we denote the $Z_{cs} \in \mathcal{R}^{H_c \times W_c\times C}$, which is obtained by reshaping. 

In Eq.~\ref{eq:content-style-aligned-feature}, all style features are taken into consideration when computing the content-style aligned features $Z_{cs}$. We find that this yields inferior results because many irrelevant features are involved into the computation.
We therefore introduce the Content-based Gating Modulation (CGM) operator which creates dynamic thresholds that further filter out irrelevant style features. 


\begin{table*}[ht!]
\centering
\setlength{\tabcolsep}{2.2pt}{
\caption{ Average inference time, measured
on a Tesla V100 GPU, for different methods with a batch size of 1 and an input image of $256 \times 256$ and $512 \times 512$.  }\label{fig:time}
\vspace{-0.3cm} 
    \begin{tabular}{l|cccc c c c c c c c}
\toprule
	\small Method&\small AdaIN'17 &\small LST'19&\small AvatarNet'18 &\small SANet'19  &\small IEC'21 & \small DRB-GAN'21 &\small AdaAttN'21 &\small MAST'21 &\small AdaConv'21 &\small StrTr$^2$'22&\small Ours \\
    \midrule
$T_{256}$ $\downarrow$& 0.004&0.004&0.873&0.011&0.012&0.005&0.024&0.126&0.019&0.034&0.017\\
    \midrule
 $T_{256}$ $\downarrow$ & 0.005&0.005&0.962&0.037&0.021&0.006&0.042&0.239&0.036&0.086&0.035\\
 \bottomrule
	\end{tabular}}
 \vspace{-0.45cm}	
\end{table*}

\noindent{\bf Content-based Gating Modulation} is used to further adapt the attention map by the content feature $Z_c$. Taking $Z_c$ as input, we employ a convolution network $E_f$ to generate a scale parameter $\lambda$ and a bias parameter $\beta$. We then perform a row average operator on the attention matrix $\mathcal{A}$ by:
\begin{equation}
    \mathcal{A}_r(i) = \frac{1}{H_c} \sum_{j=1}^{H_c} \mathcal{A}_{i,j}, \forall i \in[1, W_c].
\end{equation}
where $\mathcal{A}_r$ is the obtained column vector after the average operator. Then, we obtain the threshold matrix $\mathcal{T}$ by the following equation:
\begin{equation}
   \mathcal{T} = \mathcal{A}\left( \lambda\mathcal{A}_r +\beta\right).
\end{equation}
Finally, we update the attention $\mathcal{A}$ by:
\begin{align}
\small
\begin{split}
\bar{\mathcal{A}} = \mathcal{A}\cdot \texttt{Sign}(\mathcal{A} -\mathcal{T}), 
\end{split}
\end{align}
where the dot indicates element-wise multiplication and $\texttt{Sign}$ is a function returning 1 given positive input and 0 otherwise. We use $\bar{\mathcal{A}}$ instead of $\mathcal{A}$ in Eq.~\ref{eq:content-style-aligned-feature} to compute $Z_{cs}$.

$\bar{\mathcal{A}}$ is a dynamically learned mask, with which our model creates focused areas and filter out irrelevant style features. This allows our model to aggregate features from fairly correlated features, which is essential to preserve the completeness of semantic regions. In Figure \ref{fig:attn} we compare the intermediate visualizations and stylized results. We see that the CGM works to recognize focused areas, where the style information is aggregated by the query point. By contrast, the attention learned by the SANet is distributed over the entire image, including visually less correlated points, such as corners and background. 


\subsection{Style Kernel Generation}  \label{SDC}
We propose the Style Kernel (SK) Generation network to further transform the content-style aligned features of SAE to dynamic style kernels.
It takes the globally content-style aligned feature $Z_{cs}$ as input to predict the dynamic style filters $F, B = \{F_1^{ij}, F_2^{ij}, B^{ij}\}$.


Given the content-style aligned feature $Z_{cs}$, we take two convolution layers to predict the filters $F \in \mathbb{R}^{H_c \times W_c \times C \times 2k}$, which has two 1-dim filters $F_1 = \{F_1^{ij}\}$ and $F_2 = \{F_2^{ij}\}$ of the sizes $k \times 1$ and $1 \times k$, respectively, and one bias vector $B=\{B^{ij}\} \in \mathbb{R}^{H_c \times W_c \times C \times 1} $.
Finally, we conduct convolution operations on the content feature $Z_c$ with the learned dynamic style kernels $F_1$, $F_2$ and $B$. The output after the convolutions is the feature called $\bar{Z}_{cs}$: 
\begin{align}
\small
\label{eqx}
\begin{split}
\bar{Z}_{cs} =  \frac{Z_{c} - \mu}{\sigma} * F_1*F_2 + B,
\end{split}
\end{align}
where $\mu$ and $\sigma$ are the affine parameters in normalization layer and $*$ is the convolution operator. $\bar{Z}_{cs}$ is then passed to the decoder to generate the stylized image $I_{cs}$.

\begin{figure*}[ht!]
\captionsetup[subfigure]{justification=raggedright, singlelinecheck=false, labelformat=empty}  
\centering

  \includegraphics[height=0.09\textwidth,width=0.10\textwidth]{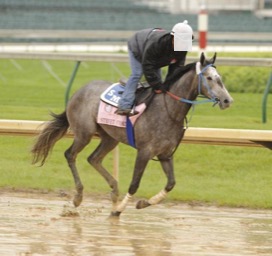} 
    \hspace{-4pt}
  \includegraphics[height=0.09\textwidth,width=0.08\textwidth]{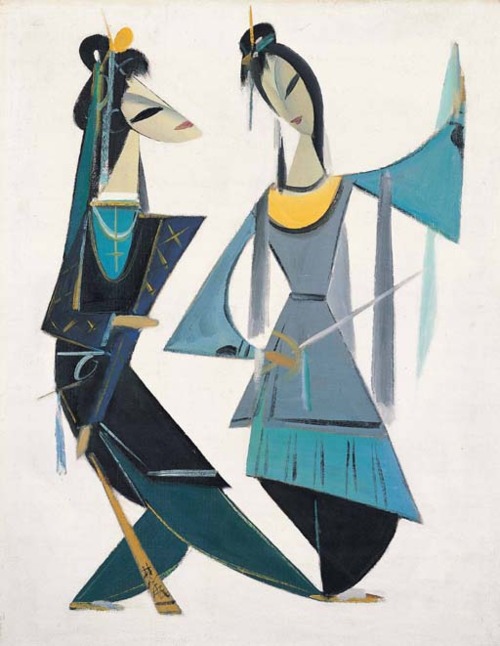} 
  \hspace{-4pt}  
  \includegraphics[height=0.09\textwidth,width=0.10\textwidth]{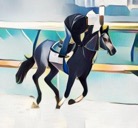}  
  \hspace{-4pt}
  \includegraphics[height=0.09\textwidth,width=0.10\textwidth]{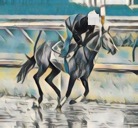}  
  \hspace{-4pt} 
  \includegraphics[height=0.09\textwidth,width=0.10\textwidth]{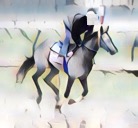}  
  \hspace{-4pt} 
  \includegraphics[height=0.09\textwidth,width=0.10\textwidth]{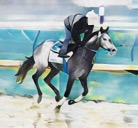}  
  \hspace{-4pt} 
  \includegraphics[height=0.09\textwidth,width=0.10\textwidth]{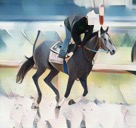}  
  \hspace{-4pt} 
   \includegraphics[height=0.09\textwidth,width=0.10\textwidth]{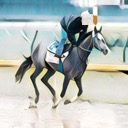}  
  \hspace{-4pt} 
  \includegraphics[height=0.09\textwidth,width=0.10\textwidth]{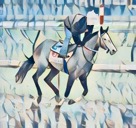}  
  \hspace{-4pt}   
   \includegraphics[height=0.09\textwidth,width=0.10\textwidth]{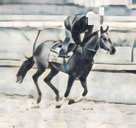}  \\

  
  \includegraphics[height=0.09\textwidth,width=0.1\textwidth]{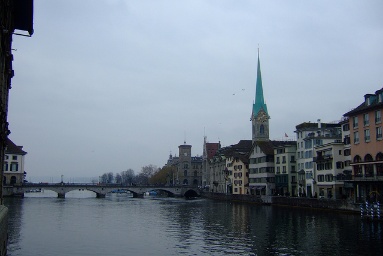}  
  \hspace{-4pt}
  \includegraphics[height=0.09\textwidth,width=0.08\textwidth]{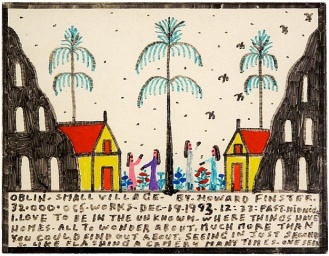}  
  \hspace{-4pt}
  \includegraphics[height=0.09\textwidth,width=0.1\textwidth]{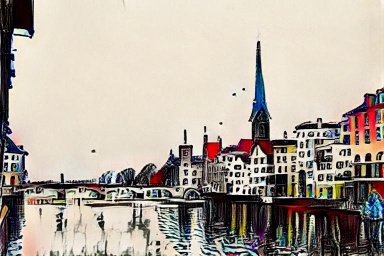} 
  \hspace{-4pt}
  \includegraphics[height=0.09\textwidth,width=0.1\textwidth]{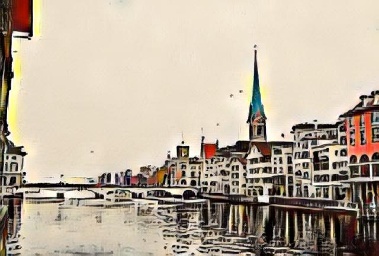}  
  \hspace{-4pt}  
  \includegraphics[height=0.09\textwidth,width=0.1\textwidth]{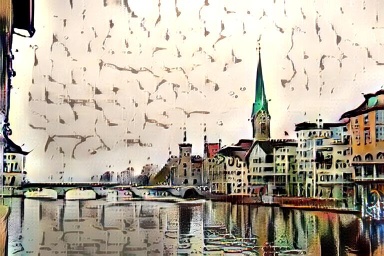}  
  \hspace{-4pt} 
   \includegraphics[height=0.09\textwidth,width=0.1\textwidth]{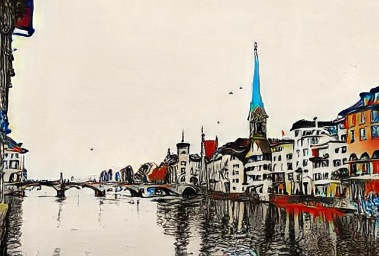}  
  \hspace{-4pt}
  \includegraphics[height=0.09\textwidth,width=0.1\textwidth]{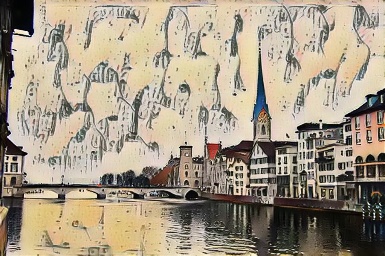}  
  \hspace{-4pt} 
     \includegraphics[height=0.09\textwidth,width=0.10\textwidth]{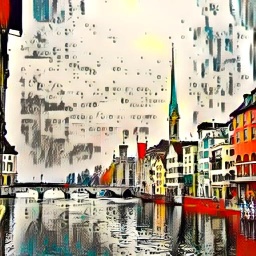}  
  \hspace{-4pt} 
  \includegraphics[height=0.09\textwidth,width=0.10\textwidth]{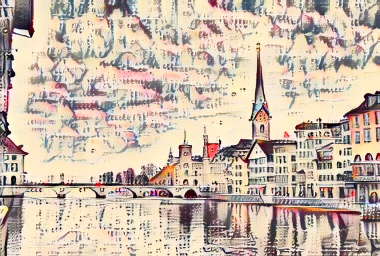}  
  \hspace{-4pt} 
   \includegraphics[height=0.09\textwidth,width=0.1\textwidth]{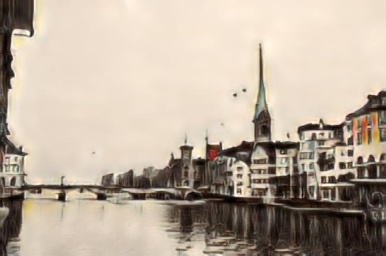}  \\
  
  
 \includegraphics[height=0.09\textwidth,width=0.1\textwidth]{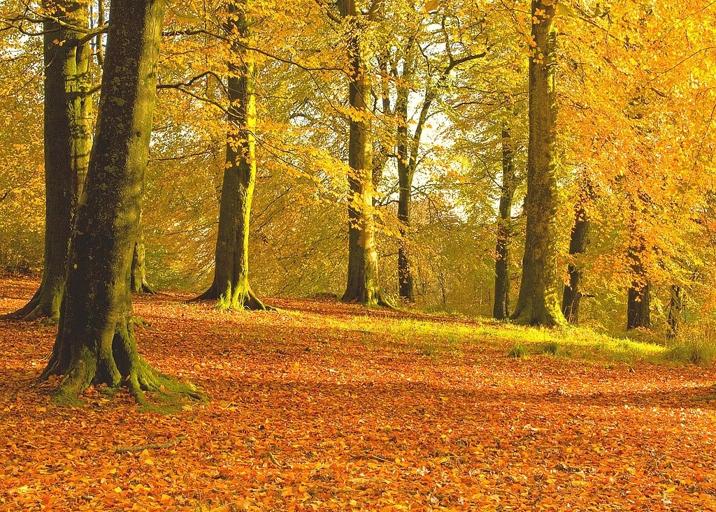}
  \hspace{-4pt} 
 \includegraphics[height=0.09\textwidth,width=0.08\textwidth]{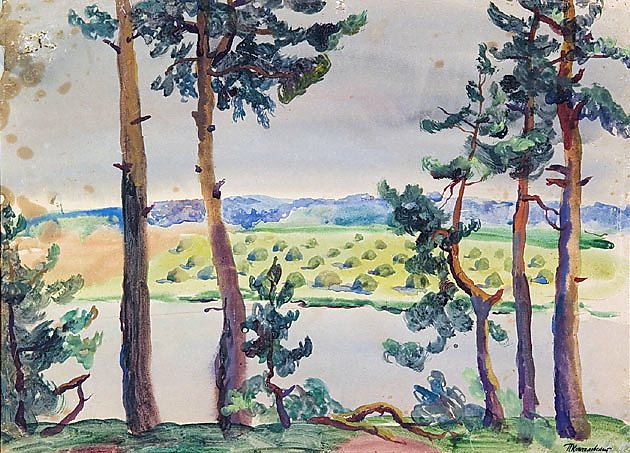}
  \hspace{-4pt}
 \includegraphics[height=0.09\textwidth,width=0.1\textwidth]{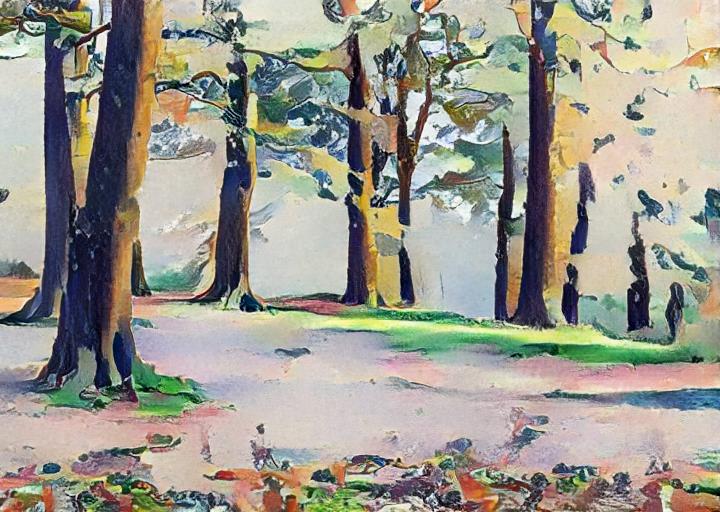}
  \hspace{-4pt} 
 \includegraphics[height=0.09\textwidth,width=0.1\textwidth]{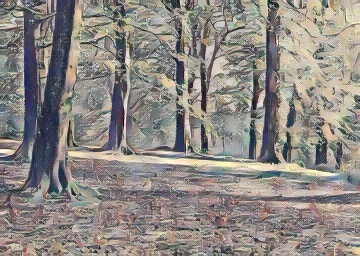}
  \hspace{-4pt}  
 \includegraphics[height=0.09\textwidth,width=0.1\textwidth]{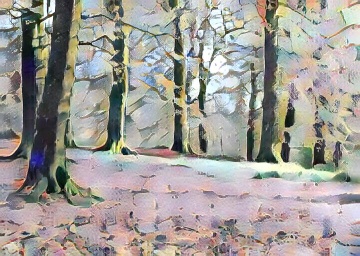}
  \hspace{-4pt} 
 \includegraphics[height=0.09\textwidth,width=0.1\textwidth]{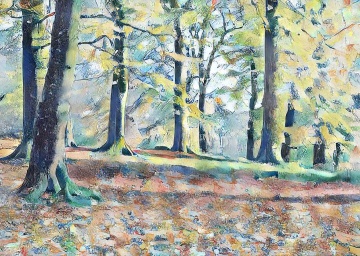}
  \hspace{-4pt} 
 \includegraphics[height=0.09\textwidth,width=0.1\textwidth]{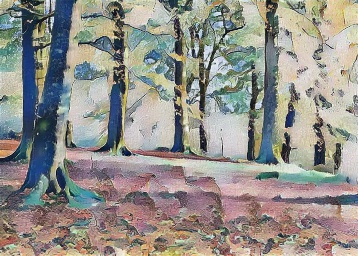}
  \hspace{-4pt}
   \includegraphics[height=0.09\textwidth,width=0.10\textwidth]{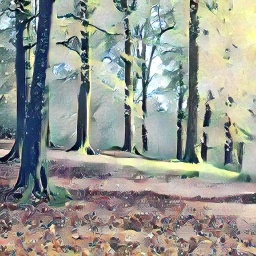}  
  \hspace{-4pt} 
  \includegraphics[height=0.09\textwidth,width=0.10\textwidth]{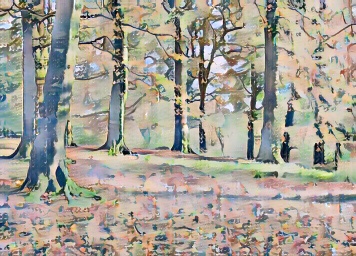}  
  \hspace{-4pt}   
 \includegraphics[height=0.09\textwidth,width=0.1\textwidth]{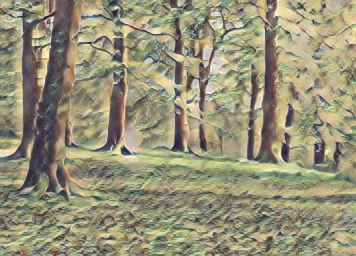}\\ 

 \includegraphics[height=0.09\textwidth,width=0.1\textwidth]{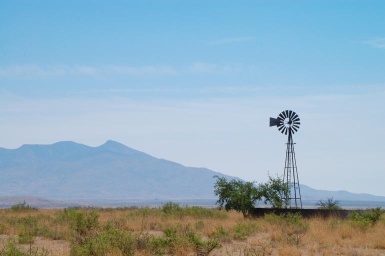}
  \hspace{-4pt} 
 \includegraphics[height=0.09\textwidth,width=0.08\textwidth]{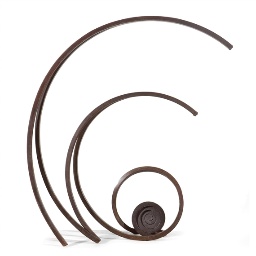}
  \hspace{-4pt}
 \includegraphics[height=0.09\textwidth,width=0.1\textwidth]{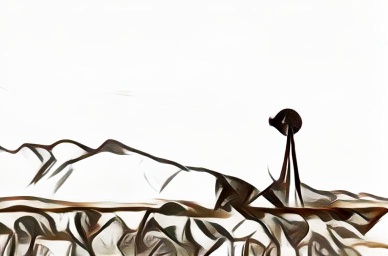}
  \hspace{-4pt} 
 \includegraphics[height=0.09\textwidth,width=0.1\textwidth]{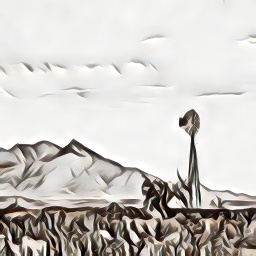}
  \hspace{-4pt}  
 \includegraphics[height=0.09\textwidth,width=0.1\textwidth]{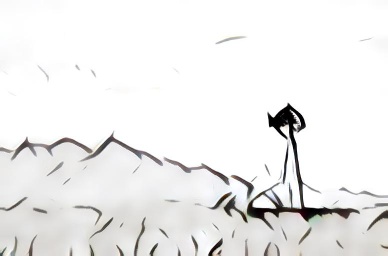}
  \hspace{-4pt} 
 \includegraphics[height=0.09\textwidth,width=0.1\textwidth]{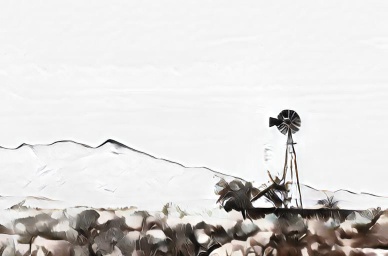}
  \hspace{-4pt} 
 \includegraphics[height=0.09\textwidth,width=0.1\textwidth]{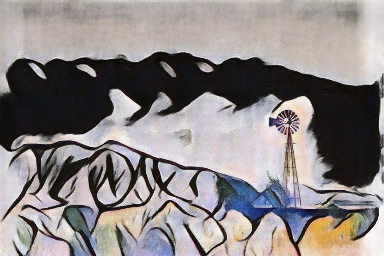}
  \hspace{-4pt}
   \includegraphics[height=0.09\textwidth,width=0.10\textwidth]{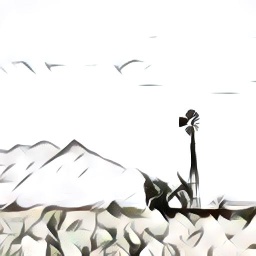}  
  \hspace{-4pt} 
  \includegraphics[height=0.09\textwidth,width=0.10\textwidth]{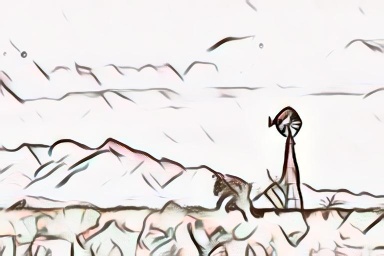}  
  \hspace{-4pt}   
 \includegraphics[height=0.09\textwidth,width=0.1\textwidth]{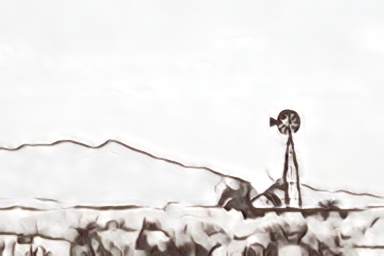}\\ 

 \includegraphics[height=0.09\textwidth,width=0.1\textwidth]{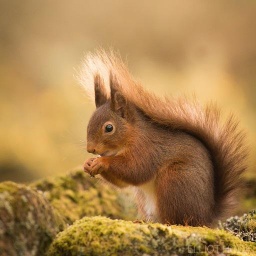}
  \hspace{-4pt} 
 \includegraphics[height=0.09\textwidth,width=0.08\textwidth]{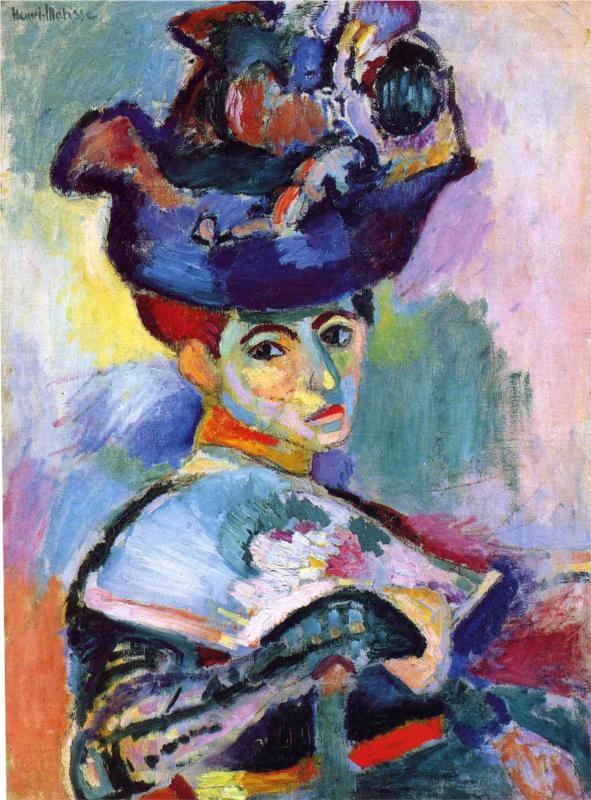}
  \hspace{-4pt}
 \includegraphics[height=0.09\textwidth,width=0.1\textwidth]{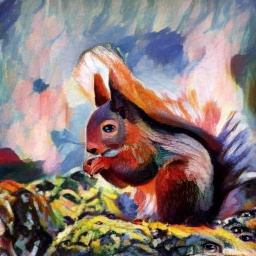}
  \hspace{-4pt} 
 \includegraphics[height=0.09\textwidth,width=0.1\textwidth]{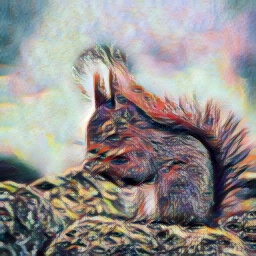}
  \hspace{-4pt}  
 \includegraphics[height=0.09\textwidth,width=0.1\textwidth]{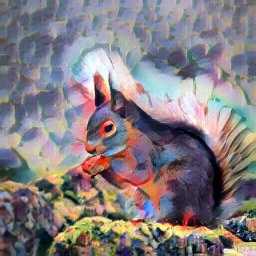}
  \hspace{-4pt} 
 \includegraphics[height=0.09\textwidth,width=0.1\textwidth]{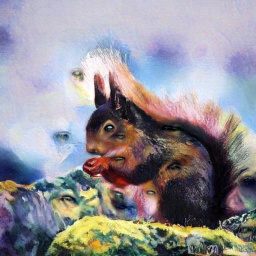}
  \hspace{-4pt} 
 \includegraphics[height=0.09\textwidth,width=0.1\textwidth]{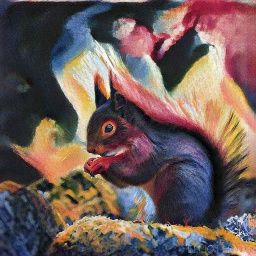}
  \hspace{-4pt}
   \includegraphics[height=0.09\textwidth,width=0.10\textwidth]{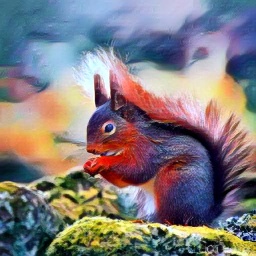}  
  \hspace{-4pt} 
  \includegraphics[height=0.09\textwidth,width=0.10\textwidth]{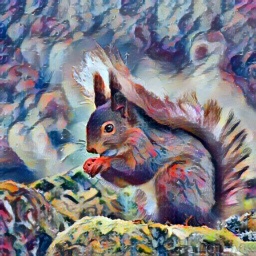}  
  \hspace{-4pt}   
 \includegraphics[height=0.09\textwidth,width=0.1\textwidth]{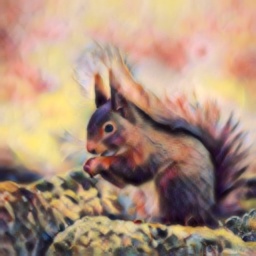}\\ 


 \includegraphics[height=0.09\textwidth,width=0.1\textwidth]{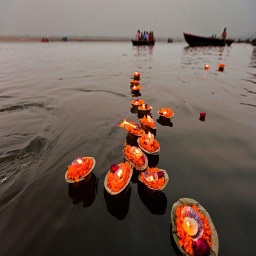}
  \hspace{-4pt} 
 \includegraphics[height=0.09\textwidth,width=0.08\textwidth]{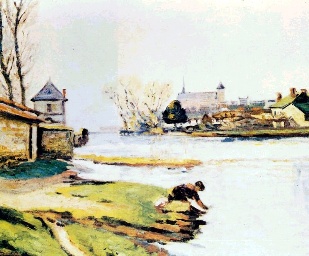}
  \hspace{-4pt}
 \includegraphics[height=0.09\textwidth,width=0.1\textwidth]{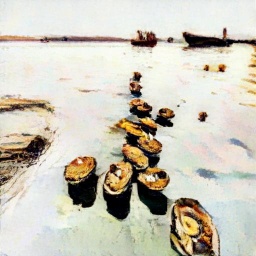}
  \hspace{-4pt} 
 \includegraphics[height=0.09\textwidth,width=0.1\textwidth]{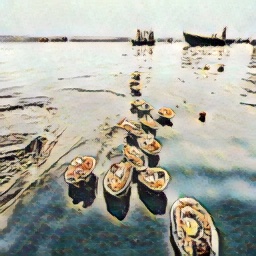}
  \hspace{-4pt}  
 \includegraphics[height=0.09\textwidth,width=0.1\textwidth]{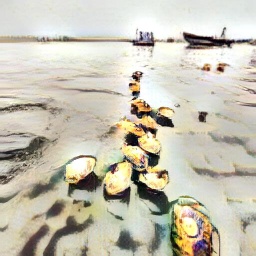}
  \hspace{-4pt} 
 \includegraphics[height=0.09\textwidth,width=0.1\textwidth]{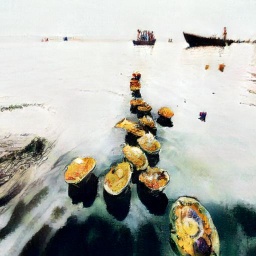}
  \hspace{-4pt} 
 \includegraphics[height=0.09\textwidth,width=0.1\textwidth]{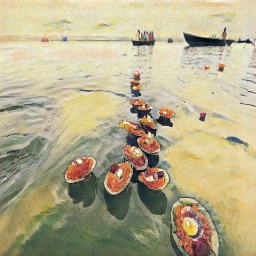}
  \hspace{-4pt}
   \includegraphics[height=0.09\textwidth,width=0.10\textwidth]{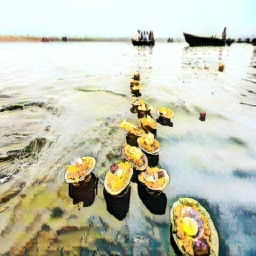}  
  \hspace{-4pt} 
  \includegraphics[height=0.09\textwidth,width=0.10\textwidth]{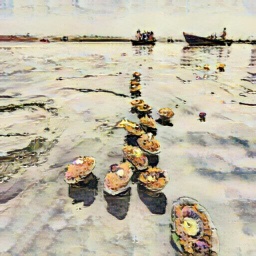}  
  \hspace{-4pt}   
 \includegraphics[height=0.09\textwidth,width=0.1\textwidth]{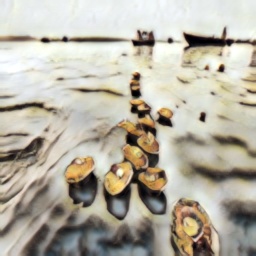}\\ 

 \includegraphics[height=0.09\textwidth,width=0.1\textwidth]{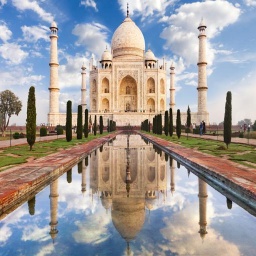}
  \hspace{-4pt} 
 \includegraphics[height=0.09\textwidth,width=0.08\textwidth]{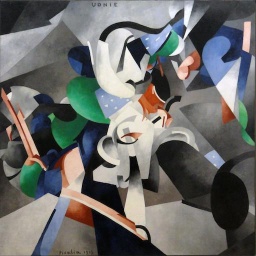}
  \hspace{-4pt}
 \includegraphics[height=0.09\textwidth,width=0.1\textwidth]{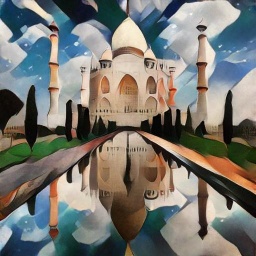}
  \hspace{-4pt} 
 \includegraphics[height=0.09\textwidth,width=0.1\textwidth]{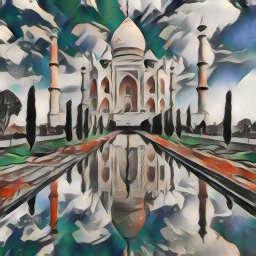}
  \hspace{-4pt}  
 \includegraphics[height=0.09\textwidth,width=0.1\textwidth]{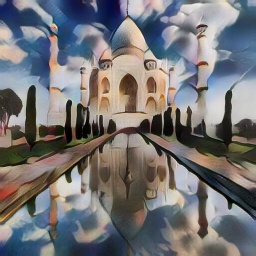}
  \hspace{-4pt} 
 \includegraphics[height=0.09\textwidth,width=0.1\textwidth]{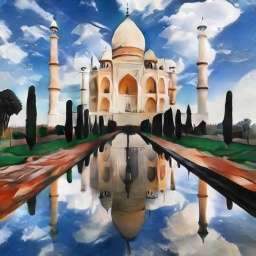}
  \hspace{-4pt} 
 \includegraphics[height=0.09\textwidth,width=0.1\textwidth]{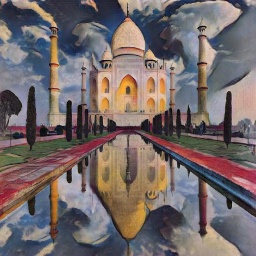}
  \hspace{-4pt}
   \includegraphics[height=0.09\textwidth,width=0.10\textwidth]{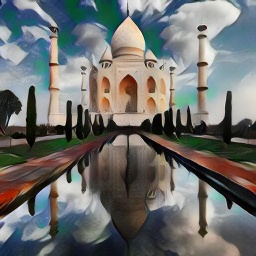}  
  \hspace{-4pt} 
  \includegraphics[height=0.09\textwidth,width=0.10\textwidth]{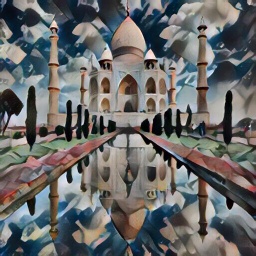}  
  \hspace{-4pt}   
 \includegraphics[height=0.09\textwidth,width=0.1\textwidth]{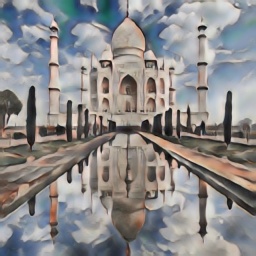}\\ 
 \raisebox{0.005\textwidth }{\makebox[0.03\textwidth]{\parbox{0.95\linewidth}{\scriptsize \hspace{ 0pt} Content\hspace{26pt} Style \hspace{ 26pt} Ours  \hspace{23pt} AdaAttN'21 \hspace{ 18pt} AdaIN'17 \hspace{ 24pt} IEC'21 \hspace{14pt} DRB-GAN'21 \hspace{14pt} StrTr$^2$'22\hspace{18pt} AdaConv'21 \hspace{13pt} MAST'21}}}
 
\vspace{-0.3cm}
\caption{Qualitative performance comparison on stylized results.}
\label{fig:comp_arbitrary_style_transfer}
\vspace{-0.3cm} 
\end{figure*} 

\begin{table*}[ht!]
\centering
\setlength{\tabcolsep}{2.2pt}{
\caption{ Quantitative comparison with state-of-the-art methods. We compute the average style loss and the LPIPS score of different methods to indicate how
well the style and content are transferred.}\label{fig:score}
\vspace{-0.3cm} 
    \begin{tabular}{l|cccc c c c c c c c}
\toprule
	\small Method&\small AdaIN'17 &\small LST'19&\small AvatarNet'18 &\small SANet'19  &\small IEC'21 & \small DRB-GAN'21 &\small AdaAttN'21 &\small MAST'21 &\small AdaConv'21 &\small StrTr$^2$'22&\small Ours \\
    \midrule
\small Sty Loss$\downarrow$&1.77&2.67&5.91&1.41&1.81&1.51&1.14&-&1.04&1.07&0.98  \\
    \midrule
 \small LPIPS $\downarrow$& 0.37&0.33&0.34&0.36&0.31&0.33&0.32&0.32&0.42&0.33&0.30\\
 \bottomrule
	\end{tabular}}
 \vspace{-0.45cm}	
\end{table*}

\bfsection{Remarks} Note that different from \textit{``style code"}~\cite{xu2021drb} modeled as the parameters (\eg mean and variance) which are shared over all spatial positions globally without taking different semantic regions into consideration, our learned dynamic style kernels are position point-wisely inferred from the globally style-content aligned feature, and therefore are able to fully explore the point-wise semantic structure correlation to modulate local content feature for better artistic style transfer. Our dynamic style kernels make it possible to transfer both the globally aggregated content-style aligned features to local regions in the content images. This ensures our new scheme \textit{``style kernel"} works well on consistent stylization with well preserved structure, and alleviates the shortages of existing style transfer methods~\cite{huang2017arbitrary,li2017universal,liu2021adaattn,huo2021manifold} which either globally modulate the content feature ignoring local details or overly focus on the local structure details leading to style leakage.

\subsection{Loss Functions} \label{LI}
The loss function is formulated with an adversarial loss $\mathcal{L}_{adv}$, a reconstruction loss $\mathcal{L}_{rec}$, a REMD loss $\mathcal{L}_{REMD}$, a style loss $\mathcal{L}_{sty}$ \cite{johnson2016perceptual} and a content loss $\mathcal{L}_{cont}$ \cite{park2019arbitrary} as:
\begin{align}
\small
\label{eqx}
\begin{split}
\mathcal{L} =\mathcal{L}_{adv}+\lambda_{rec}\mathcal{L}_{rec}+\lambda_{cont}\mathcal{L}_{cont}&+\lambda_{sty}\mathcal{L}_{sty} +\lambda_{REMD}\mathcal{L}_{REMD}
\end{split}
\end{align}
where $\lambda_{adv}$, $\lambda_{rec}$, $\lambda_{cont}$, $\lambda_{sty}$ and $\lambda_{REMD}$ are the hyper-parameters used to balance the loss during training.



\noindent{\bf Reconstruction Loss} $\mathcal{L}_{rec}$ is defined as: 
\begin{align}
\small
\label{eqx}
\begin{split}
\mathcal{L}_{rec} &= \lambda_{rec1}(||I_{cc}-I_c||_2 + ||I_{ss}-I_s||_2)\\ &+ \lambda_{rec2}\sum_{l=1}^L (||\phi_l(I_{cc})-\phi_l(I_c)||_2+||\phi_l(I_{ss})-\phi_l(I_s)||_2),
\end{split}
\end{align}
where $I_{cc}$ and $I_{ss}$ are the reconstructed images when both the input content and style images are $I_c$ and $I_s$, respectively. $\phi_l$ refers to the $l_{th}$ layer in VGG. We chose features from Relu2\_1, Relu3\_1, Relu4\_1 and Relu5\_1 layers.

\noindent{\bf REMD Loss} $\mathcal{L}_{REMD}$~\cite{kolkin2019style} adapts the relaxed earth mover distance (REMD) to align the manifold surface of style features. It is formulated as:
\begin{align}
\small
\label{eqx}
\begin{split}
\mathcal{L}_{REMD} =\max(\frac{1}{W_s H_s} \sum_{i}\min_{j}C_{ij}, \frac{1}{W_c H_c} \sum_{j}\min_{i}C_{ij}),
\end{split}
\end{align}
where $C_{ij}$ denotes the pair-wise cosine distance matrix between the $i_{th}$ and $j_{th}$ feature vector in $Z_{cs}$ and $Z_s$.

\section{Experiments}
We train our proposed method on content images from the MS-COCO \cite{lin2014microsoft} dataset, and style images from WikiArt \cite{karayev2013recognizing} database. Each dataset has around 80k images. The training images are first resized to 512, and then $256\times256$ patches are randomly cropped from the images as inputs. Note that our model can be applied to images of any resolution at the testing stage. Our model is implemented by PyTorch \cite{paszke2017automatic}. The Adam \cite{kingma2014adam} is adopted as the optimization solver. We train our model for 160k iterations with a batch size of 16. The learning rate is set to 0.0001.

The hyperparameters in loss functions are set to $\lambda_{cont} = 1$,  $\lambda_{rec} = 1$, $\lambda_{sty} = 1$, $\lambda_{rec1}=20$, $\lambda_{rec2}=0.5$, $\lambda_{REMD} = 3$, and $\lambda_{adv} = 1$. The encoder is a pretrained VGG-16 network \cite{simonyan2014very}, whose parameters are fixed during model training. A multi-scale discriminator is adopted from \cite{wang2018high}. We update the discriminator one time after two generator iterations.

\subsection{Comparison with State-of-the-Art Methods}
\bfsection{Qualitative Evaluation} 
As shown in Figure \ref{fig:comp_arbitrary_style_transfer}, we present qualitative results of different style transfer methods {\em i.e.}, AdaConv~\cite{chandran2021adaptive}, DRB-GAN~\cite{xu2021drb}, statistics matching based method like AdaIN \cite{huang2017arbitrary}, attention based methods including AdaAttN \cite{liu2021adaattn} and IEC \cite{chen2021artistic}, and feature modification methods like StrTr$^2$~\cite{deng2022stytr2} and MAST~\cite{huo2021manifold}. For a fair comparison, we deploy all competing methods for style transfer on images of the smaller dimension resized to $512$ while the aspect ratio is preserved. Note that we use the public released code and follow the default configurations for testing.

The comparison in Figure~\ref{fig:comp_arbitrary_style_transfer} shows the outperformance of our method in terms of visual quality. These images produced by our method faithfully reflect the style characters (\eg, stroke sizes and colors) with no artifact in the regions, and most importantly, they preserve the structural similarity of the content images. On the contrary, AdaIN \cite{huang2017arbitrary} and MAST~\cite{huo2021manifold} fail to generate sharp details and fine strokes. AdaConv~\cite{chandran2021adaptive} yields distorted patterns as it cannot always recover the original style patterns. We also observe non-negligible artificial structures in those images obtained by StrTr$^2$~\cite{deng2022stytr2}, DRB-GAN \cite{xu2021drb} and AdaIN \cite{huang2017arbitrary}. These methods struggle to preserve the consistency in semantic regions such as the sky. In addition, AdaAttN~\cite{liu2021adaattn} and IEC~\cite{chen2021artistic} and MAST~\cite{huo2021manifold} tend to overly preserve the structural similarity and fail to transfer the style to the content image, which makes the results look like the content images.


\begin{figure*}[ht!]
\captionsetup[subfigure]{justification=raggedright, singlelinecheck=false, labelformat=empty}    
 \centering
  \includegraphics[width=\textwidth]{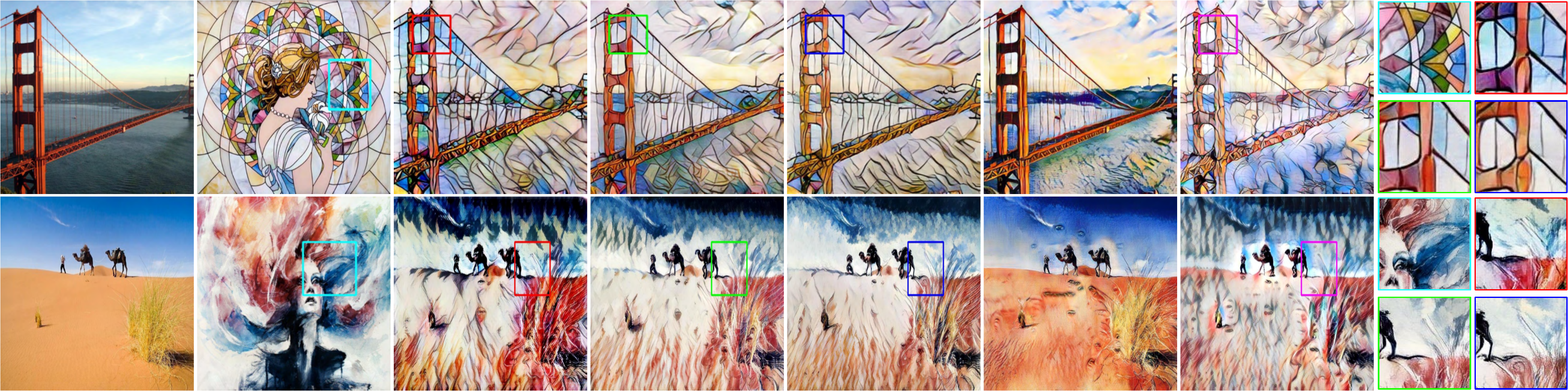}  
 \centering
\raisebox{0.005\textwidth }{\makebox[0.03\textwidth]{\parbox{0.95\linewidth}{\scriptsize \hspace{ 8pt} Content\hspace{40pt} Style \hspace{ 45pt}  w/ GS \hspace{38pt} w/o GS \hspace{ 40pt} w/ SK \hspace{ 40pt} w/o SK \hspace{ 40pt} SANet \hspace{ 25pt} zoom-in details}}} 
  \vspace{-0.3cm}
\caption{Qualitative performance comparison on stylized results of different model variants. The results of SANet are also listed as reference. \textit{Please zoom in to observe the detailed difference.}}
\label{fig:comp_AB}
\vspace{-0.3cm} 
\end{figure*}  

\begin{figure}[ht!]
\captionsetup[subfigure]{justification=raggedright, singlelinecheck=false, labelformat=empty} 
\centering
  \includegraphics[height=0.1\textwidth,width=0.11\textwidth]{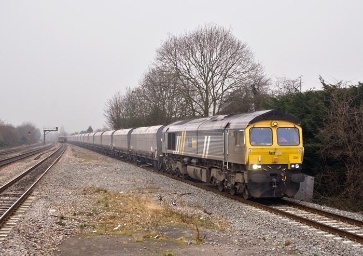}  
  \hspace{-4pt} 
 \includegraphics[height=0.1\textwidth,width=0.11\textwidth]{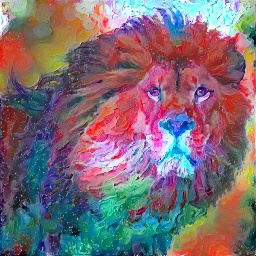}
  \hspace{-4pt}
  \includegraphics[height=0.1\textwidth,width=0.11\textwidth]{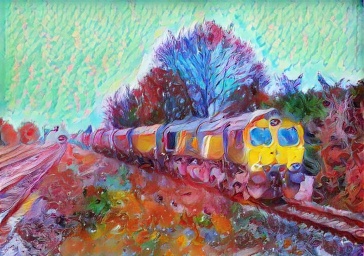} 
    \hspace{-4pt}
  \includegraphics[height=0.1\textwidth,width=0.11\textwidth]{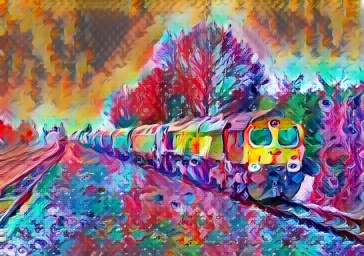}  \\
    \includegraphics[height=0.1\textwidth,width=0.11\textwidth]{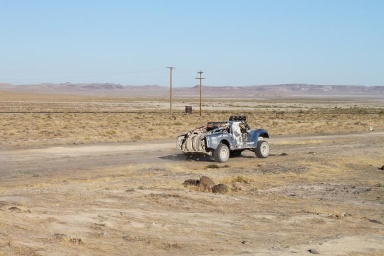}
      \hspace{-4pt} 
    \includegraphics[height=0.1\textwidth,width=0.11\textwidth]{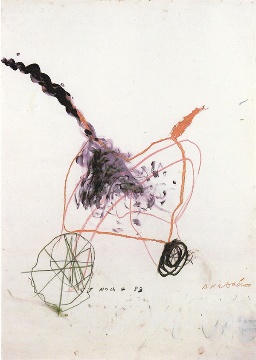}
      \hspace{-4pt} 
    \includegraphics[height=0.1\textwidth,width=0.11\textwidth]{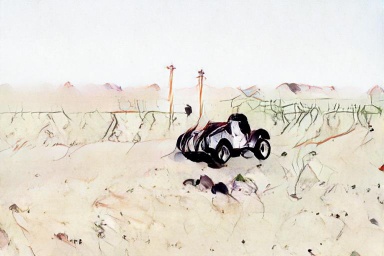}
      \hspace{-4pt} 
     \includegraphics[height=0.1\textwidth,width=0.11\textwidth]{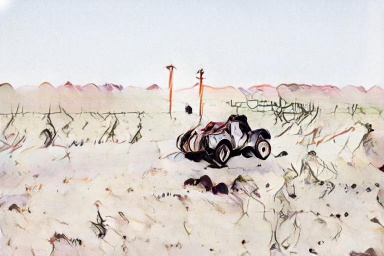}\\ 
\raisebox{0.005\textwidth }{\makebox[0.03\textwidth]{\parbox{0.95\linewidth}{\scriptsize \hspace{ 15pt} Content\hspace{35pt} Style \hspace{ 35pt}  w/ CGM  \hspace{28pt}  w/o CGM}}}     
\vspace{-0.3cm} 
\caption{Effectiveness of the Content-based Gating Module (CGM). The results of AdaAttN are also listed as reference.}
\label{fig:comp_CGM}
\vspace{-0.3cm} 
\end{figure}

\begin{figure}[ht!]
\captionsetup[subfigure]{justification=raggedright, singlelinecheck=false, labelformat=empty}    
\centering
  \includegraphics[height=0.2\textwidth,width=0.3\textwidth]{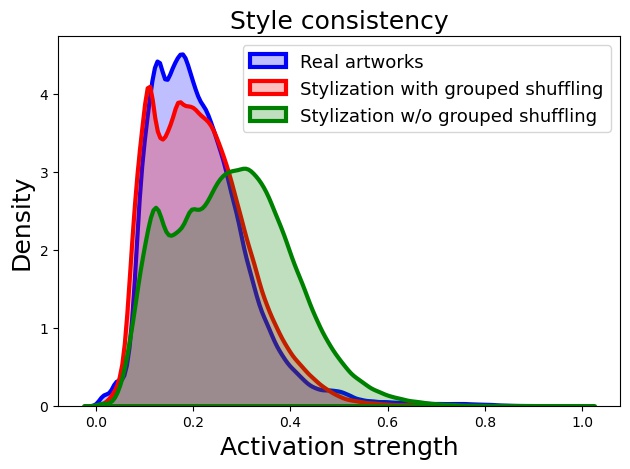}  
\vspace{-0.3cm} 
\caption{Impact of grouped shuffling: we randomly take 100 content-style pairs to create stylizations. From their feature maps extracted by VGG network, we measure the mean activation value channel-wisely and report the overall distribution. }
\label{fig:style_consistency}
\vspace{-0.3cm} 
\end{figure}

\bfsection{Quantitative Evaluation} (1) \textbf{Style Loss.} Following WCT~\cite{li2017universal}, we adopt the style loss to measure the style consistency between the generated stylizations and the style references. The results of different methods are reported in Table~\ref{fig:score}. As we can observe, our method achieves the lowest style loss, which indicates that our method mostly transfers the style information to the output. (2) \textbf{LPIPS.} We conduct LPIPS to measure the stability and
consistency of rendered video clips by following IEC \cite{chen2021artistic}. This metric is to compute the average perceptual distances between consecutive frames and the lower LPIPS score indicates a better stable and consistent performance. We synthesize 10 video clips for each method. In Table \ref{fig:score} we observe that our approach produces the lowest LPIPS score among all methods. 



\bfsection{Efficiency Analysis} In Table \ref{fig:time}, we compare the inference time of different methods on image resolutions of $256\times 256$ and $512 \times 512$. All experiments are conducted using a single Tesla V100 GPU. Our method can achieve 25 fps on $512\times 512$ images, which is comparable with SOTA methods such as AdaAttN and IEC. It worth pointing out that the operation of our style kernel takes FLOPs cost of $H_c \! \times \! W_c \!\times \!  C_{in} \!\times \!(k\!+\! k\! +\! 1)$ while vanilla convolution operation has FLOPS cost of $H_c \! \times \! W_c \!\times \!  C_{out} \!\times \! (C_{in} \times\! k \times\! k +1)$. In Table \ref{tab:add}, we compare the inference time given different kernel sizes. Note we set k=3 as a trade-off between the efficiency and performance.

\subsection{Ablation Study}
We conduct ablation studies to highlight the effectiveness of different modeling components used by our method. 




\bfsection{Effectiveness of Style Kernel} This module works to fully explore the point-wise semantic
structure correlation to modulate local content features for
better artistic style transfer. As shown in Figure \ref{fig:comp_AB}, our dynamic style kernels make
it possible to transfer both the globally aggregated content-
style aligned features to local regions in the content images. This ensures our new scheme “style kernel” works
well on consistent stylization with well preserved structure.

\begin{table}[t!]
 \centering
\setlength{\tabcolsep}{7pt}{
\caption{Results of different sizes of style kernel.}\label{tab:add}
\vspace{-0.3cm}

\begin{tabular}{lcccc} \toprule
Setting & Sty Loss$\downarrow$ & LPIPS $\downarrow$  & $T_{256}$ $\downarrow$  & $T_{512}$$\downarrow$  \\  \midrule
k=1 & 1.04 & 0.30  & 0.015 & 0.034 \\
k=3 & 0.98 & 0.30  & 0.017 & 0.035 \\
k=5 & 0.97 & 0.31  & 0.021 & 0.044 \\
 \bottomrule
\end{tabular}}
\vspace{-0.45cm}	
\end{table}

\begin{figure*}[ht!]
\captionsetup[subfigure]{justification=raggedright, singlelinecheck=false, labelformat=empty}    
\centering
  \includegraphics[height=0.09\textwidth,width=0.14\textwidth]{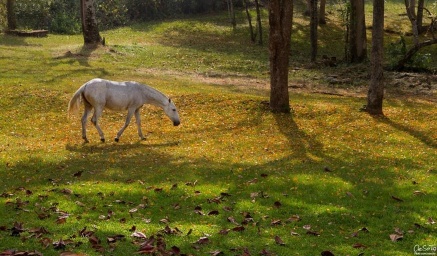}  
 \hspace{-4pt} 
 \includegraphics[height=0.09\textwidth,width=0.14\textwidth]{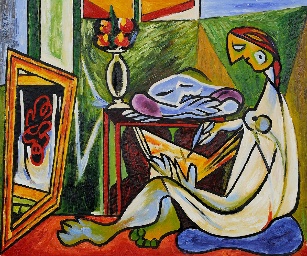}
  \hspace{-4pt}
  \includegraphics[height=0.09\textwidth,width=0.14\textwidth]{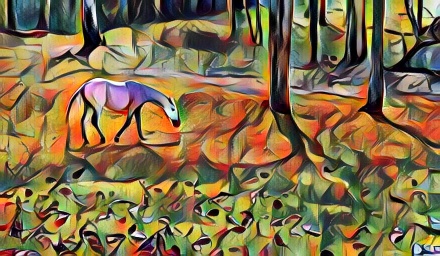} 
    \hspace{-4pt}
  \includegraphics[height=0.09\textwidth,width=0.14\textwidth]{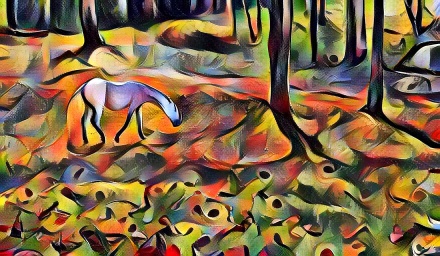}  
  \hspace{-4pt}  
  \includegraphics[height=0.09\textwidth,width=0.14\textwidth]{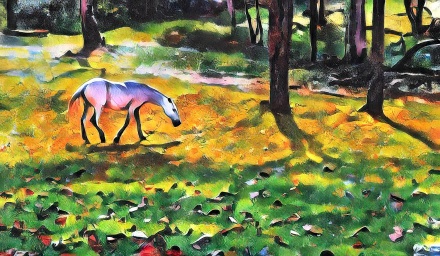} 
    \hspace{-4pt}
  \includegraphics[height=0.09\textwidth,width=0.14\textwidth]{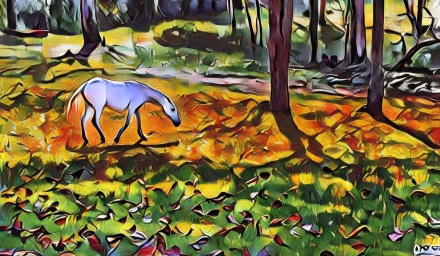}  
  \hspace{-4pt}  
   \includegraphics[height=0.09\textwidth,width=0.14\textwidth]{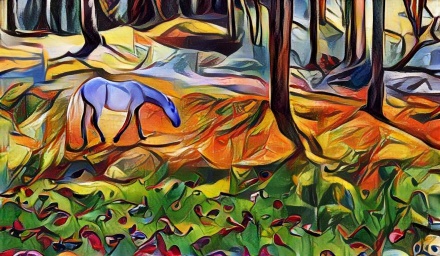}  \\
  \includegraphics[height=0.09\textwidth,width=0.14\textwidth]{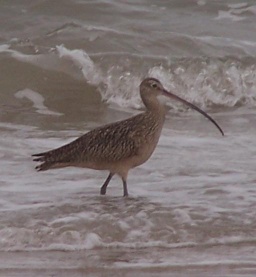}  
 \hspace{-4pt} 
 \includegraphics[height=0.09\textwidth,width=0.14\textwidth]{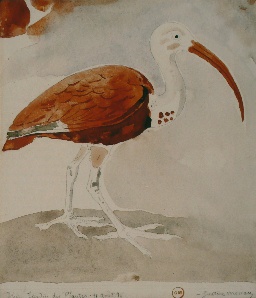}
  \hspace{-4pt}
  \includegraphics[height=0.09\textwidth,width=0.14\textwidth]{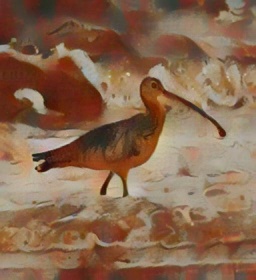} 
    \hspace{-4pt}
  \includegraphics[height=0.09\textwidth,width=0.14\textwidth]{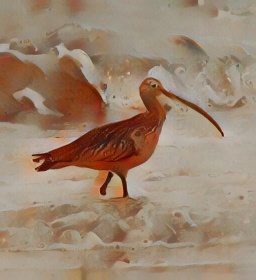}  
  \hspace{-4pt}  
  \includegraphics[height=0.09\textwidth,width=0.14\textwidth]{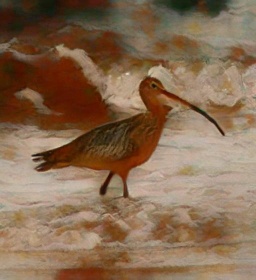} 
    \hspace{-4pt}
  \includegraphics[height=0.09\textwidth,width=0.14\textwidth]{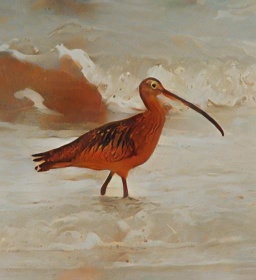}  
  \hspace{-4pt}  
   \includegraphics[height=0.09\textwidth,width=0.14\textwidth]{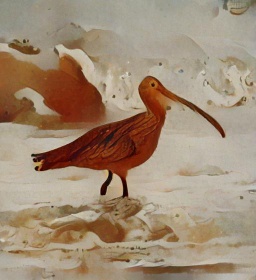}  \\
  \includegraphics[height=0.09\textwidth,width=0.14\textwidth]{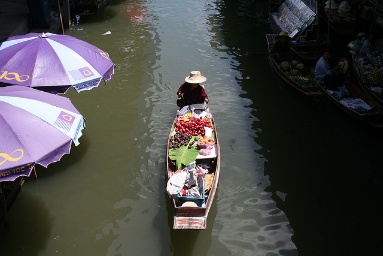}  
 \hspace{-4pt} 
 \includegraphics[height=0.09\textwidth,width=0.14\textwidth]{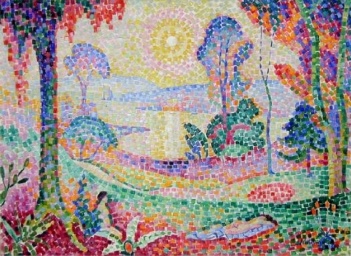}
  \hspace{-4pt}
  \includegraphics[height=0.09\textwidth,width=0.14\textwidth]{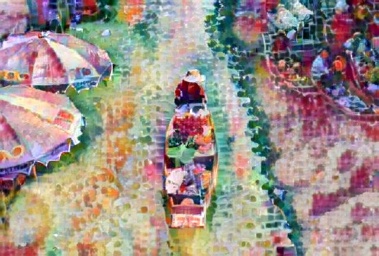} 
    \hspace{-4pt}
  \includegraphics[height=0.09\textwidth,width=0.14\textwidth]{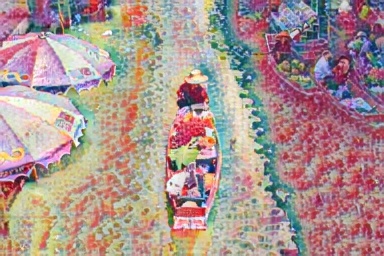}  
  \hspace{-4pt}  
  \includegraphics[height=0.09\textwidth,width=0.14\textwidth]{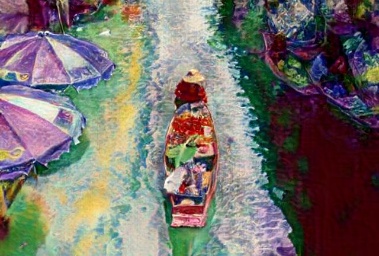} 
    \hspace{-4pt}
  \includegraphics[height=0.09\textwidth,width=0.14\textwidth]{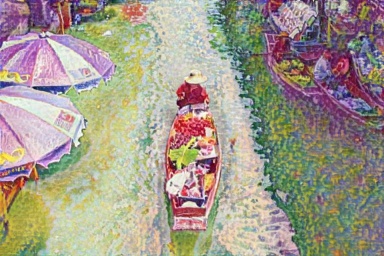}  
  \hspace{-4pt}  
   \includegraphics[height=0.09\textwidth,width=0.14\textwidth]{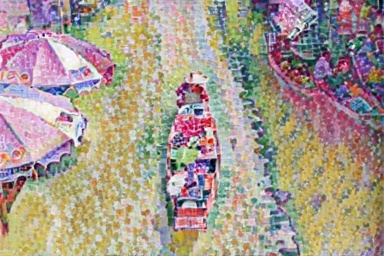}  \\  

\raisebox{0.005\textwidth }{\makebox[0.03\textwidth]{\parbox{0.95\linewidth}{\scriptsize \hspace{ 22pt} Content\hspace{48pt} Style \hspace{ 45pt}  SANet \hspace{38pt}  SANet + CGM \hspace{ 40pt}  IEC \hspace{38pt}  IEC + CGM \hspace{38pt} Ours}}} 
\vspace{-0.3cm} 
\caption{Generalization of our Content-based Gating Module (CGM). By replacing the attention module in SANet and IEC with our CGM, the artifacts and distorted regions in the results of original SANet and IEC are removed and the semantic regions are well preserved.}
\label{fig:plug-in}
\vspace{-0.3cm} 
\end{figure*}

\bfsection{Effectiveness of Grouped Shuffling} To verify this module, we report the statistic distribution of activation strength in Table \ref{fig:style_consistency}. The distribution gap between the results and real artworks is significantly reduced via using grouped shuffling. Without using GS module, the model fails to produce stylizations consistent to the style reference as shown in Figure \ref{fig:comp_AB}. This demonstrates that the GS shrinks the statistic gap between the generated images and the style images.

\bfsection{Effectiveness of CGM} We summarized the results of our model w/ and w/o CGM in Figure \ref{fig:comp_CGM}. The model w/o CGM produces images containing regions with corrupted patterns and the stylization is not consistent to the style reference. While our model w/ CGM faithfully preserves the completeness of semantic regions. This is because our CGM aims to select focusing regions, where query points can aggregate style information from fairly correlated nodes.

\begin{figure}[t!]
\vspace{-0.1cm}
\centering
\captionsetup[subfigure]{justification=raggedright, singlelinecheck=false, labelformat=empty}   
 \centering
    \includegraphics[height=0.1\textwidth, width=0.156\textwidth]{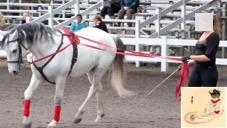}
      \hspace{-4pt}     
    \includegraphics[height=0.1\textwidth, width=0.156\textwidth]{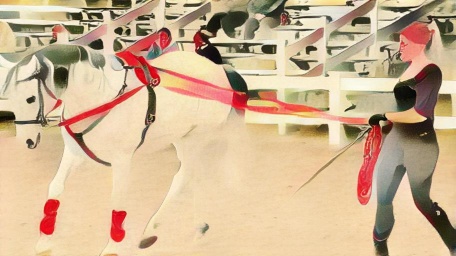}
      \hspace{-4pt} 
    \includegraphics[height=0.1\textwidth, width=0.156\textwidth]{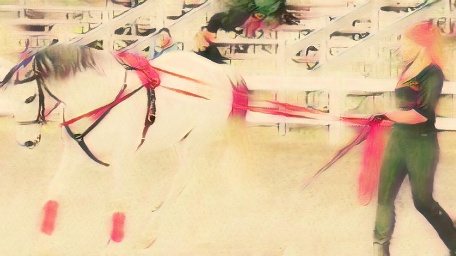}\\
    \includegraphics[height=0.1\textwidth, width=0.156\textwidth]{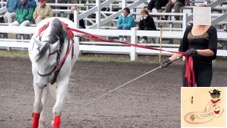}
      \hspace{-4pt} 
    \includegraphics[height=0.1\textwidth, width=0.156\textwidth]{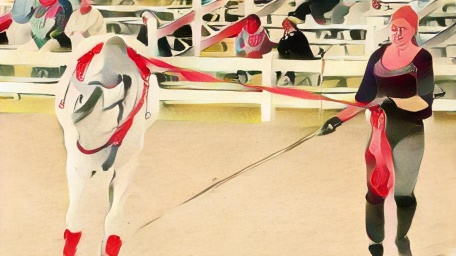}
      \hspace{-4pt} 
    \includegraphics[height=0.1\textwidth, width=0.156\textwidth]{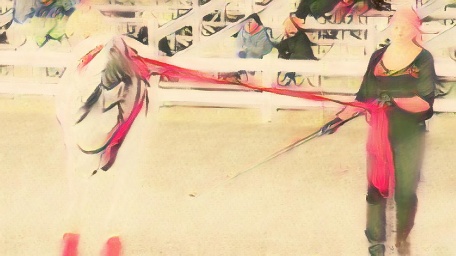}\\ 
 \raisebox{0.005\textwidth }{\makebox[0.03\textwidth]{\parbox{0.95\linewidth}{\scriptsize  \hspace{10pt} Content \& Style\hspace{ 45pt} Ours\hspace{ 55pt} DRB-GAN}}}     
   \vspace{-0.3cm}   
\caption{Comparisons on video style transfer. The 1st column lists 2 frames from a video clip as content images. The style image is at the bottom-right corner.} 
\label{fig:video}
\vspace{-0.5cm} 
\end{figure}

\subsection{More Discussions}
\bfsection{Generalization of CGM} Our content-based gating modulation can be flexibly deployed as a plug-in module. To verify the generalization, we conduct experiments by replacing the attention module in SANet and IEC with our CGM module. The refactored models are denoted as SANet + CGM and IEC + CGM. We train these models with the default settings and compare the performances in Figure \ref{fig:plug-in}. We can observe there are nonneglectable artifacts and distorted regions in the results of original SANet and IEC. By plugging-in our CGM, the artifacts are removed and the semantic regions are well preserved. This further proves the effectiveness of our CGM in aggregating semantically consistent information.  



\begin{figure}[t!]
\vspace{-0.1cm}
\centering
\captionsetup[subfigure]{justification=raggedright, singlelinecheck=false, labelformat=empty} 
 \centering
    \includegraphics[height=0.18\textwidth, width=0.3\textwidth]{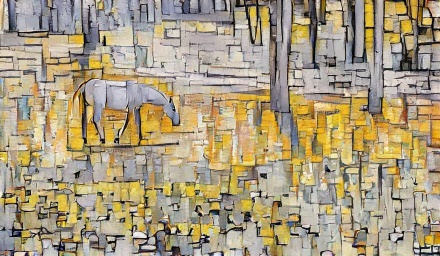}
      \hspace{-2pt}     
    \includegraphics[height=0.18\textwidth, width=0.13\textwidth]{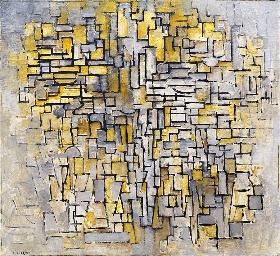}\\
    \includegraphics[height=0.18\textwidth, width=0.3\textwidth]{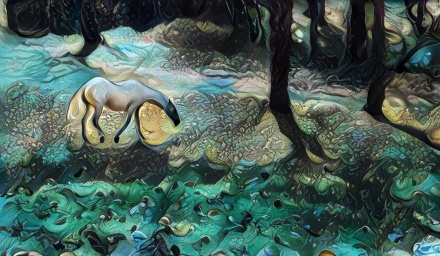}
      \hspace{-2pt} 
    \includegraphics[height=0.18\textwidth, width=0.13\textwidth]{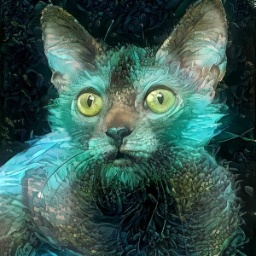}\\
 \raisebox{0.005\textwidth }{\makebox[0.03\textwidth]{\parbox{0.95\linewidth}{\scriptsize  \hspace{ 70pt} Ours \hspace{ 95pt} Style}}}  
   \vspace{-0.3cm}   
\caption{Demonstration of the performance of our method in terms of style consistency and structural similarity.} 
\label{fig:demo2}
\vspace{-0.5cm} 
\end{figure}

\bfsection{Real-world Video Style Transfer} To verify the performance on real-world video stylization, we collect a 1080P video clip consist of 1377 images and compare our method with DRB-GAN \cite{xu2021drb}. As shown in Figure \ref{fig:video}, our approach outperforms existing style transfer methods in terms of style consistency and stability. This can be attributed to the fact that our style kernels are good at transferring the globally aggregated and locally semantic-aligned style features to local regions. 

\bfsection{Social Impact and Future Work} Our model achieves impressive image stylizations with well preserved content structure and consistent style characteristics, as shown in Figure \ref{fig:demo2}. This can definitely benefit our society.  In the future, we will incorporate text information for text conditioned image stylization.


\section{Conclusion}
In this paper, we present a new scheme \textit{``style kernel"} for artistic style transfer. Our method learns an attention map with focusing regions using the proposed content-based gating operator. The style feature is then aligned to match the semantics in the content feature based on the learned attention map. In the style kernel generation module, the dynamic parameters \textit{``style kernel"} are learned from the content-style aligned feature and then applied to modulate the content feature for style transfer. Extensive experimental results demonstrate the remarkable performance of our model in generating synthetic style images with better quality than the state-of-the-art. 

\section*{Acknowledgement}
This research was sponsored by Prof. Yongwei Nie’s Natural Science Foundation of China (62072191).

\clearpage
%
%
{\small
\bibliographystyle{ieee_fullname}
\bibliography{egbib}
}

\clearpage

\appendix
\section{Implementation Details}
{\noindent\bf Configuration of Our Networks.}
We take a pretrained VGG-16 network as the encoder, which is fixed during training. The activation map extracted from ReLU\_4 of the VGG-16 network is utilized as the encoded feature representation. In style-alignment encoding module, we set the number of groups in multi-head attention module $G = 8$. For each attention head, we normalize the query and key features, and take three $3\times3$ convolution blocks to map the query, key and value features, respectively. The $\psi$ comprising of three $3\times3$ convolution blocks maps the content feature $Z_c^{H_c \times W_c \times C}$ into scale parameter $\gamma^G \in \mathbb{R}^{H_c \times W_c \times G}$ and bias parameter $\beta^G \in \mathbb{R}^{H_c \times W_c \times G}$, from which we obtain the $\gamma \in \mathbb{R}^{H_c W_c }$ and $\beta \in \mathbb{R}^{H_c W_c}$ used in each attention head via splitting and reshaping. $Z_{cs}$ is the concatenation of outputs from all attention heads. The $\varphi$ consisting of two $3\times3$ convolution blocks takes in $Z_{cs}$ to predict the filters $F \in \mathbb{R}^{H_c \times W_c \times C(2k + 1)}$. For each spatial point in $Z_{c}$, we take the separable convolution to modulate the content feature with the predicted kernel. The modulated output $\Bar{Z}_{cs}$ is fed into our decoder to generate the stylied image. The structure of our decoder is listed in Table \ref{tab:decoder}. This decoder consists of three residual blocks, upsampling layers, and Convolution layers.

\setlength{\tabcolsep}{3pt}
\begin{table}[ht!]
\centering
\caption{ The network architecture of our decoder. }
\vspace{-0.2cm}
\footnotesize 
\begin{tabular}{l|c|c|c|c|c}
\hlineB{2} 
Model&Block&\multicolumn{2}{c|}{Layer}& Input Shape & Out Shape\\
\hline
\multirow{21}{*}{Decoder} & \multirow{5}{*}{ResBlock}&ReLu &\multirow{2}{*}{ReLu}&[H, W, C]& [H, W, C//2] \\
\cline{3-3} \cline{5-6}
&& Conv &&[H, W, C//2]& [H, W, C//2]\\
\cline{3-6}
&& ReLu &\multirow{2}{*}{Conv}&[H, W, C//2]& [H, W, C//2]\\
\cline{3-3} \cline{5-6}
&& Conv &&[H, W, C//2]& [H, W, C//2]\\
\cline{3-6}
&& \multicolumn{2}{c|}{Sum} &[H, W, C//2]& [H, W, C//2]\\
\cline{2-6}
& Up & \multicolumn{2}{c|}{Upsample}&[H, W, C//2]& [2H, 2W, C//2] \\
\cline{2-6}
& \multirow{5}{*}{ResBlock}&ReLu &\multirow{2}{*}{ReLu}&[2H, 2W, C//2]& [2H, 2W, C//4] \\
\cline{3-3} \cline{5-6}
&& Conv &&[2H, 2W, C//4]& [2H, 2W, C//4]\\
\cline{3-6}
&& ReLu &\multirow{2}{*}{Conv}&[2H, 2W, C//4]& [2H, 2W, C//4]\\
\cline{3-3} \cline{5-6}
&& Conv &&[2H, 2W, C//4]& [2H, 2W, C//4]\\
\cline{3-6}
&& \multicolumn{2}{c|}{Sum} &[2H, 2W, C//4]& [2H, 2W, C//4]\\
\cline{2-6}
& Up & \multicolumn{2}{c|}{Upsample}&[4H, 4W, C//4]& [4H, 4W, C//4] \\
\cline{2-6}
& \multirow{5}{*}{ResBlock}&ReLu &\multirow{2}{*}{ReLu}&[4H, 4W, C//4]& [4H, 4W, C//8] \\
\cline{3-3} \cline{5-6}
&& Conv &&[4H, 4W, C//8]& [4H, 4W, C//8]\\
\cline{3-6}
&& ReLu &\multirow{2}{*}{Conv}&[4H, 4W, C//8]& [4H, 4W, C//8]\\
\cline{3-3} \cline{5-6}
&& Conv &&[4H, 4W, C//8]& [4H, 4W, C//8]\\
\cline{3-6}
&& \multicolumn{2}{c|}{Sum} &[4H, 4W, C//8]& [4H, 4W, C//8]\\
\cline{2-6}
& Up & \multicolumn{2}{c|}{Upsample}&[4H, 4W, C//8]& [8H, 8W, C//8] \\
\cline{2-6}
  & &\multicolumn{2}{c|}{ReLu} &[8H, 8W, C//8]& [8H, 8W, C//8] \\
  \cline{2-6}
  & &\multicolumn{2}{c|}{Conv} &[8H, 8W, C//8]& [8H, 8W, 3] \\
  \cline{2-6}
  & &\multicolumn{2}{c|}{Tanh} &[8H, 8W, 3]& [8H, 8W, 3] \\  
\hlineB{2} 
\end{tabular}
\label{tab:decoder}
\vspace{-0.2cm}
\end{table}

\begin{figure*}[ht!]
\centering
\captionsetup[subfigure]{justification=raggedright, singlelinecheck=false, labelformat=empty}   
\begin{subfigure}{\textwidth}
\centering
\raisebox{0.005\textwidth }{\makebox[0.03\textwidth]{\parbox{0.92\linewidth}{\scriptsize \hspace{ 2.5pt} Content \hspace{ 20pt} Style A \hspace{ 35pt} 1:0 \hspace{ 50pt} 0.75:0.25 \hspace{ 40pt} 0.5:0.5 \hspace{45pt} 0.25:0.75 \hspace{ 45pt} 0:1\hspace{ 43pt} Style B}}}
  \end{subfigure}
    \includegraphics[height=0.11\textwidth,width=0.12\textwidth]{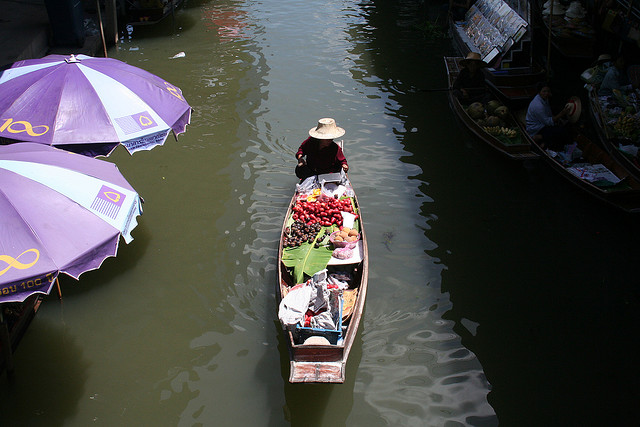}
      \hspace{-4pt} 
    \includegraphics[height=0.11\textwidth]{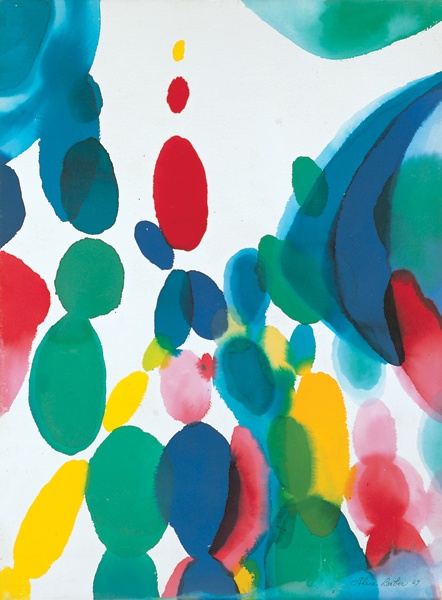}   
      \hspace{-4pt}    
    \includegraphics[height=0.11\textwidth,width=0.136\textwidth]{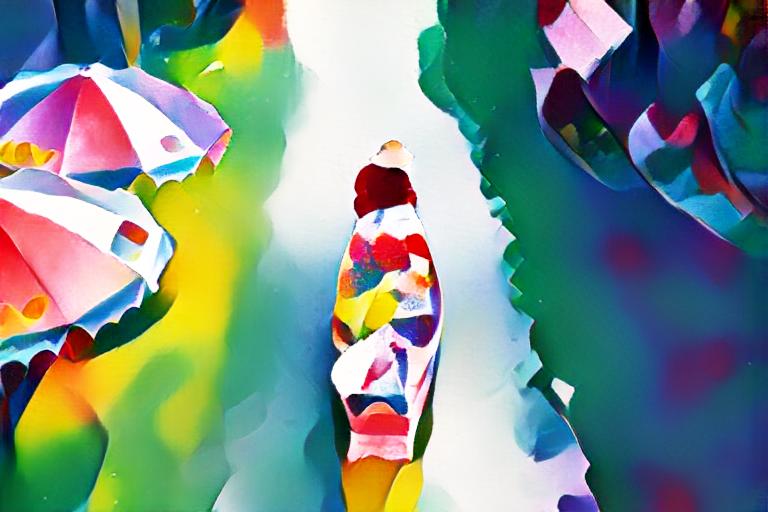}
      \hspace{-4pt} 
    \includegraphics[height=0.11\textwidth,width=0.136\textwidth]{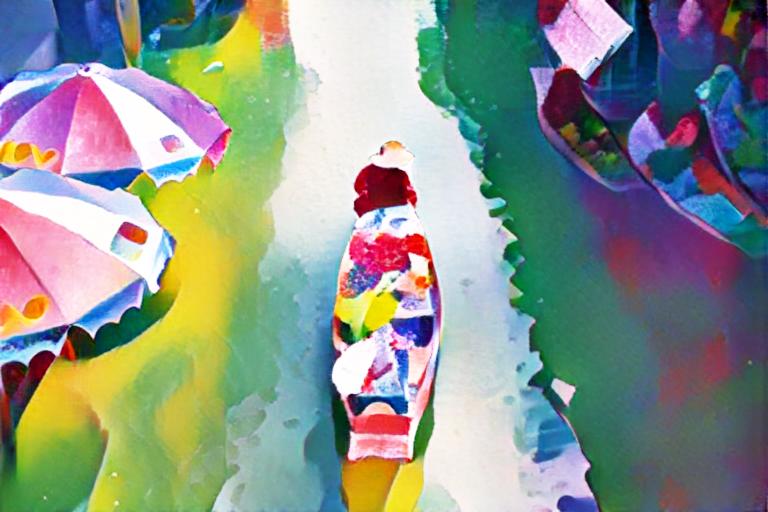}
      \hspace{-4pt} 
    \includegraphics[height=0.11\textwidth,width=0.136\textwidth]{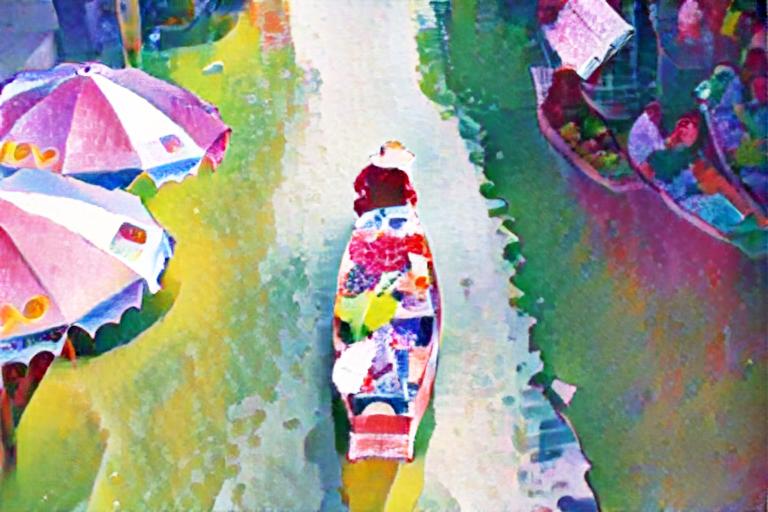}
      \hspace{-4pt}  
    \includegraphics[height=0.11\textwidth,width=0.136\textwidth]{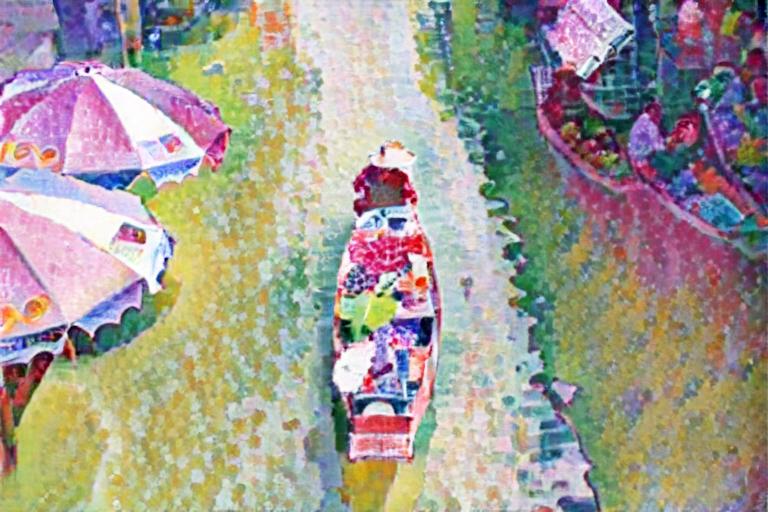}
      \hspace{-4pt} 
    \includegraphics[height=0.11\textwidth,width=0.136\textwidth]{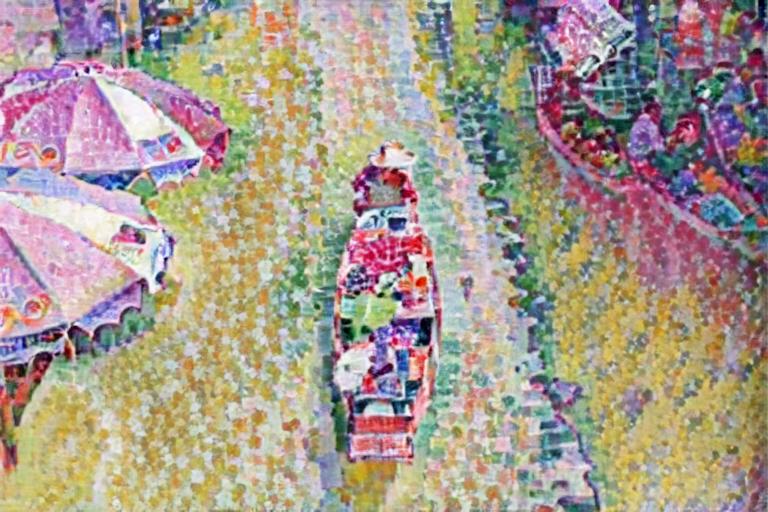}
      \hspace{-4pt} 
    \includegraphics[height=0.11\textwidth,width=0.11\textwidth]{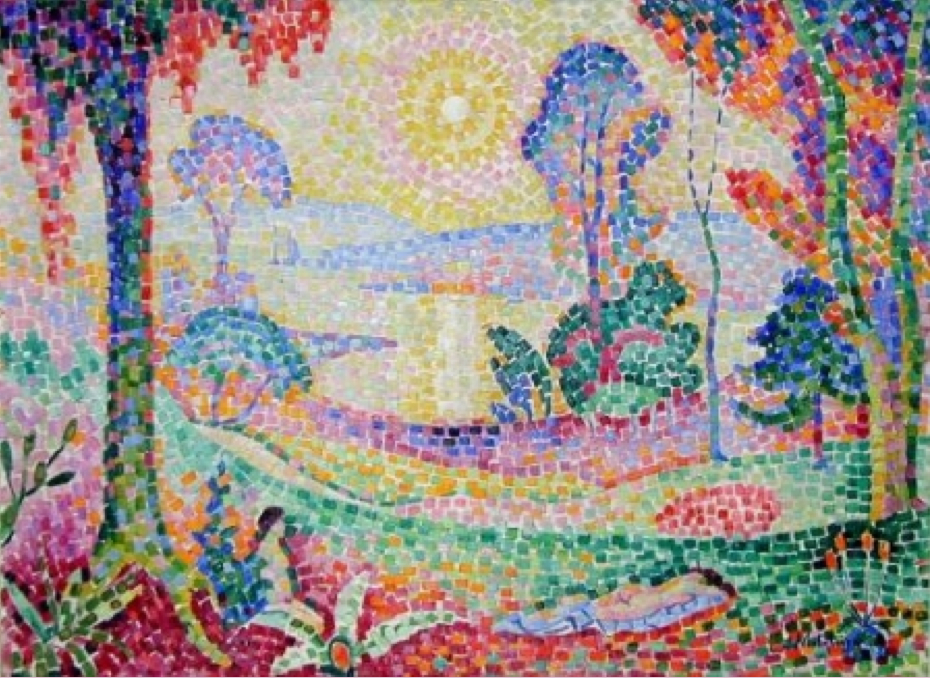}\\
   \vspace{-0.2cm}   
\caption{Performance of style interpolation.}  
\label{fig:inte}
\vspace{-0.35cm} 
\end{figure*}

\begin{figure}[ht!]
\captionsetup[subfigure]{justification=raggedright, singlelinecheck=false, labelformat=empty}  
\centering
 \includegraphics[width=0.48\textwidth]{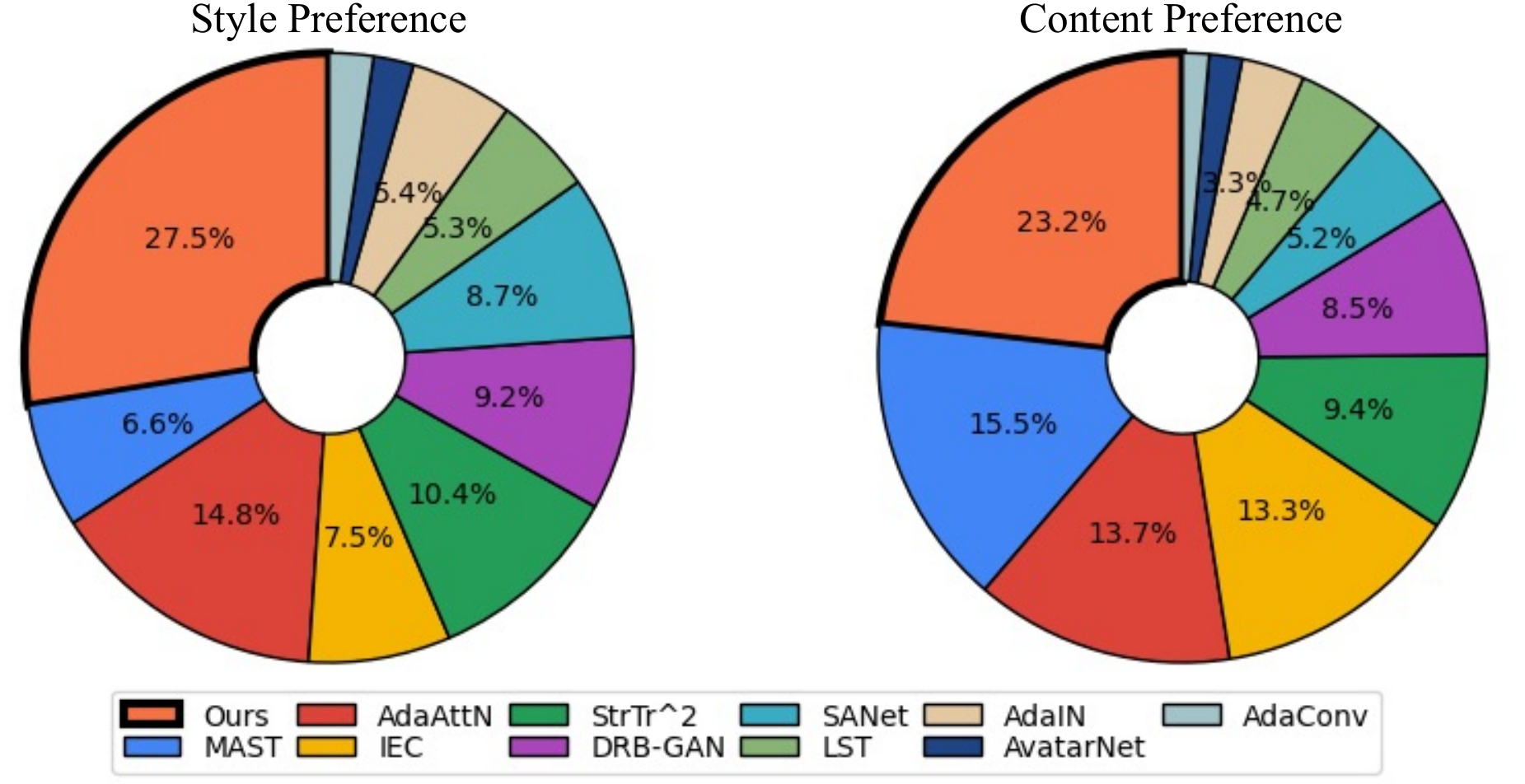}  
   \vspace{-0.4cm}   
\caption{The quantitative comparison of the user study. We report the style preference and content preference of each compared method.} 
\label{fig:User_Study}
\vspace{-0.3cm} 
\end{figure}

\begin{figure}[th!]
\centering
\captionsetup[subfigure]{justification=raggedright, singlelinecheck=false, labelformat=empty}  
\begin{subfigure}{\textwidth}
\centering
\raisebox{0.005\textwidth }{\makebox[0.03\textwidth]{\parbox{0.92\linewidth}{\scriptsize \hspace{ 2.5pt} Content \hspace{ 35pt} Style \hspace{ 40pt} Ours \hspace{ 40pt}  IEC}}}
  \end{subfigure}
\includegraphics[height=0.14\textwidth,width=0.117\textwidth]{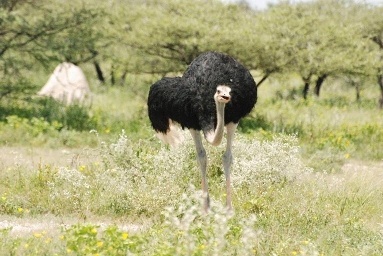}
      \hspace{-3.5pt} 
\includegraphics[height=0.14\textwidth,width=0.117\textwidth]{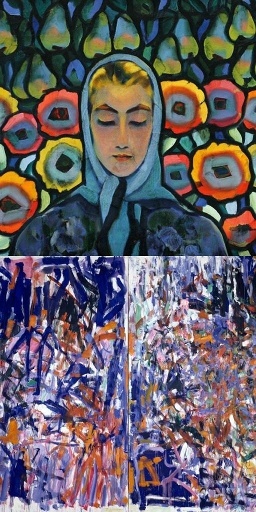}   
      \hspace{-3.5pt}     
  \includegraphics[height=0.14\textwidth,width=0.117\textwidth]{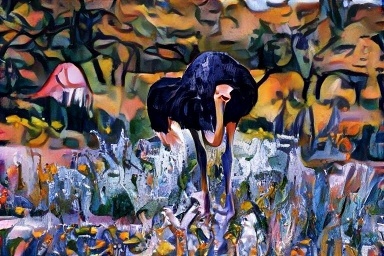} 
   \hspace{-3.5pt}
  \includegraphics[height=0.14\textwidth,width=0.117\textwidth]{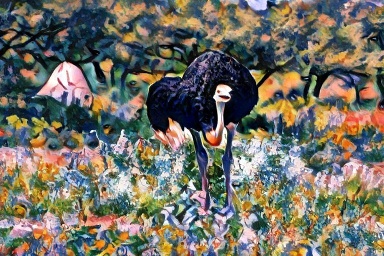} 

\includegraphics[height=0.14\textwidth,width=0.117\textwidth]{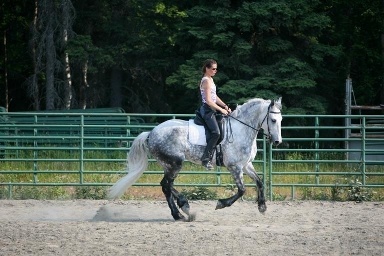}
   \hspace{-3.5pt}
\includegraphics[height=0.14\textwidth,width=0.117\textwidth]{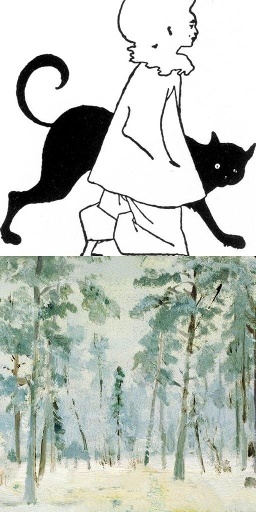}
   \hspace{-3.5pt}
 \includegraphics[height=0.14\textwidth,width=0.117\textwidth]{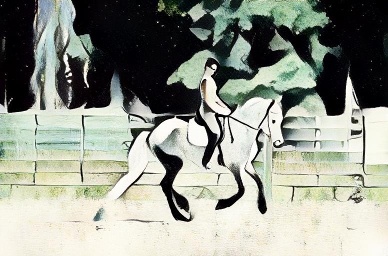}
    \hspace{-3.5pt} 
  \includegraphics[height=0.14\textwidth,width=0.117\textwidth]{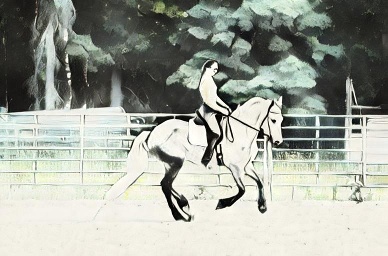} 
  
   \vspace{-0.2cm}   
\caption{Performance of multi-style transfer, where the style images are created by stitching multiple style
images into one image.}  
\label{fig:multi} 
\vspace{-0.10cm} 
\end{figure}

{\noindent\bf Style Transfer with Grouped Shuffling. }
Due to the correlation within each channel-wise feature vector in content feature $Z_c$, directly forcing the output and style images to have similar global statistics (e.g., Gram matrices or covariance matrices) is challenging. As shown in the third and fourth columns in Figure 5 of the manuscript, even though the stylized results are visually close to the style reference, the differences (colors and style patterns) are obvious. To break down the correlation within the feature vector and improve the style consistency between stylized images and style references, we propose a simple but effective method, denoted as grouped shuffling, which is to group the content feature channel-wisely and randomly shuffle the order of groups. 
The grouping operation could be written as:
\begin{align}
\small
\label{eqx}
\begin{split}
Z_c = \overset{O}{ \underset{gs}{||}} Z_c^{gs}
\end{split}
\end{align}
where O refers to the order of groups. $gs \in O$ is the index of each group and $||$ stands for channel-wise concatenation. $O = \{0,1,2,3,4,5,6,7 \} $ and $Z_c^{gs} \in \mathbb{R}^{H_c\times W_c \times C/8}$ given there are 8 groups.  For shuffling, we randomly shuffle the order of groups, such as $O = \{2,3,5,4,6,1,7,0 \} $. The re-concatenated content feature will be fed into the decoder for stylization.

\begin{figure*}[ht!]
\captionsetup[subfigure]{justification=raggedright, singlelinecheck=false, labelformat=empty}  
\centering
 \includegraphics[height=0.12\textwidth,width=0.17\textwidth]{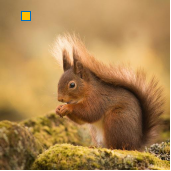}
  \hspace{-4pt}   
 \includegraphics[height=0.12\textwidth,width=0.14\textwidth]{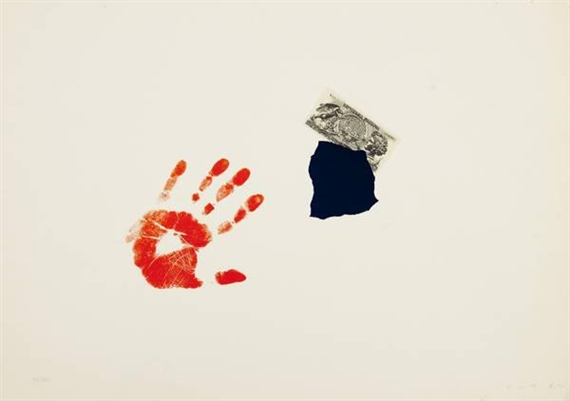}
  \hspace{-4pt}   
  \includegraphics[height=0.12\textwidth,width=0.17\textwidth]{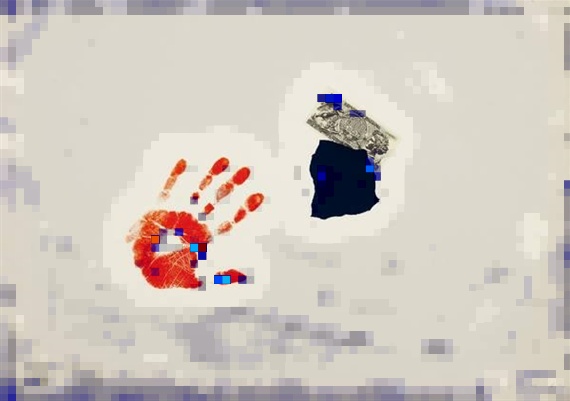}  
  \hspace{-4pt}   
  \includegraphics[height=0.12\textwidth,width=0.17\textwidth]{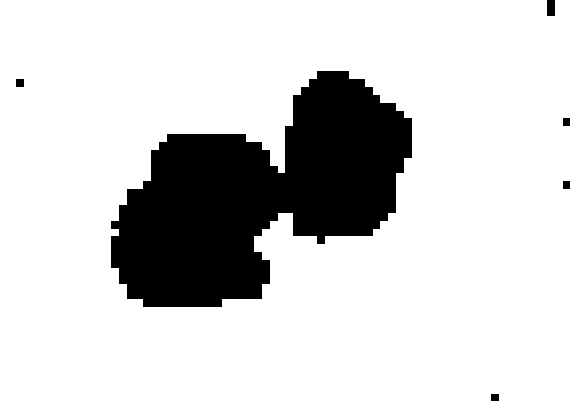}  
  \hspace{-4pt}   
  \includegraphics[height=0.12\textwidth,width=0.17\textwidth]{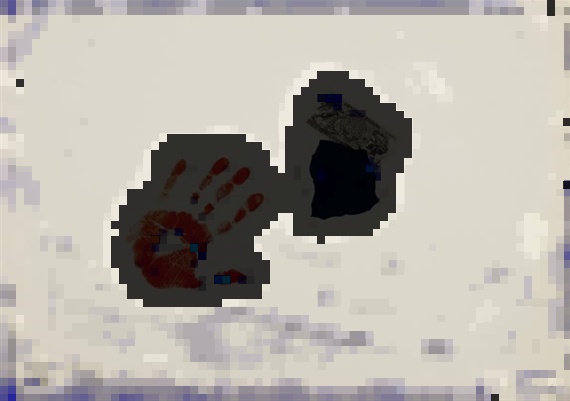}  
  \hspace{-4pt}   
  \includegraphics[height=0.12\textwidth,width=0.17\textwidth]{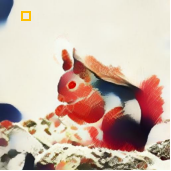}   \\

 \includegraphics[height=0.12\textwidth,width=0.17\textwidth]{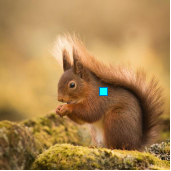}
  \hspace{-4pt}   
 \includegraphics[height=0.12\textwidth,width=0.14\textwidth]{Masked-Attention/1/20}
  \hspace{-4pt}   
  \includegraphics[height=0.12\textwidth,width=0.17\textwidth]{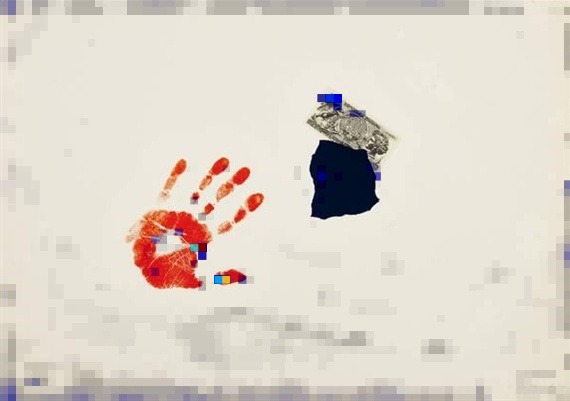}  
  \hspace{-4pt}   
  \includegraphics[height=0.12\textwidth,width=0.17\textwidth]{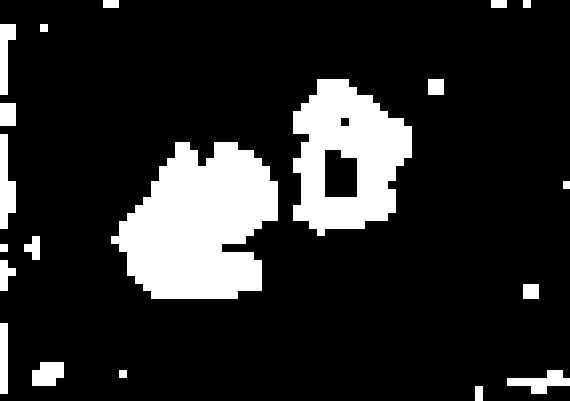}  
  \hspace{-4pt}    
  \includegraphics[height=0.12\textwidth,width=0.17\textwidth]{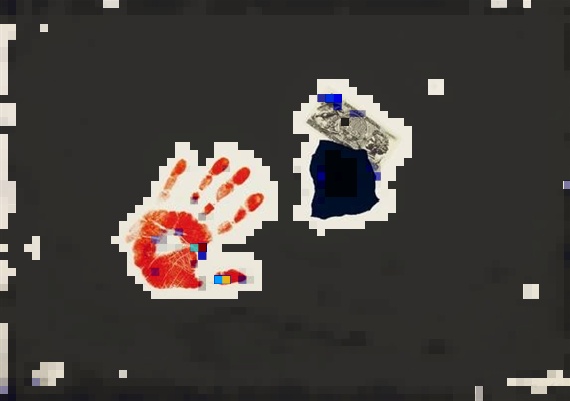}  
  \hspace{-4pt}    
  \includegraphics[height=0.12\textwidth,width=0.17\textwidth]{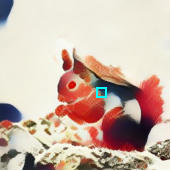}   \\  

  \includegraphics[height=0.12\textwidth,width=0.17\textwidth]{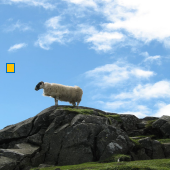}  
  \hspace{-4pt}    
  \includegraphics[height=0.12\textwidth,width=0.14\textwidth]{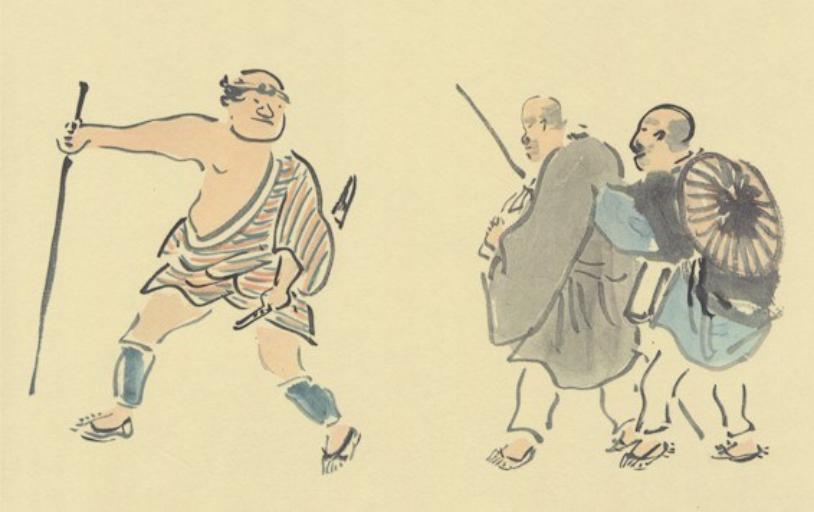}  
  \hspace{-4pt}    
  \includegraphics[height=0.12\textwidth,width=0.17\textwidth]{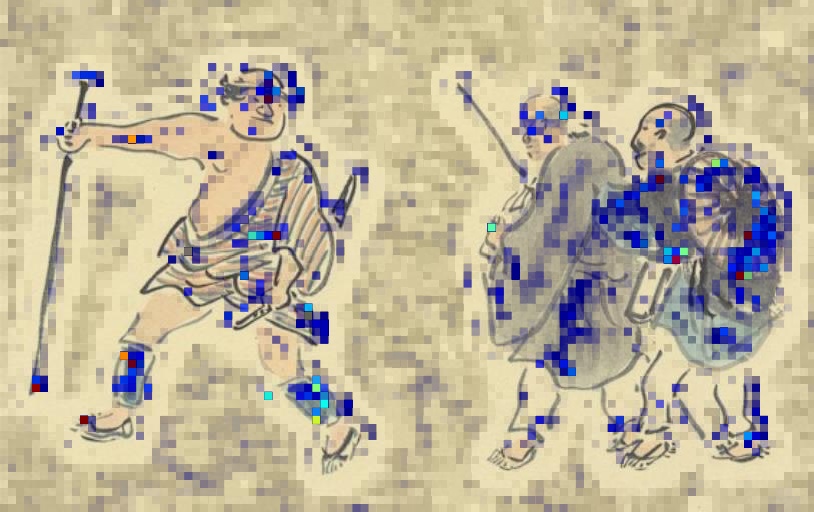}
  \hspace{-4pt}    
  \includegraphics[height=0.12\textwidth,width=0.17\textwidth]{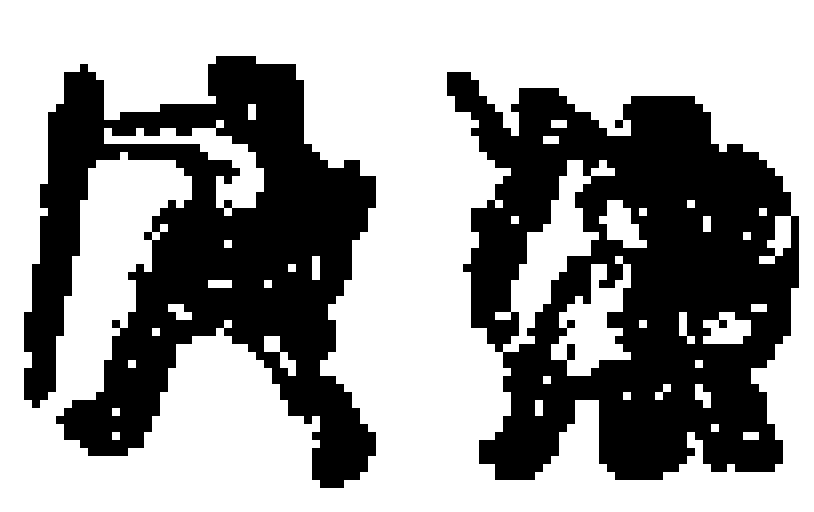}
  \hspace{-4pt}     
  \includegraphics[height=0.12\textwidth,width=0.17\textwidth]{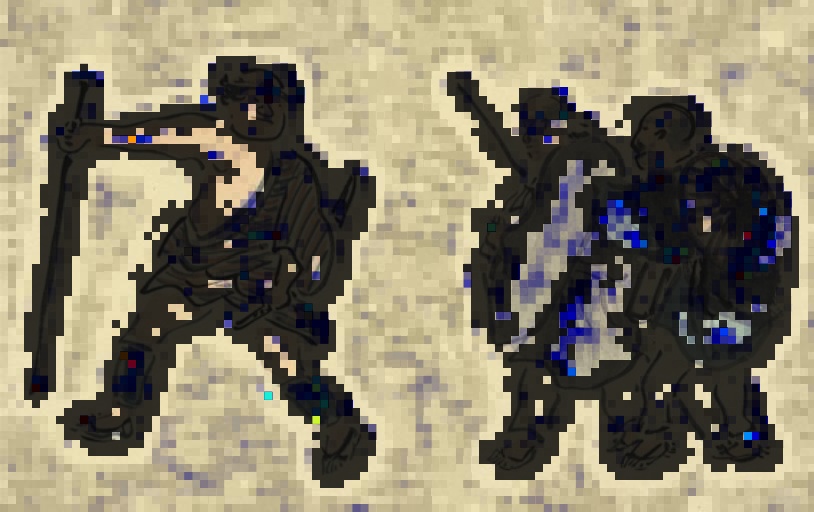}
  \hspace{-4pt}   
    \includegraphics[height=0.12\textwidth,width=0.17\textwidth]{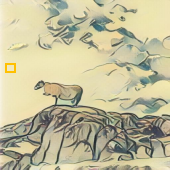}\\

 \includegraphics[height=0.12\textwidth,width=0.17\textwidth]{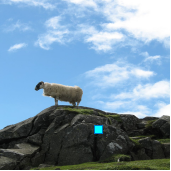}  
  \hspace{-4pt}   
  \includegraphics[height=0.12\textwidth,width=0.14\textwidth]{Masked-Attention/2/2}  
  \hspace{-4pt}   
  \includegraphics[height=0.12\textwidth,width=0.17\textwidth]{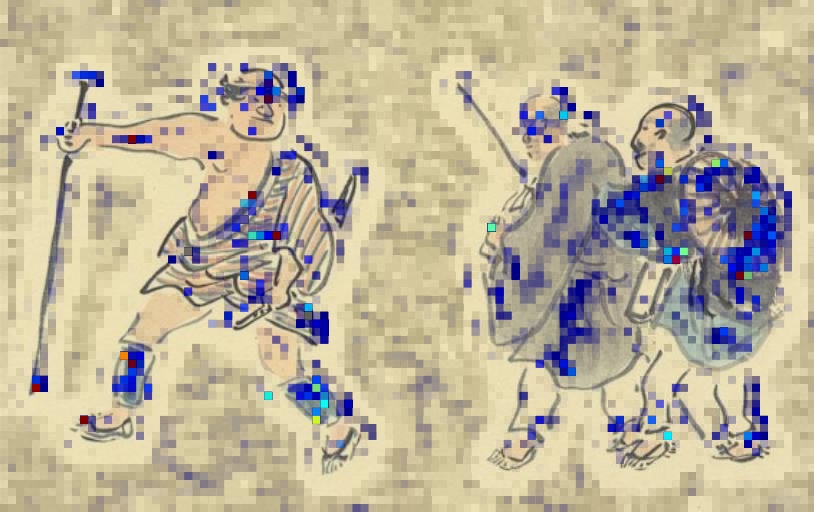}
  \hspace{-4pt}   
  \includegraphics[height=0.12\textwidth,width=0.17\textwidth]{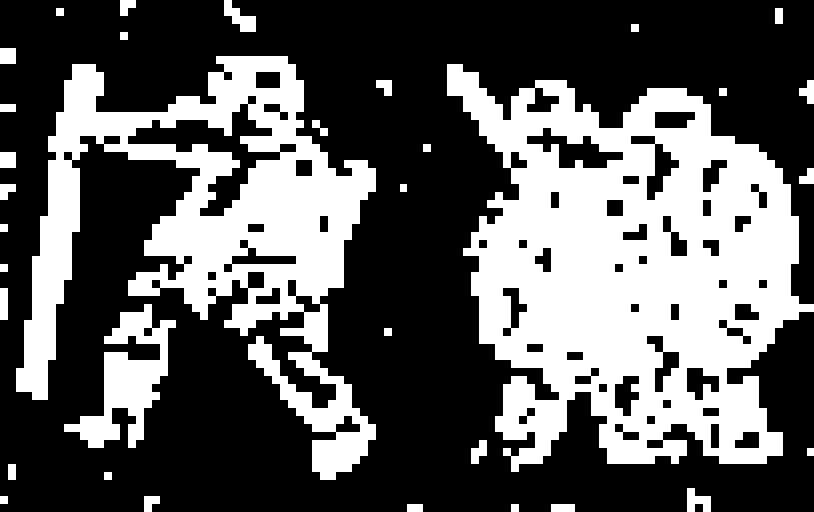}
  \hspace{-4pt}   
  \includegraphics[height=0.12\textwidth,width=0.17\textwidth]{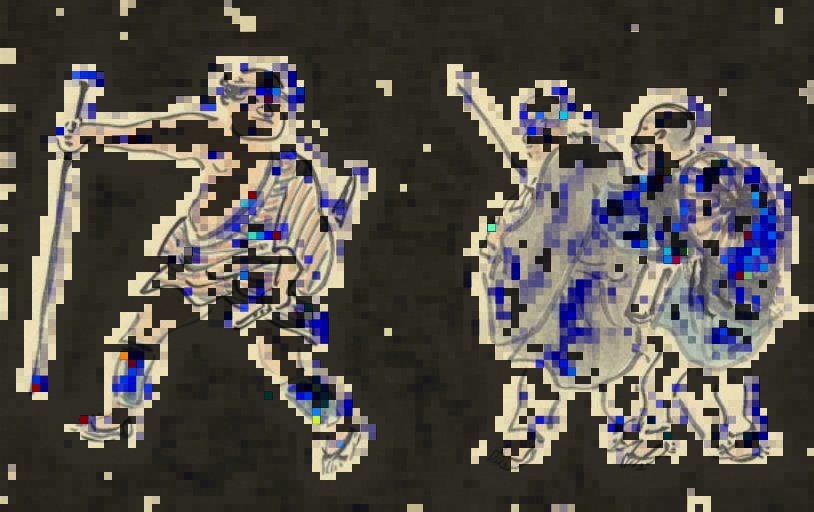}
  \hspace{-4pt}   
    \includegraphics[height=0.12\textwidth,width=0.17\textwidth]{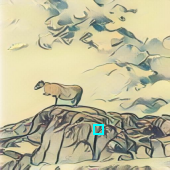}\\

 \includegraphics[height=0.12\textwidth,width=0.17\textwidth]{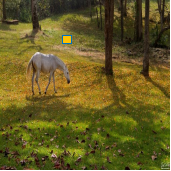}
  \hspace{-4pt}   
 \includegraphics[height=0.12\textwidth,width=0.14\textwidth]{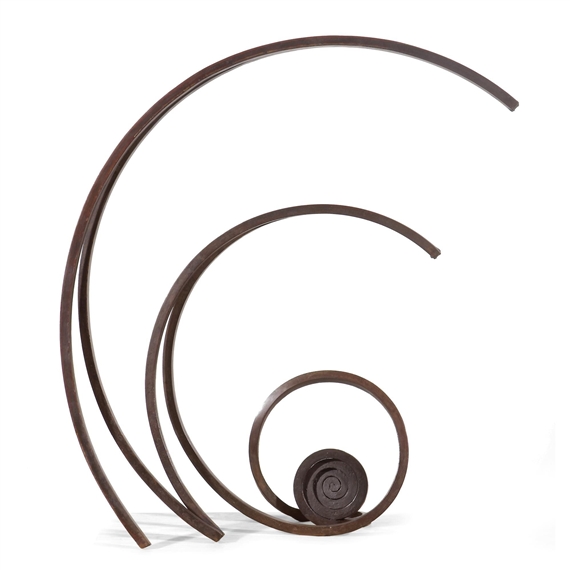}
  \hspace{-4pt}   
    \includegraphics[height=0.12\textwidth,width=0.17\textwidth]{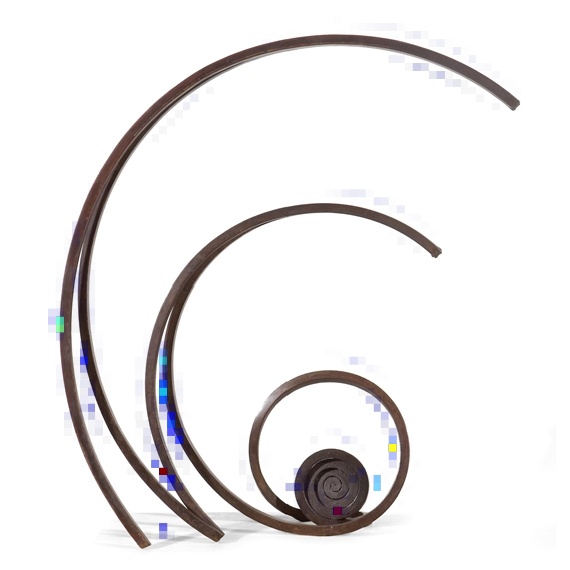}  
  \hspace{-4pt}  
  \includegraphics[height=0.12\textwidth,width=0.17\textwidth]{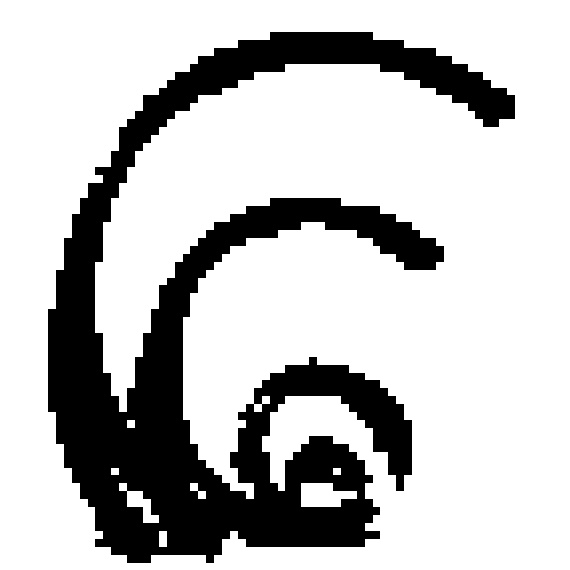}  
  \hspace{-4pt}   
  \includegraphics[height=0.12\textwidth,width=0.17\textwidth]{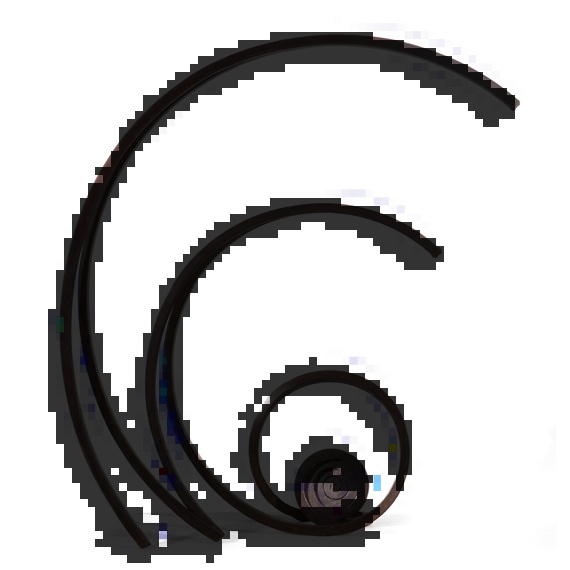}  
  \hspace{-4pt}   
    \includegraphics[height=0.12\textwidth,width=0.17\textwidth]{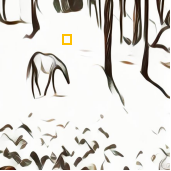}  \\
  
 \includegraphics[height=0.12\textwidth,width=0.17\textwidth]{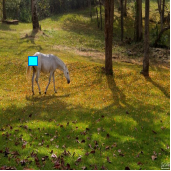}
  \hspace{-4pt}  
 \includegraphics[height=0.12\textwidth,width=0.14\textwidth]{Masked-Attention/3/2}
  \hspace{-4pt}  
    \includegraphics[height=0.12\textwidth,width=0.17\textwidth]{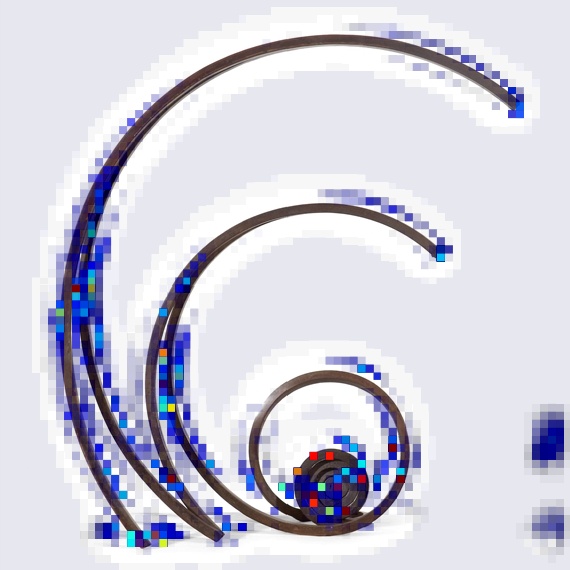}  
  \hspace{-4pt}   
  \includegraphics[height=0.12\textwidth,width=0.17\textwidth]{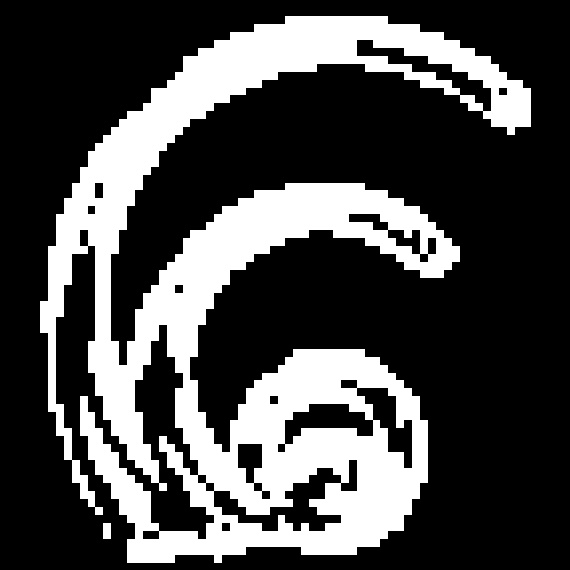}  
  \hspace{-4pt} 
  \includegraphics[height=0.12\textwidth,width=0.17\textwidth]{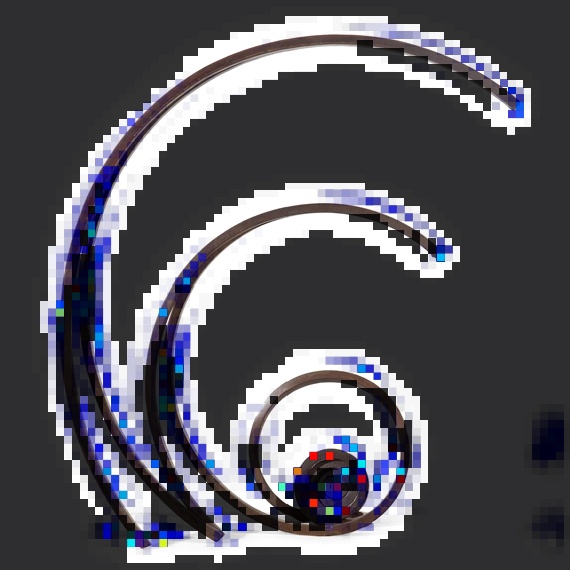}  
  \hspace{-4pt}  
    \includegraphics[height=0.12\textwidth,width=0.17\textwidth]{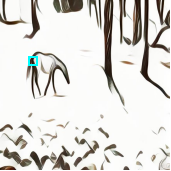}  \\

 \raisebox{0.005\textwidth }{\makebox[0.03\textwidth]{\parbox{0.95\linewidth}{\scriptsize \hspace{ 18pt} Content\hspace{55pt} Style \hspace{ 40pt} Attention Map $\mathcal{A}$  \hspace{40pt} Focused Region  \hspace{ 22pt} Masked Attention Map $\bar{\mathcal{A}}$ \hspace{38pt} Ours }}}
 
\vspace{-0.2cm}
\caption{ Visualizations of the focused region. Given a query point in the content image (yellow and Cyan rectangles), the corresponding focused region is learned to filter out points with low correlation. The focused region contains a complete semantic area, enabling our method to create style-consistent stylizations with well-preserved content structure. Please zoom in for better visualization. The attention distributed over the image is highlighted in colors.}  
\label{fig:focusing_region} 
\vspace{-0.3cm} 
\end{figure*}

\begin{figure*}[ht!]
\centering
\raisebox{0.02\textwidth }{\makebox[0.09\textwidth]{\footnotesize Style \begin{turn}{90} \raisebox{0.00002\textwidth }{ \footnotesize Content}\end{turn}}}
\hspace{-4pt}  
  \includegraphics[height=0.12\textwidth,width=0.14\textwidth]{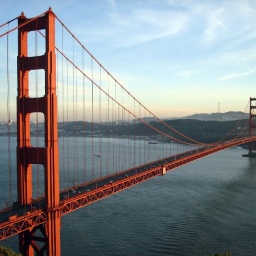}  
 \hspace{-4pt} 
 \includegraphics[height=0.12\textwidth,width=0.14\textwidth]{Supp/SOTAComp/cc/3}
    \hspace{-4pt}
  \includegraphics[height=0.12\textwidth,width=0.1\textwidth]{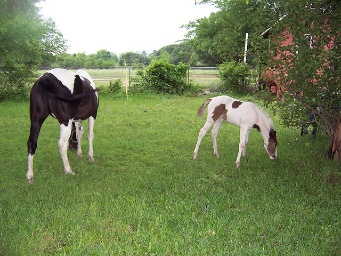}  
  \hspace{-4pt}
  \includegraphics[height=0.12\textwidth,width=0.13\textwidth]{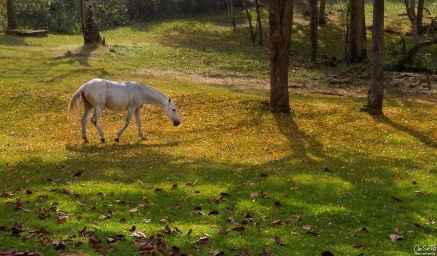}  
  \hspace{-4pt}  
  \includegraphics[height=0.12\textwidth,width=0.14\textwidth]{Supp/SOTAComp/cc/17}  
  \hspace{-4pt}
 \includegraphics[height=0.12\textwidth,width=0.14\textwidth]{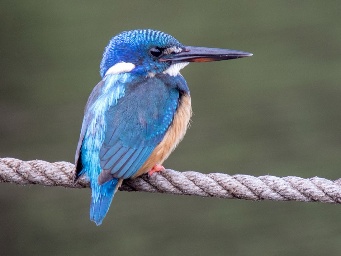}
  \hspace{-4pt}
  \includegraphics[height=0.12\textwidth,width=0.11\textwidth]{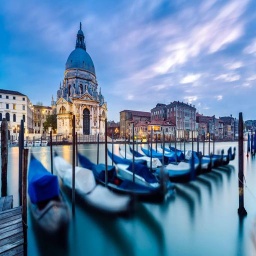} \\
  \includegraphics[height=0.12\textwidth,width=0.09\textwidth]{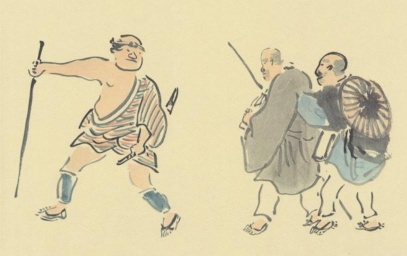}
 \hspace{-4pt} 
  \includegraphics[height=0.12\textwidth,width=0.14\textwidth]{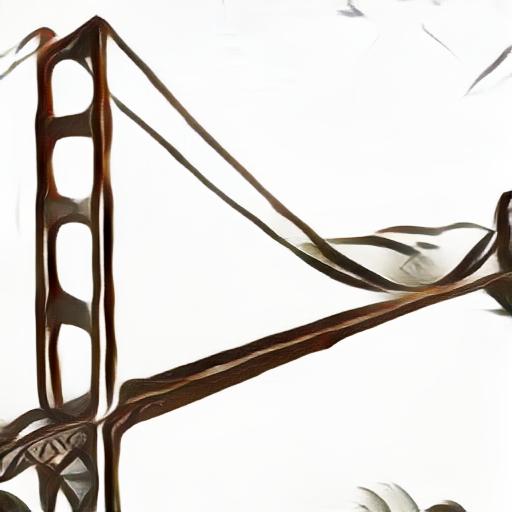}
 \hspace{-4pt} 
 \includegraphics[height=0.12\textwidth,width=0.14\textwidth]{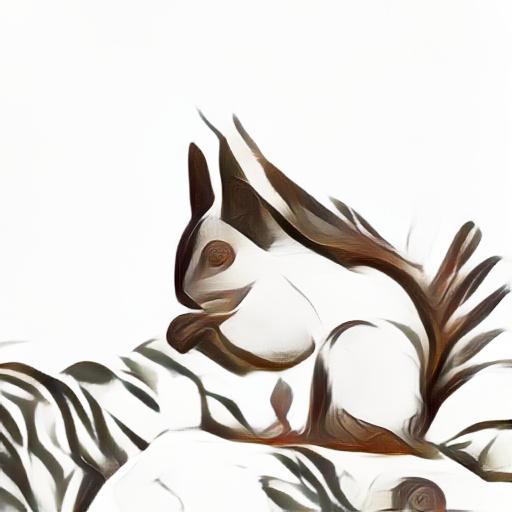}
  \hspace{-4pt}
  \includegraphics[height=0.12\textwidth,width=0.1\textwidth]{Supp/SOTAComp/ours/8_1}  
  \hspace{-4pt}
  \includegraphics[height=0.12\textwidth,width=0.13\textwidth]{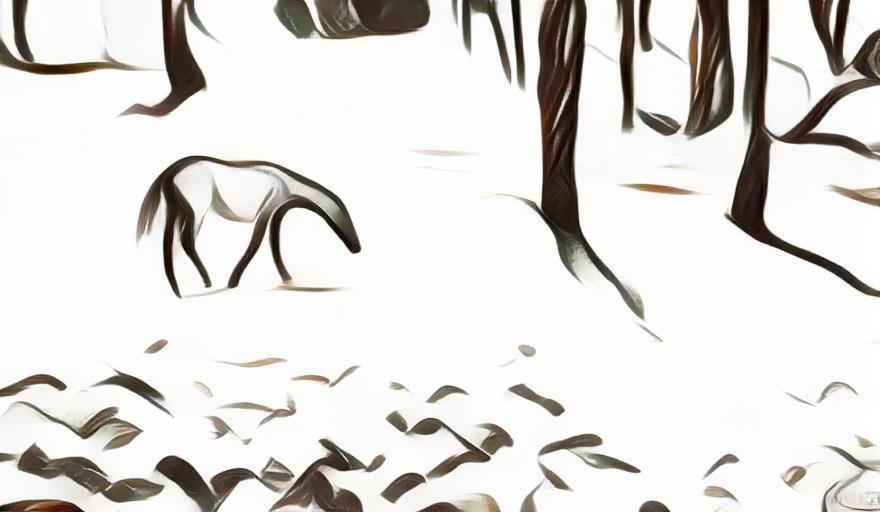}  
  \hspace{-4pt}  
  \includegraphics[height=0.12\textwidth,width=0.14\textwidth]{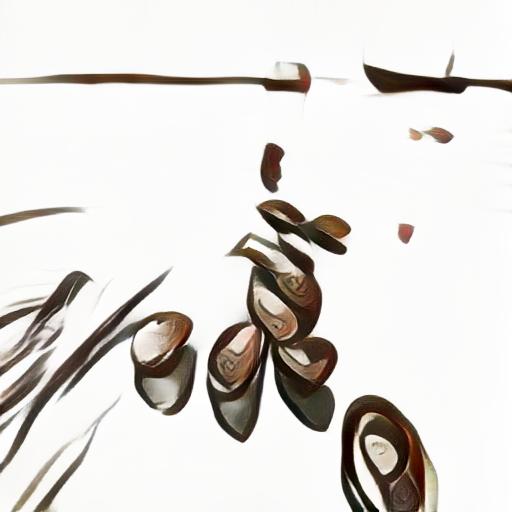}  
  \hspace{-4pt}
 \includegraphics[height=0.12\textwidth,width=0.14\textwidth]{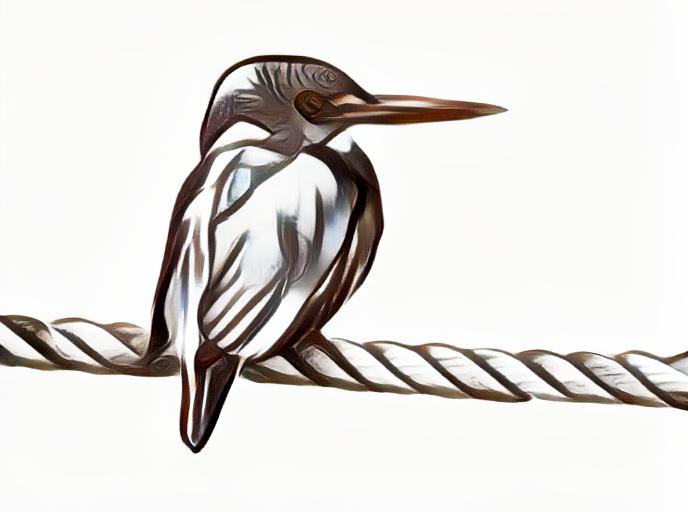}
  \hspace{-4pt}
  \includegraphics[height=0.12\textwidth,width=0.11\textwidth]{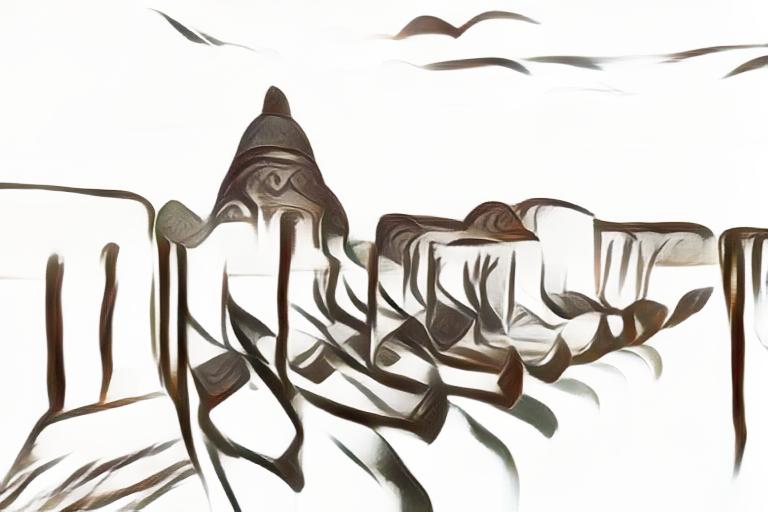} \\
  \includegraphics[height=0.12\textwidth,width=0.09\textwidth]{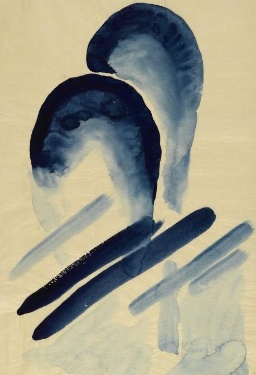}
 \hspace{-4pt} 
  \includegraphics[height=0.12\textwidth,width=0.14\textwidth]{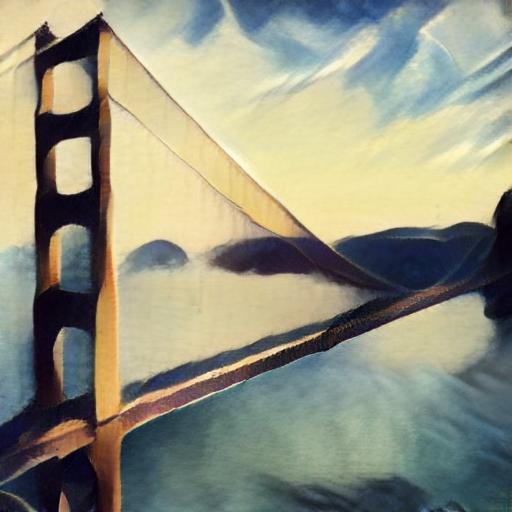}
 \hspace{-4pt} 
 \includegraphics[height=0.12\textwidth,width=0.14\textwidth]{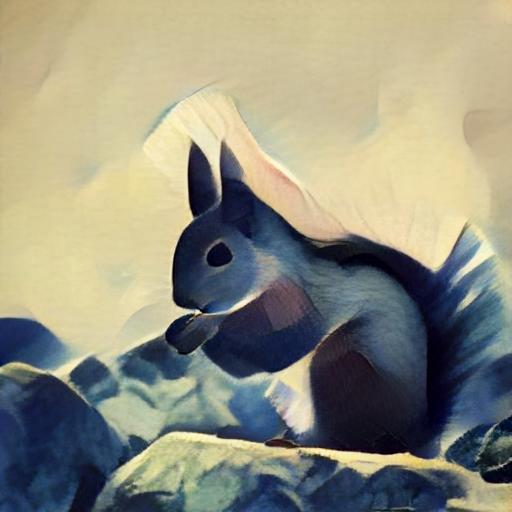}
  \hspace{-4pt}
  \includegraphics[height=0.12\textwidth,width=0.1\textwidth]{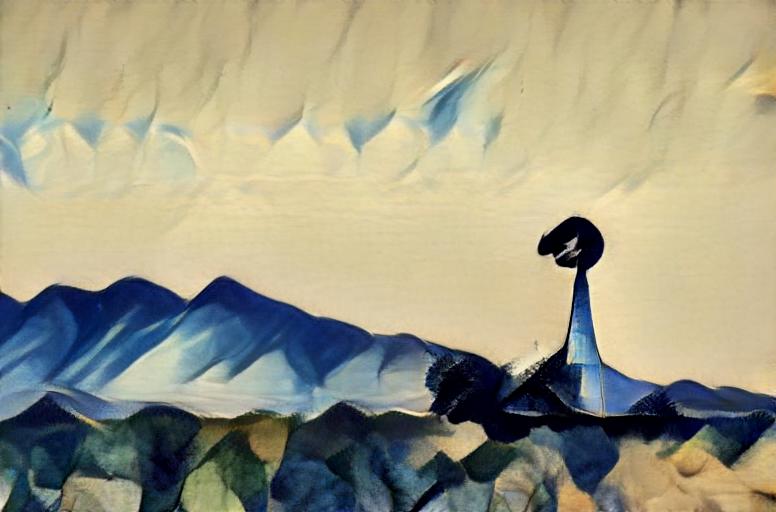}  
  \hspace{-4pt}
  \includegraphics[height=0.12\textwidth,width=0.13\textwidth]{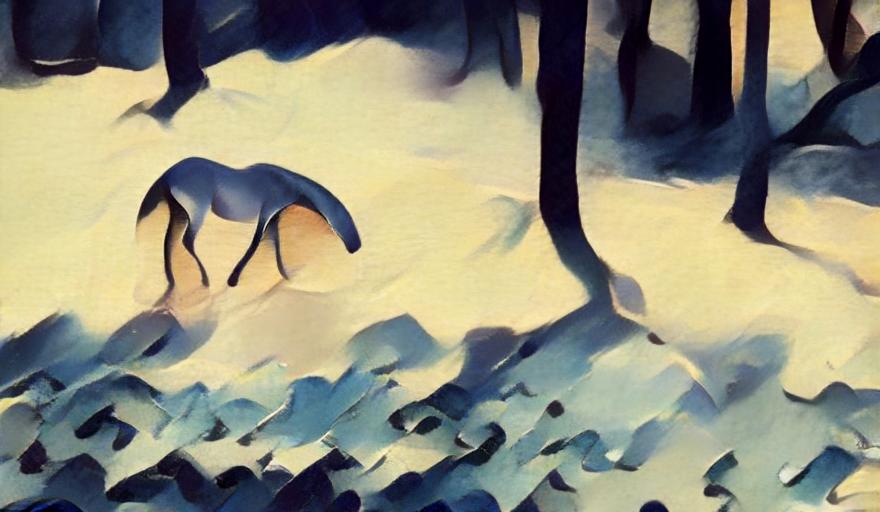}  
  \hspace{-4pt}  
  \includegraphics[height=0.12\textwidth,width=0.14\textwidth]{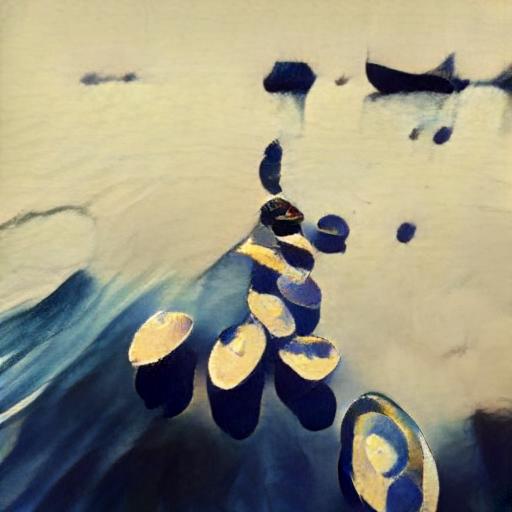}  
  \hspace{-4pt}
 \includegraphics[height=0.12\textwidth,width=0.14\textwidth]{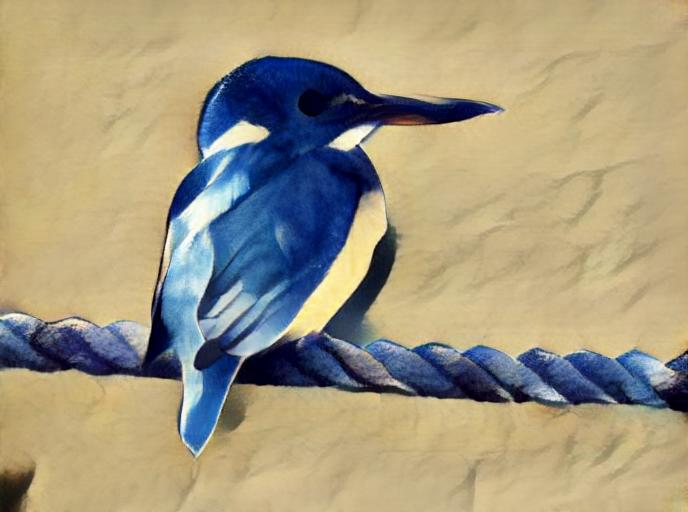}
  \hspace{-4pt}
  \includegraphics[height=0.12\textwidth,width=0.11\textwidth]{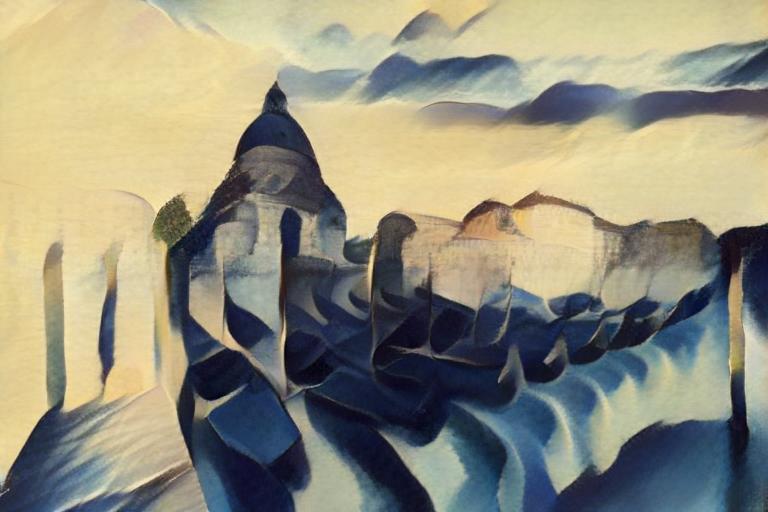} \\
  \includegraphics[height=0.12\textwidth,width=0.09\textwidth]{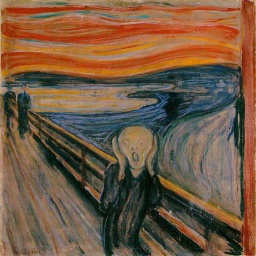}
 \hspace{-4pt} 
  \includegraphics[height=0.12\textwidth,width=0.14\textwidth]{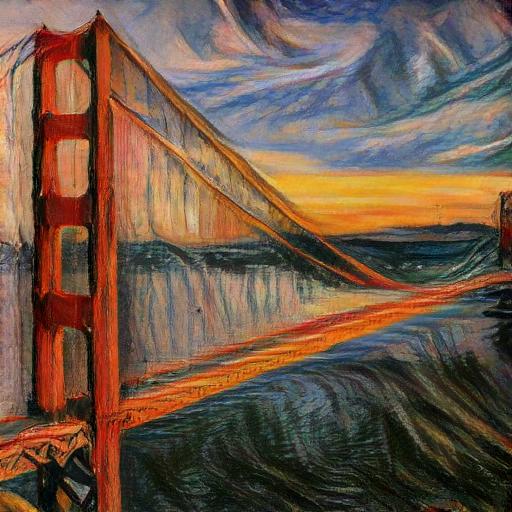}
 \hspace{-4pt} 
 \includegraphics[height=0.12\textwidth,width=0.14\textwidth]{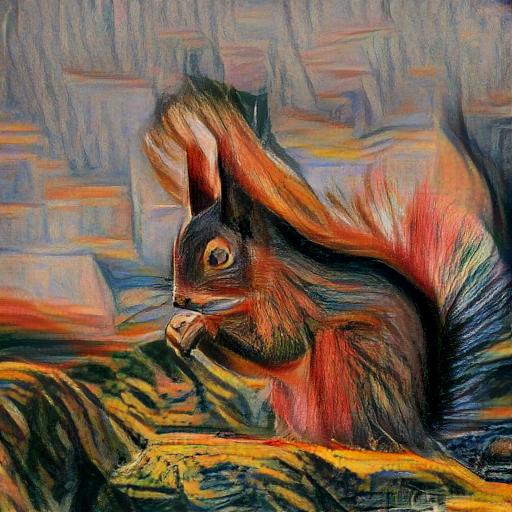}
  \hspace{-4pt}
  \includegraphics[height=0.12\textwidth,width=0.1\textwidth]{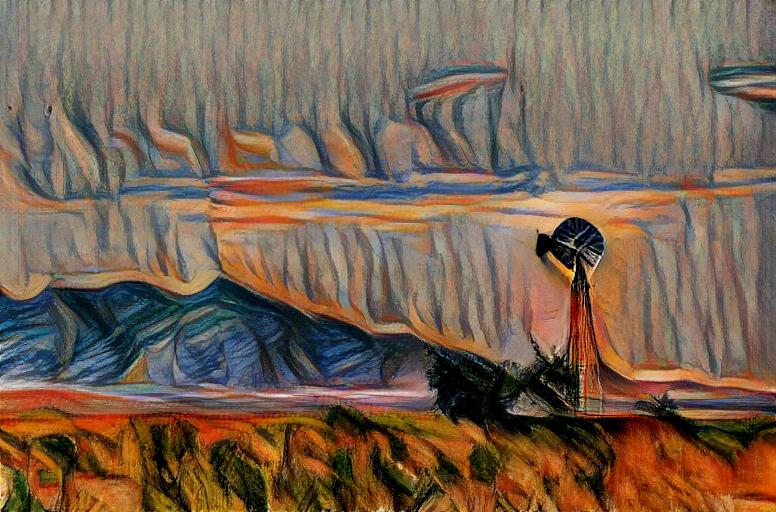}  
  \hspace{-4pt}
  \includegraphics[height=0.12\textwidth,width=0.13\textwidth]{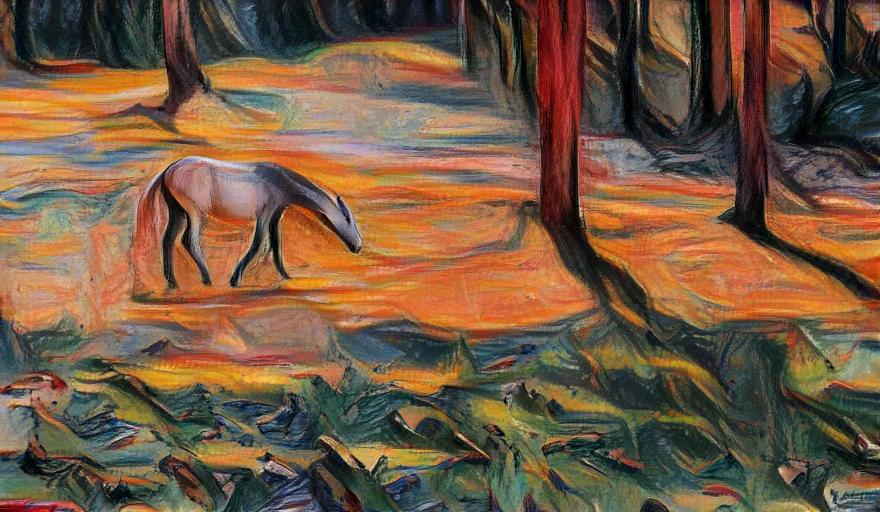}  
  \hspace{-4pt}  
  \includegraphics[height=0.12\textwidth,width=0.14\textwidth]{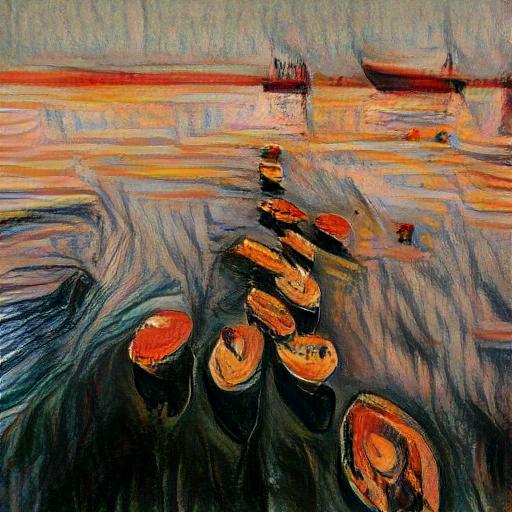}  
  \hspace{-4pt}
 \includegraphics[height=0.12\textwidth,width=0.14\textwidth]{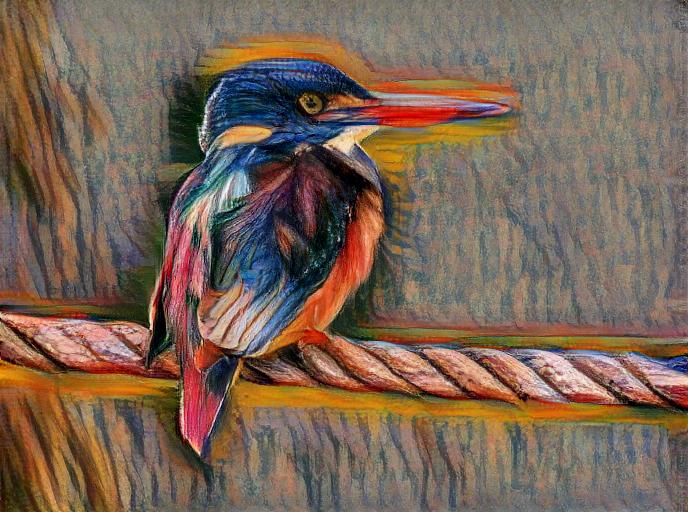}
  \hspace{-4pt}
  \includegraphics[height=0.12\textwidth,width=0.11\textwidth]{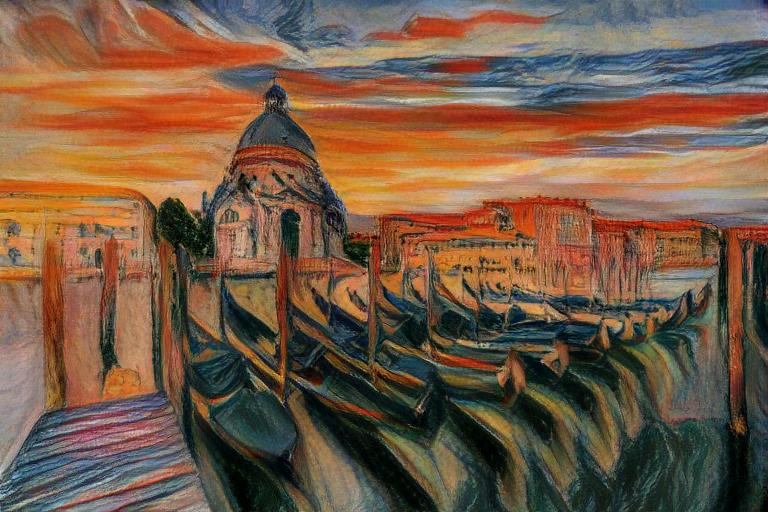} \\
  \includegraphics[height=0.12\textwidth,width=0.09\textwidth]{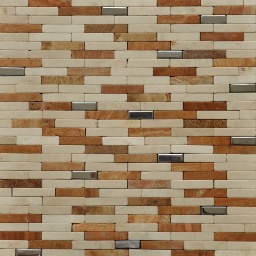}
 \hspace{-4pt} 
  \includegraphics[height=0.12\textwidth,width=0.14\textwidth]{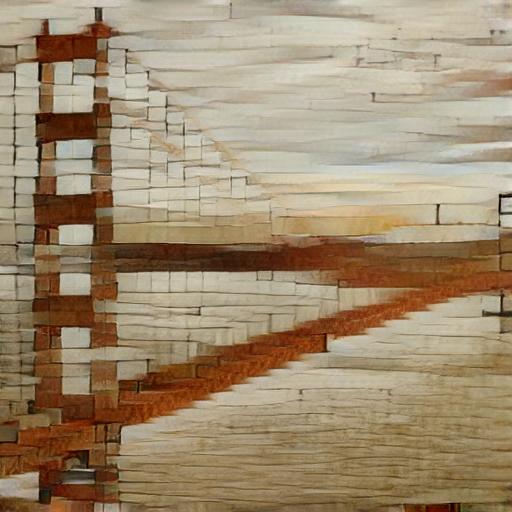}
 \hspace{-4pt} 
 \includegraphics[height=0.12\textwidth,width=0.14\textwidth]{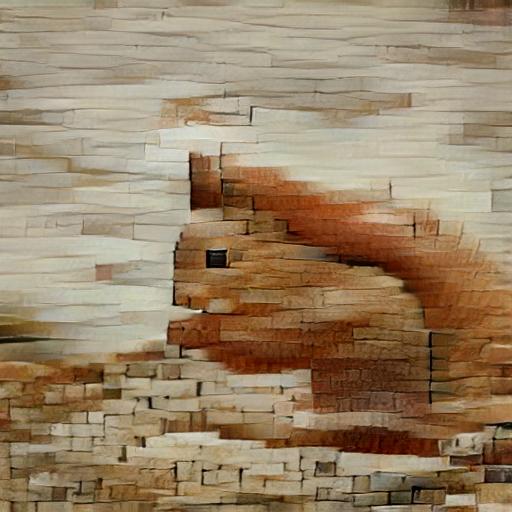}
  \hspace{-4pt}
  \includegraphics[height=0.12\textwidth,width=0.1\textwidth]{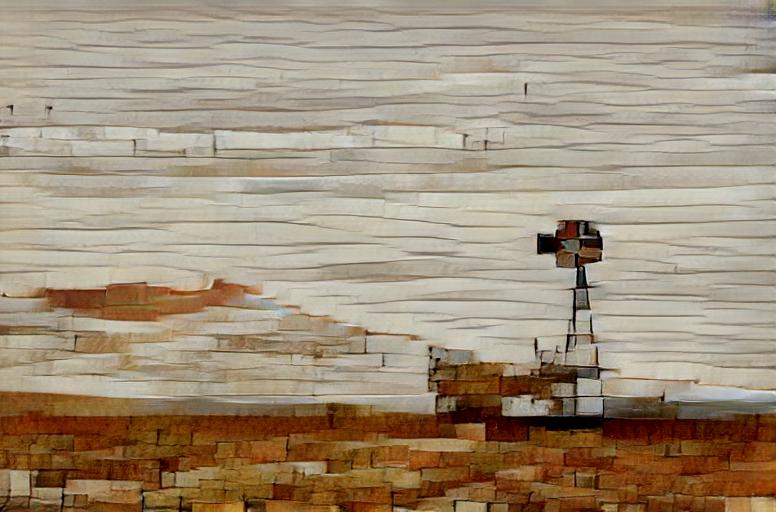}  
  \hspace{-4pt}
  \includegraphics[height=0.12\textwidth,width=0.13\textwidth]{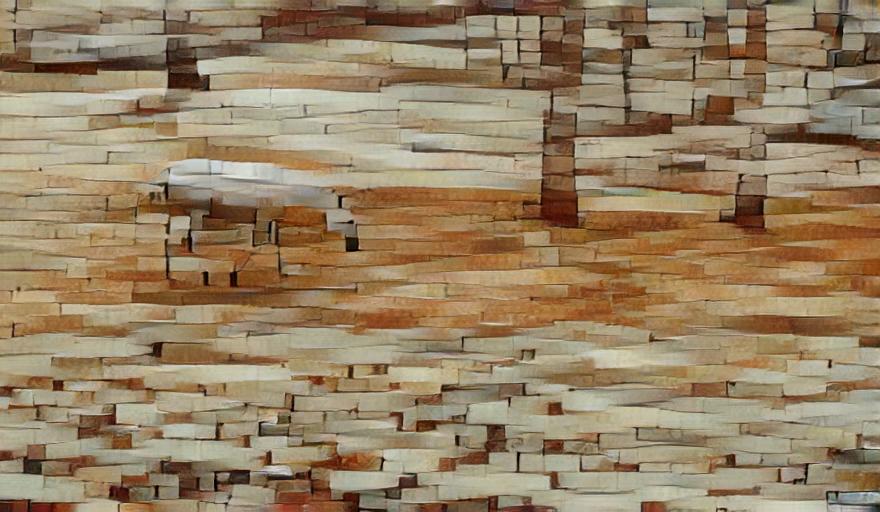}  
  \hspace{-4pt}  
  \includegraphics[height=0.12\textwidth,width=0.14\textwidth]{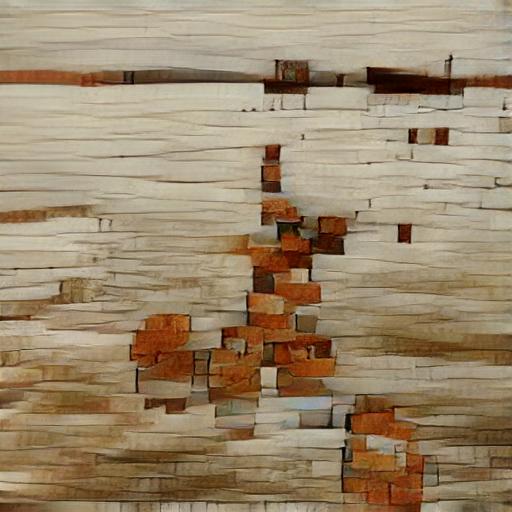}  
  \hspace{-4pt}
 \includegraphics[height=0.12\textwidth,width=0.14\textwidth]{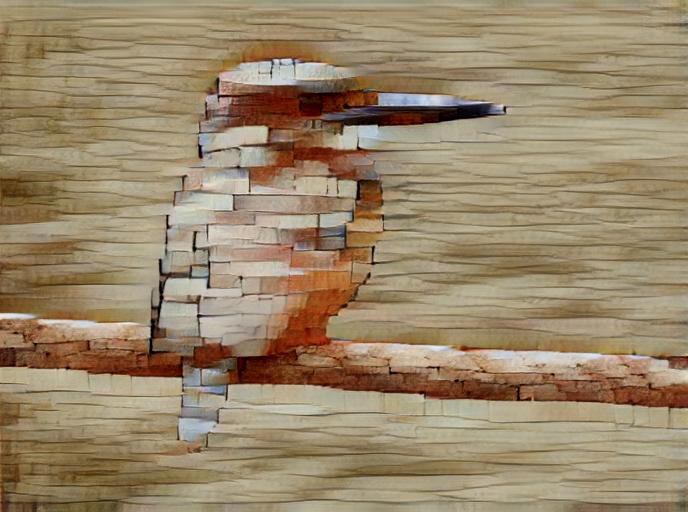}
  \hspace{-4pt}
  \includegraphics[height=0.12\textwidth,width=0.11\textwidth]{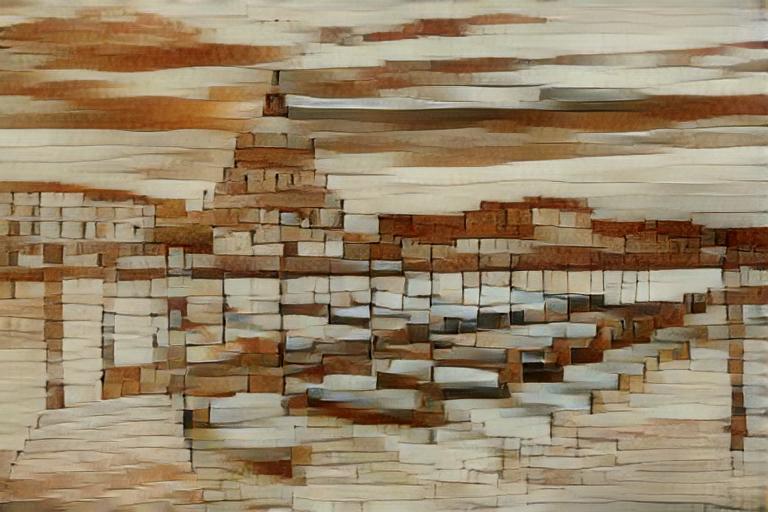} \\ 
  \includegraphics[height=0.12\textwidth,width=0.09\textwidth]{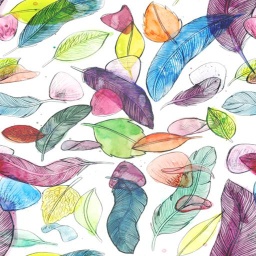}
 \hspace{-4pt} 
  \includegraphics[height=0.12\textwidth,width=0.14\textwidth]{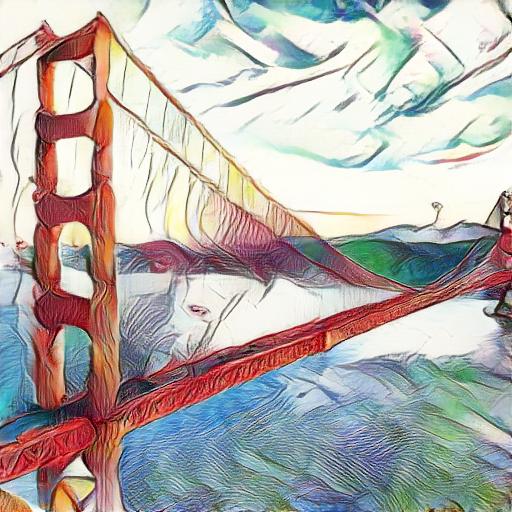}
 \hspace{-4pt} 
 \includegraphics[height=0.12\textwidth,width=0.14\textwidth]{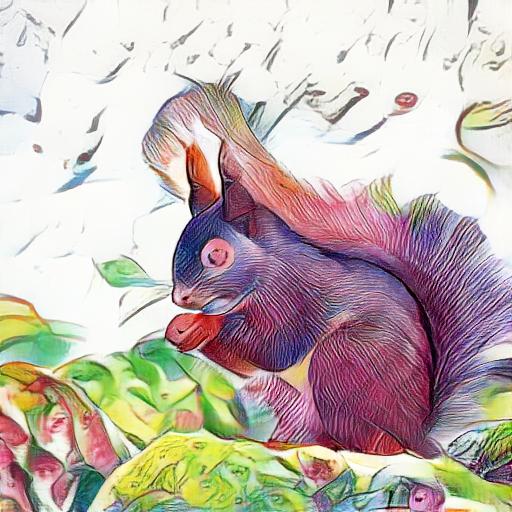}
  \hspace{-4pt}
  \includegraphics[height=0.12\textwidth,width=0.1\textwidth]{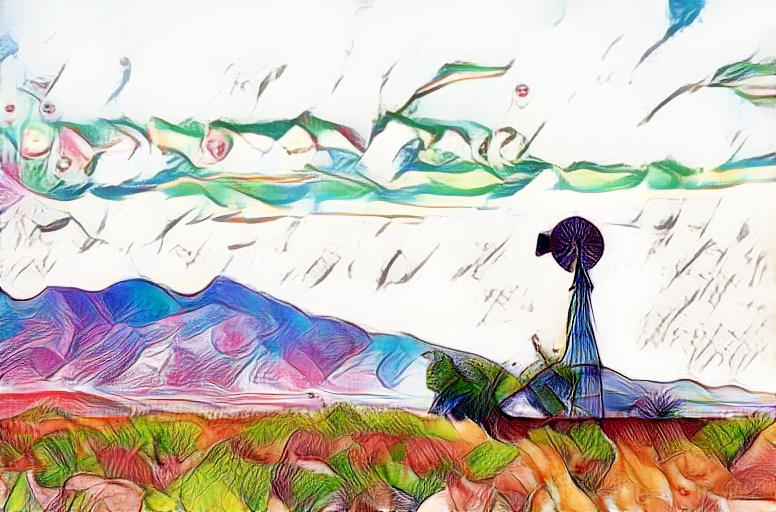}  
  \hspace{-4pt}
  \includegraphics[height=0.12\textwidth,width=0.13\textwidth]{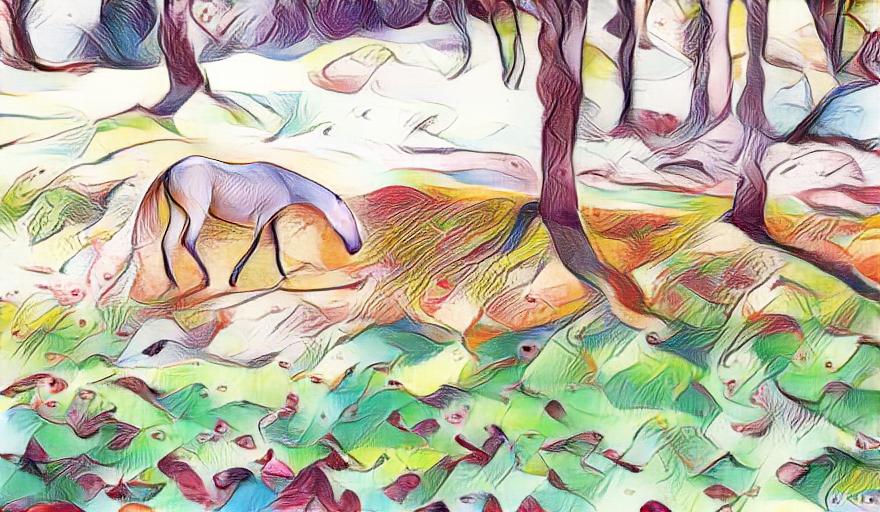}  
  \hspace{-4pt}  
  \includegraphics[height=0.12\textwidth,width=0.14\textwidth]{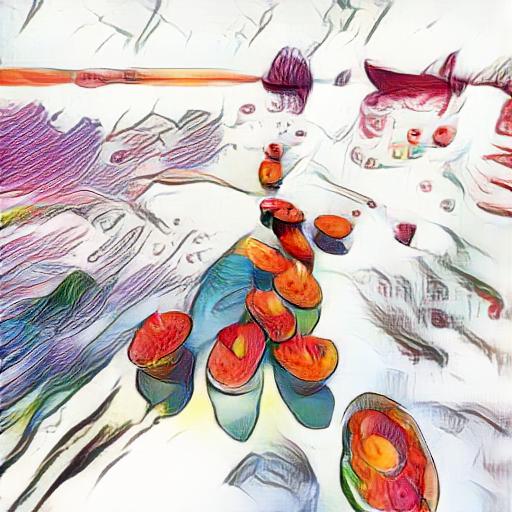}  
  \hspace{-4pt}
 \includegraphics[height=0.12\textwidth,width=0.14\textwidth]{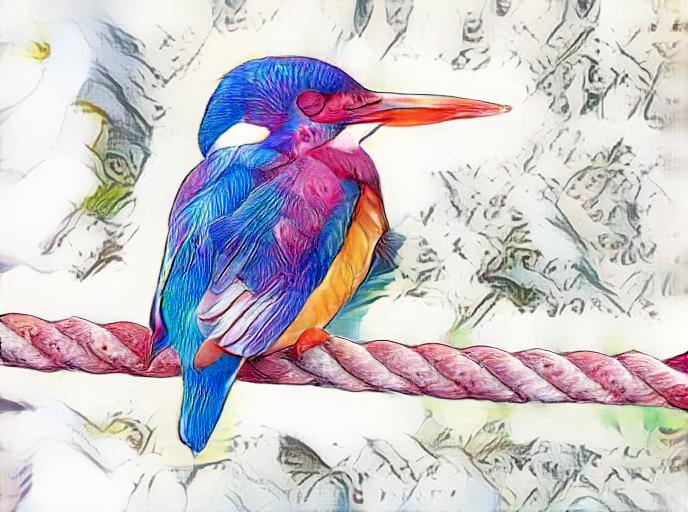}
  \hspace{-4pt}
  \includegraphics[height=0.12\textwidth,width=0.11\textwidth]{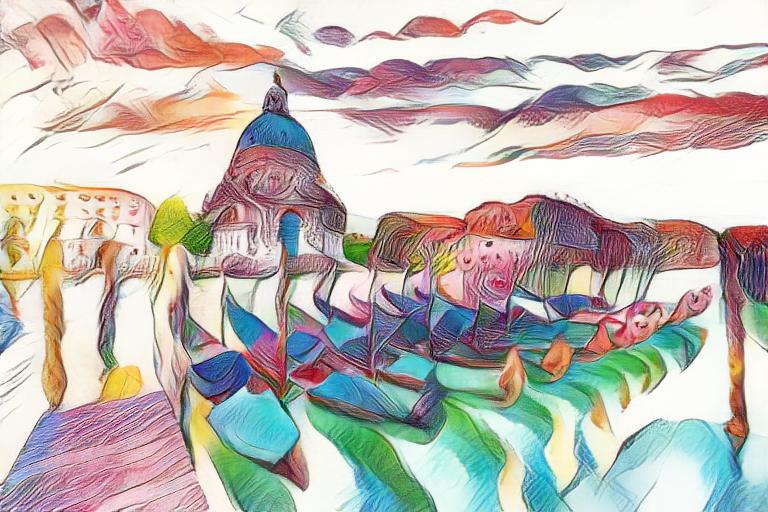} \\  
  \includegraphics[height=0.12\textwidth,width=0.09\textwidth]{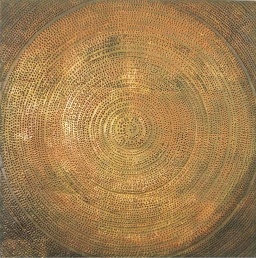}
 \hspace{-4pt} 
  \includegraphics[height=0.12\textwidth,width=0.14\textwidth]{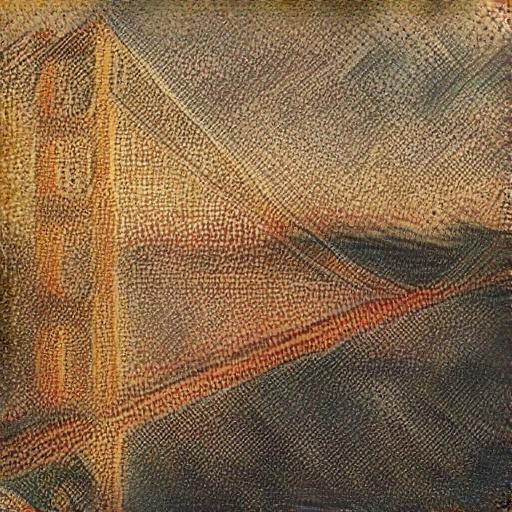}
 \hspace{-4pt} 
 \includegraphics[height=0.12\textwidth,width=0.14\textwidth]{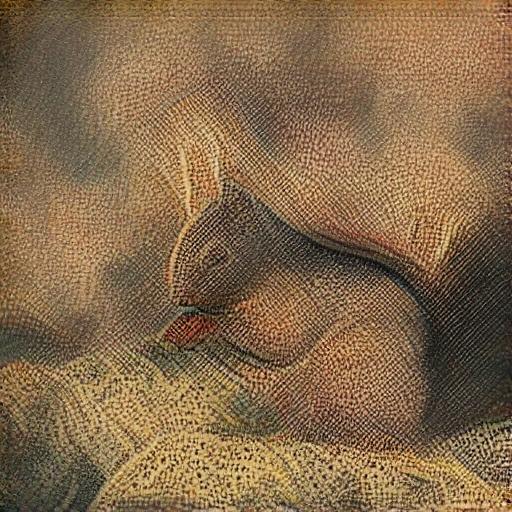}
  \hspace{-4pt}
  \includegraphics[height=0.12\textwidth,width=0.1\textwidth]{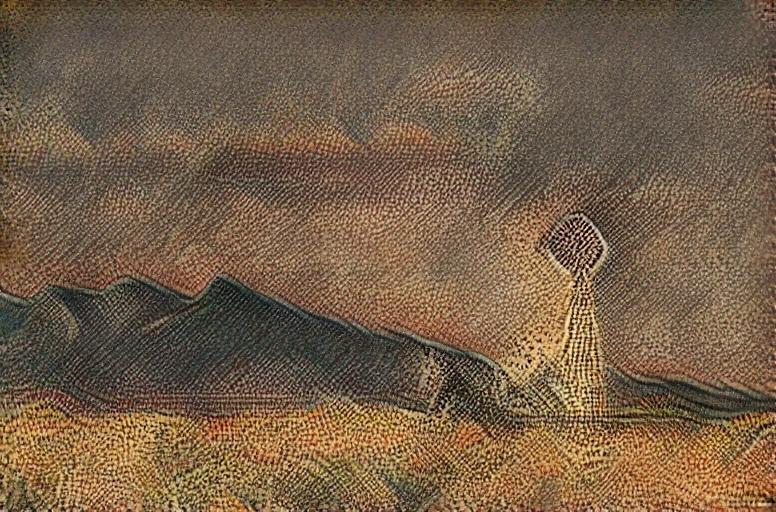}  
  \hspace{-4pt}
  \includegraphics[height=0.12\textwidth,width=0.13\textwidth]{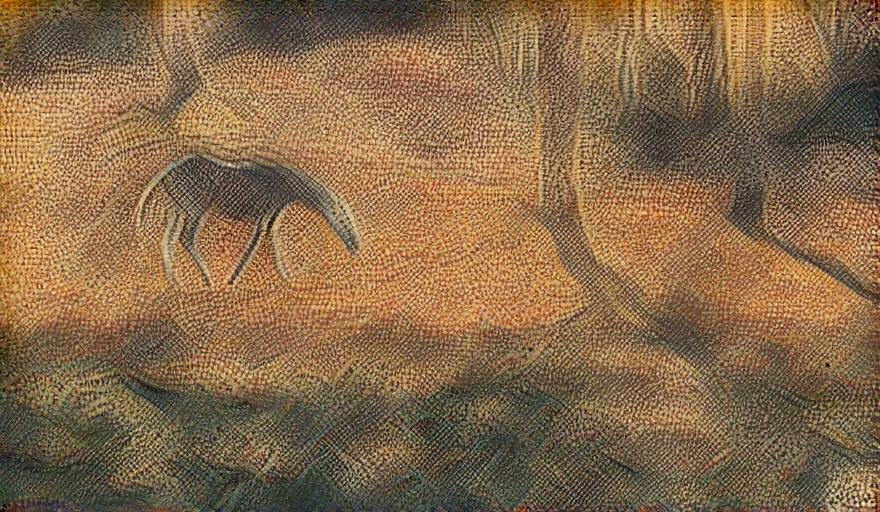}  
  \hspace{-4pt}  
  \includegraphics[height=0.12\textwidth,width=0.14\textwidth]{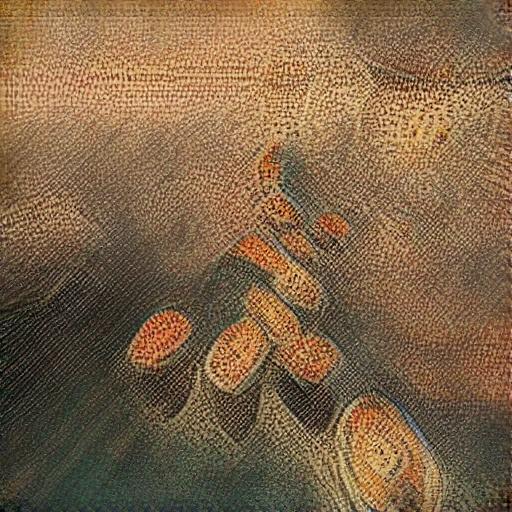}  
  \hspace{-4pt}
 \includegraphics[height=0.12\textwidth,width=0.14\textwidth]{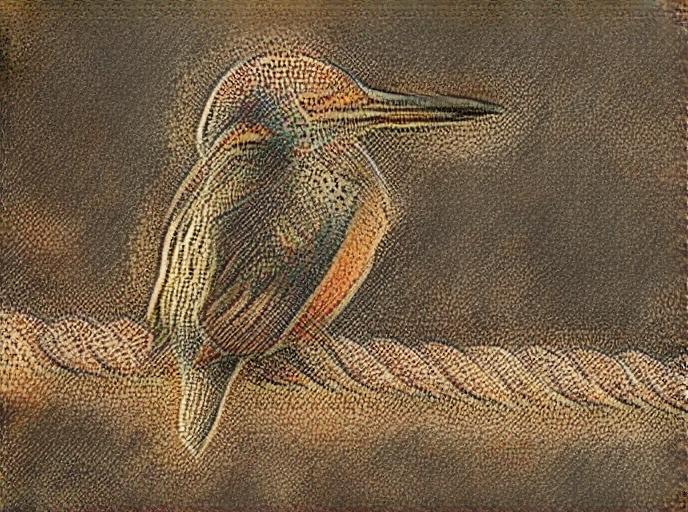}
  \hspace{-4pt}
  \includegraphics[height=0.12\textwidth,width=0.11\textwidth]{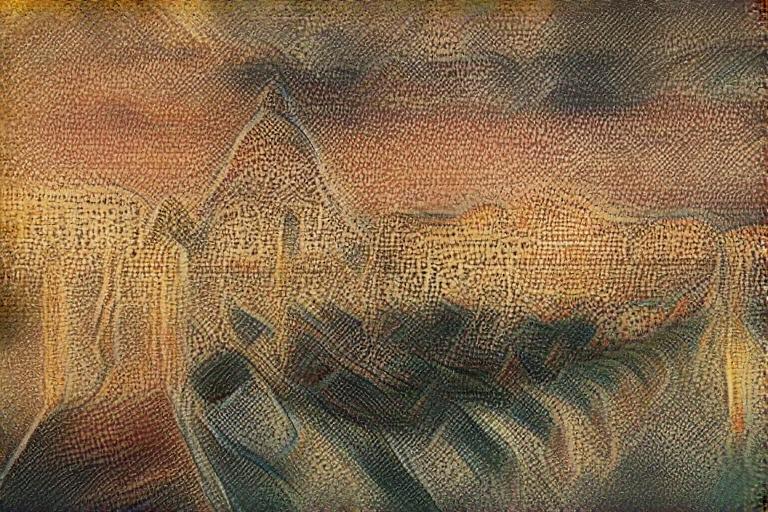} \\  
   \includegraphics[height=0.12\textwidth,width=0.09\textwidth]{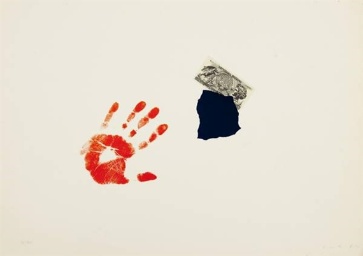}
 \hspace{-4pt} 
  \includegraphics[height=0.12\textwidth,width=0.14\textwidth]{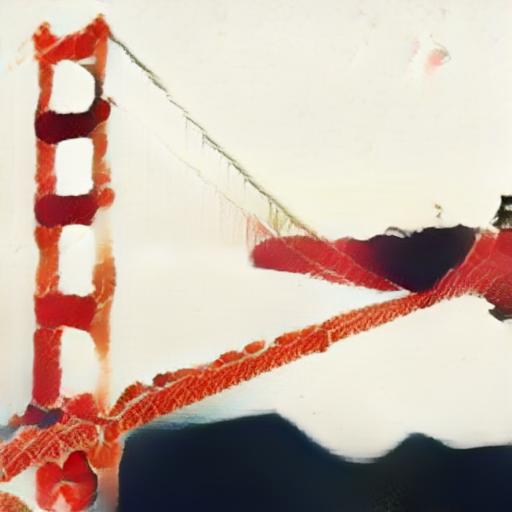}
 \hspace{-4pt} 
 \includegraphics[height=0.12\textwidth,width=0.14\textwidth]{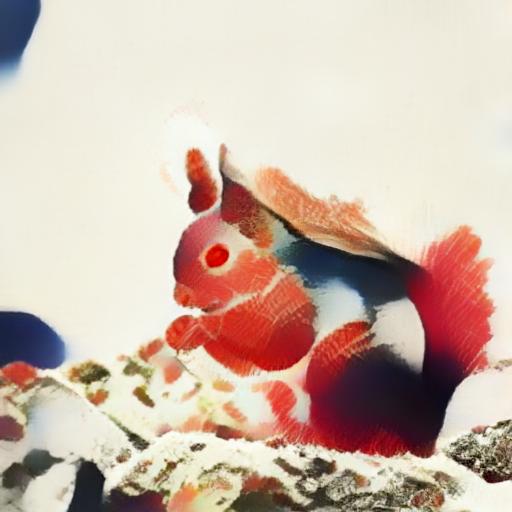}
  \hspace{-4pt}
  \includegraphics[height=0.12\textwidth,width=0.1\textwidth]{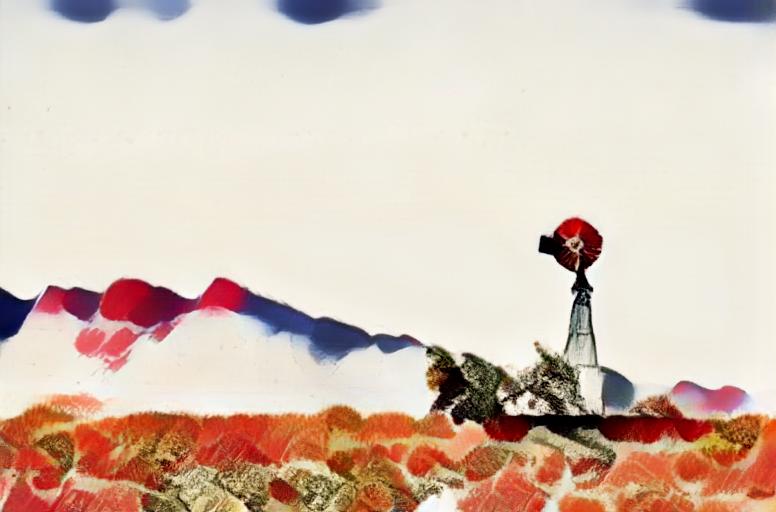}  
  \hspace{-4pt}
  \includegraphics[height=0.12\textwidth,width=0.13\textwidth]{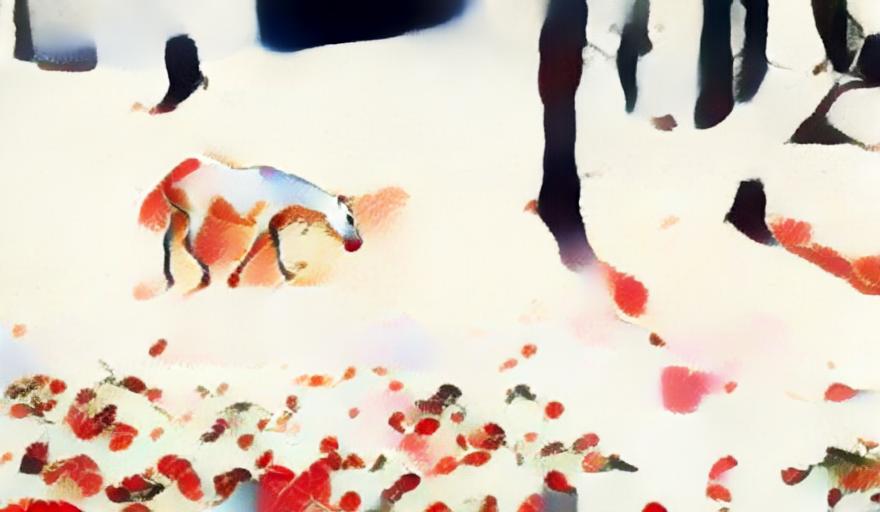}  
  \hspace{-4pt}  
  \includegraphics[height=0.12\textwidth,width=0.14\textwidth]{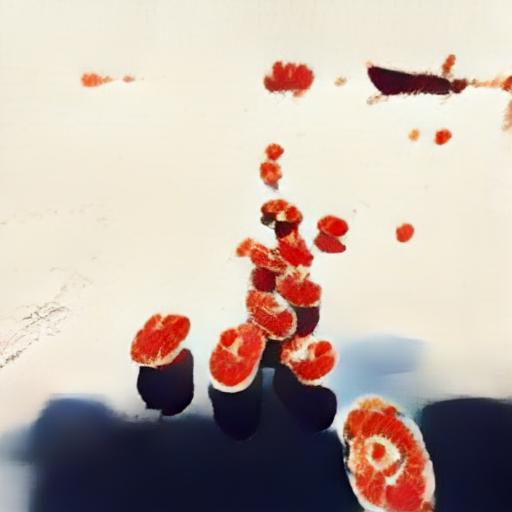}  
  \hspace{-4pt}
 \includegraphics[height=0.12\textwidth,width=0.14\textwidth]{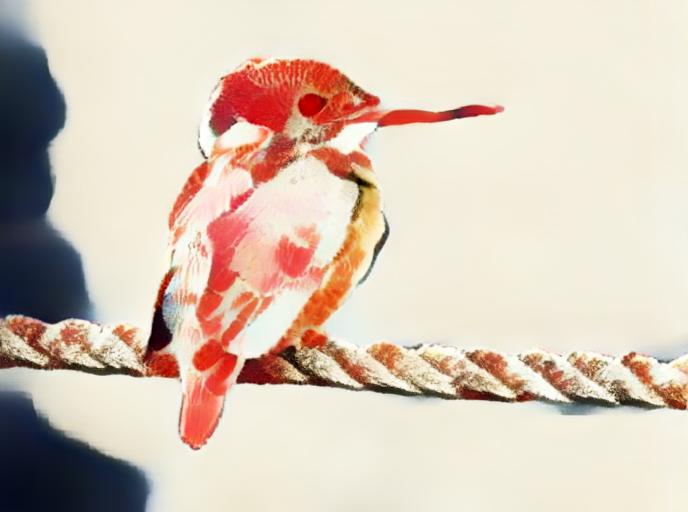}
  \hspace{-4pt}
  \includegraphics[height=0.12\textwidth,width=0.11\textwidth]{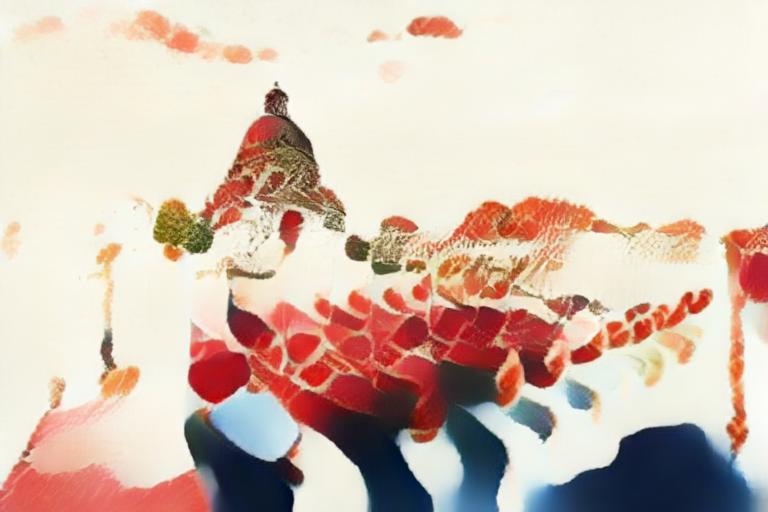} \\   
   \includegraphics[height=0.12\textwidth,width=0.09\textwidth]{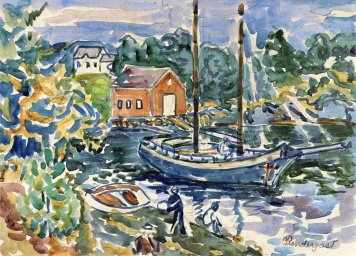}
 \hspace{-4pt} 
  \includegraphics[height=0.12\textwidth,width=0.14\textwidth]{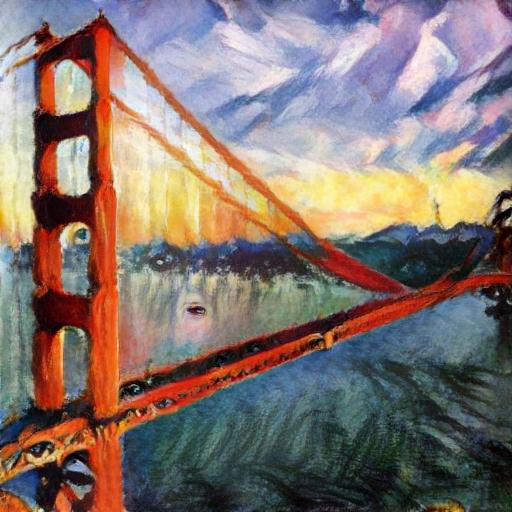}
 \hspace{-4pt} 
 \includegraphics[height=0.12\textwidth,width=0.14\textwidth]{Supp/SOTAComp/ours/2_30}
  \hspace{-4pt}
  \includegraphics[height=0.12\textwidth,width=0.1\textwidth]{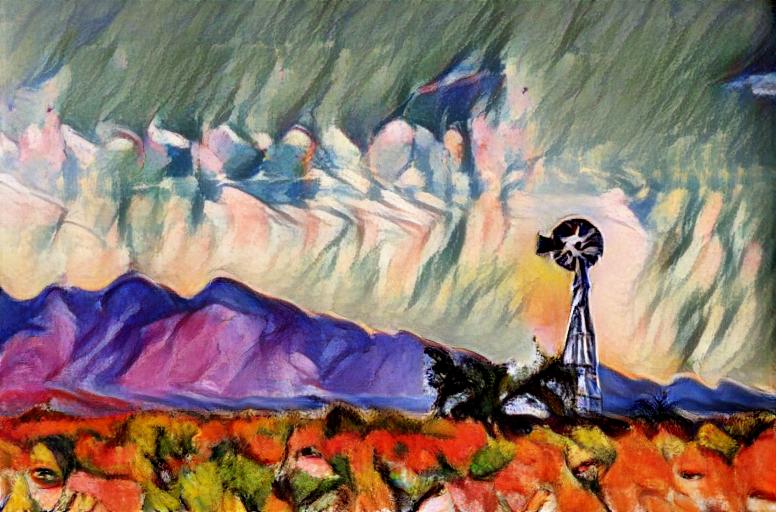}  
  \hspace{-4pt}
  \includegraphics[height=0.12\textwidth,width=0.13\textwidth]{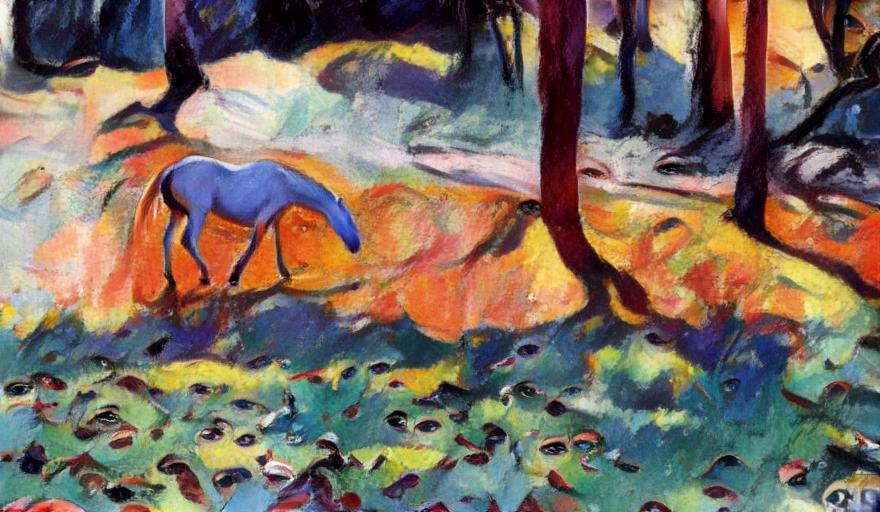}  
  \hspace{-4pt}  
  \includegraphics[height=0.12\textwidth,width=0.14\textwidth]{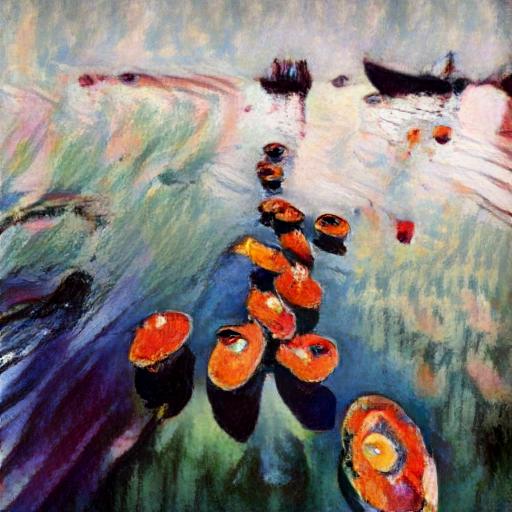}  
  \hspace{-4pt}
 \includegraphics[height=0.12\textwidth,width=0.14\textwidth]{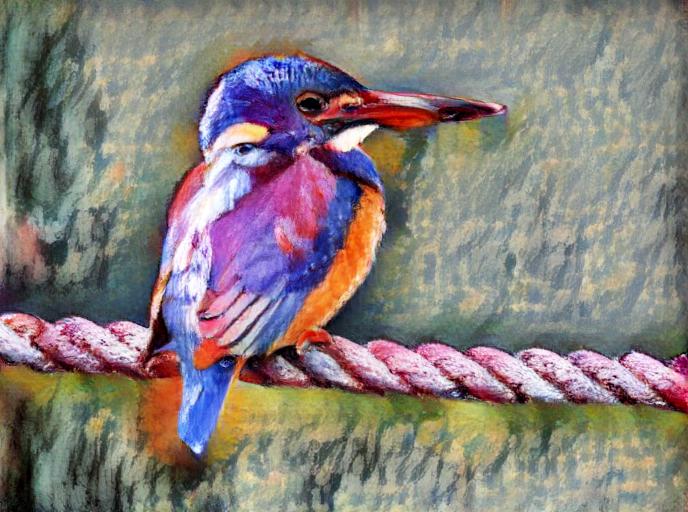}
  \hspace{-4pt}
  \includegraphics[height=0.12\textwidth,width=0.11\textwidth]{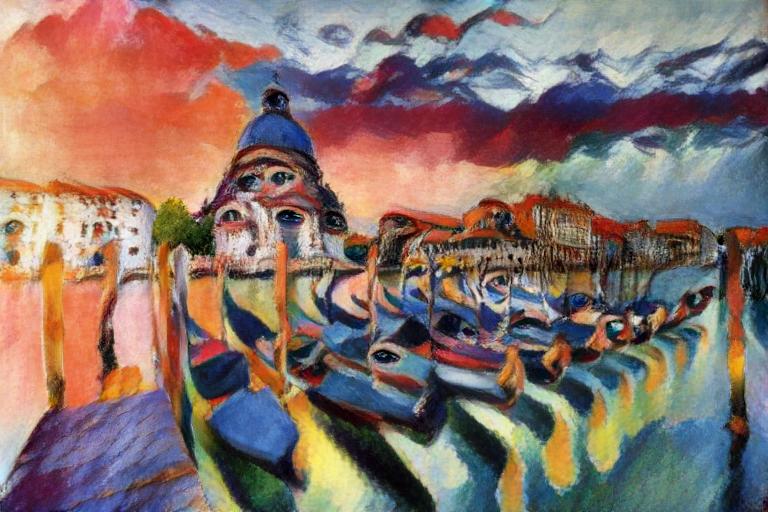} \\ 
    \includegraphics[height=0.12\textwidth,width=0.09\textwidth]{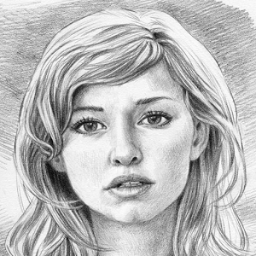}
 \hspace{-4pt} 
  \includegraphics[height=0.12\textwidth,width=0.14\textwidth]{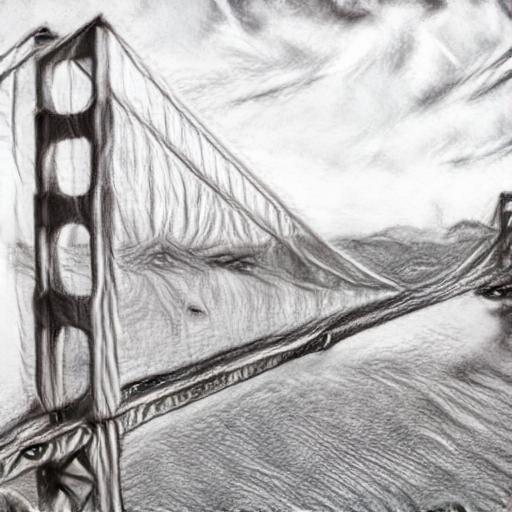}
 \hspace{-4pt} 
 \includegraphics[height=0.12\textwidth,width=0.14\textwidth]{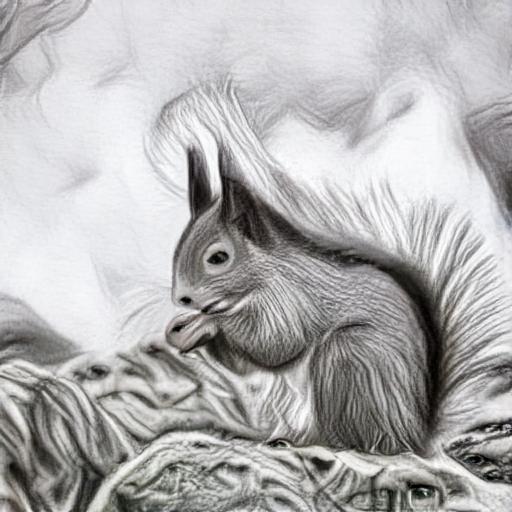}
  \hspace{-4pt}
  \includegraphics[height=0.12\textwidth,width=0.1\textwidth]{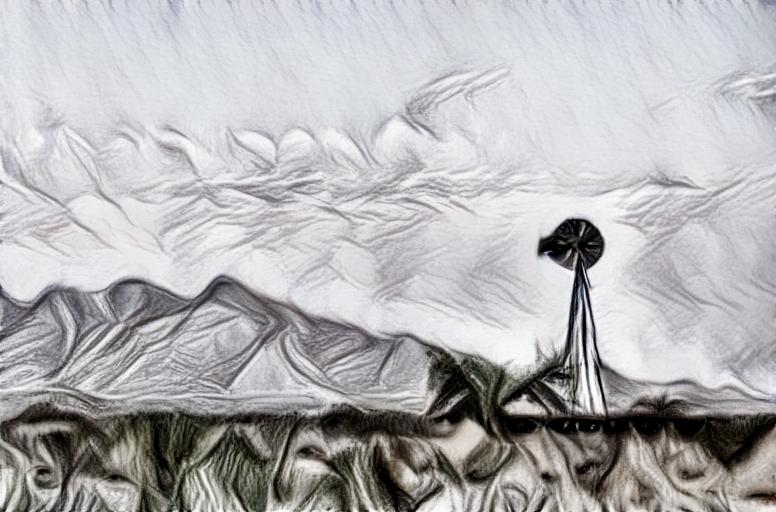}  
  \hspace{-4pt}
  \includegraphics[height=0.12\textwidth,width=0.13\textwidth]{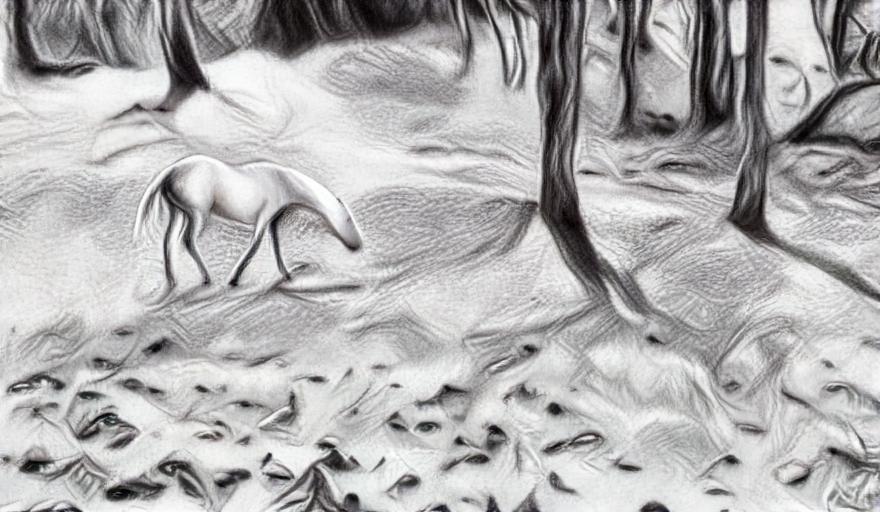}  
  \hspace{-4pt}  
  \includegraphics[height=0.12\textwidth,width=0.14\textwidth]{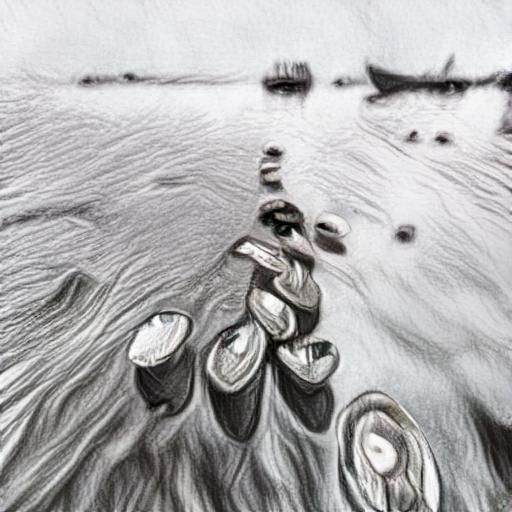}  
  \hspace{-4pt}
 \includegraphics[height=0.12\textwidth,width=0.14\textwidth]{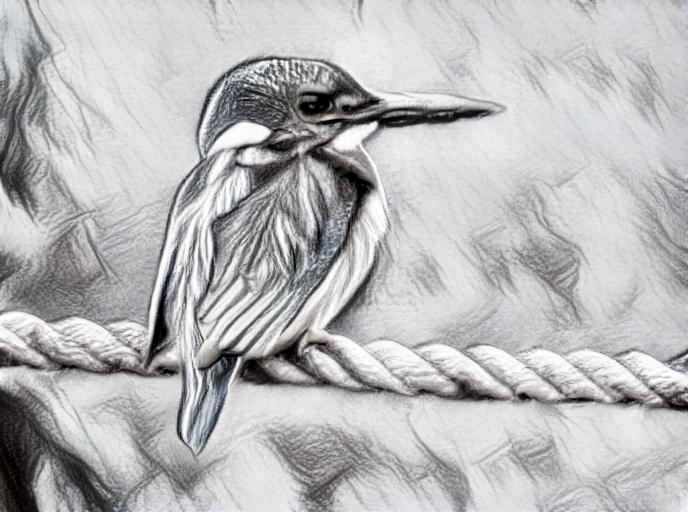}
  \hspace{-4pt}
  \includegraphics[height=0.12\textwidth,width=0.11\textwidth]{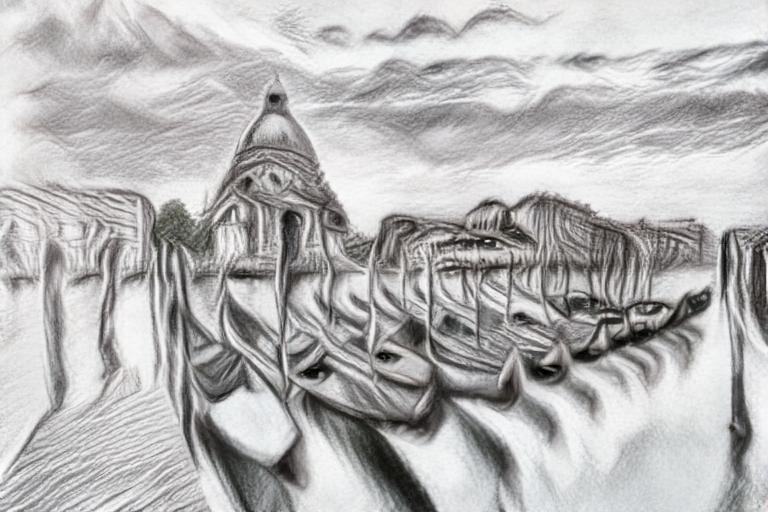} \\  
  \vspace{-0.2cm}
\caption{More qualitative performance on arbitrary style transfer. The first row and column are the content images and style images, respectively. Others are the stylization results created by our method.}
\label{fig:comp_arbitrary_style_transfer}
\vspace{-0.5cm} 
\end{figure*}  


\begin{figure*}[ht!]
\captionsetup[subfigure]{justification=raggedright, singlelinecheck=false, labelformat=empty}  
\centering

 \includegraphics[height=0.11\textwidth,width=0.16\textwidth]{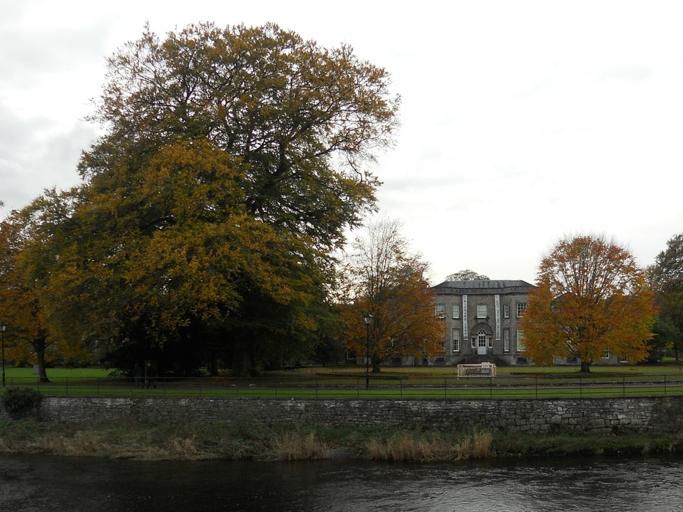}
  \hspace{-4pt}
 \includegraphics[height=0.11\textwidth,width=0.14\textwidth]{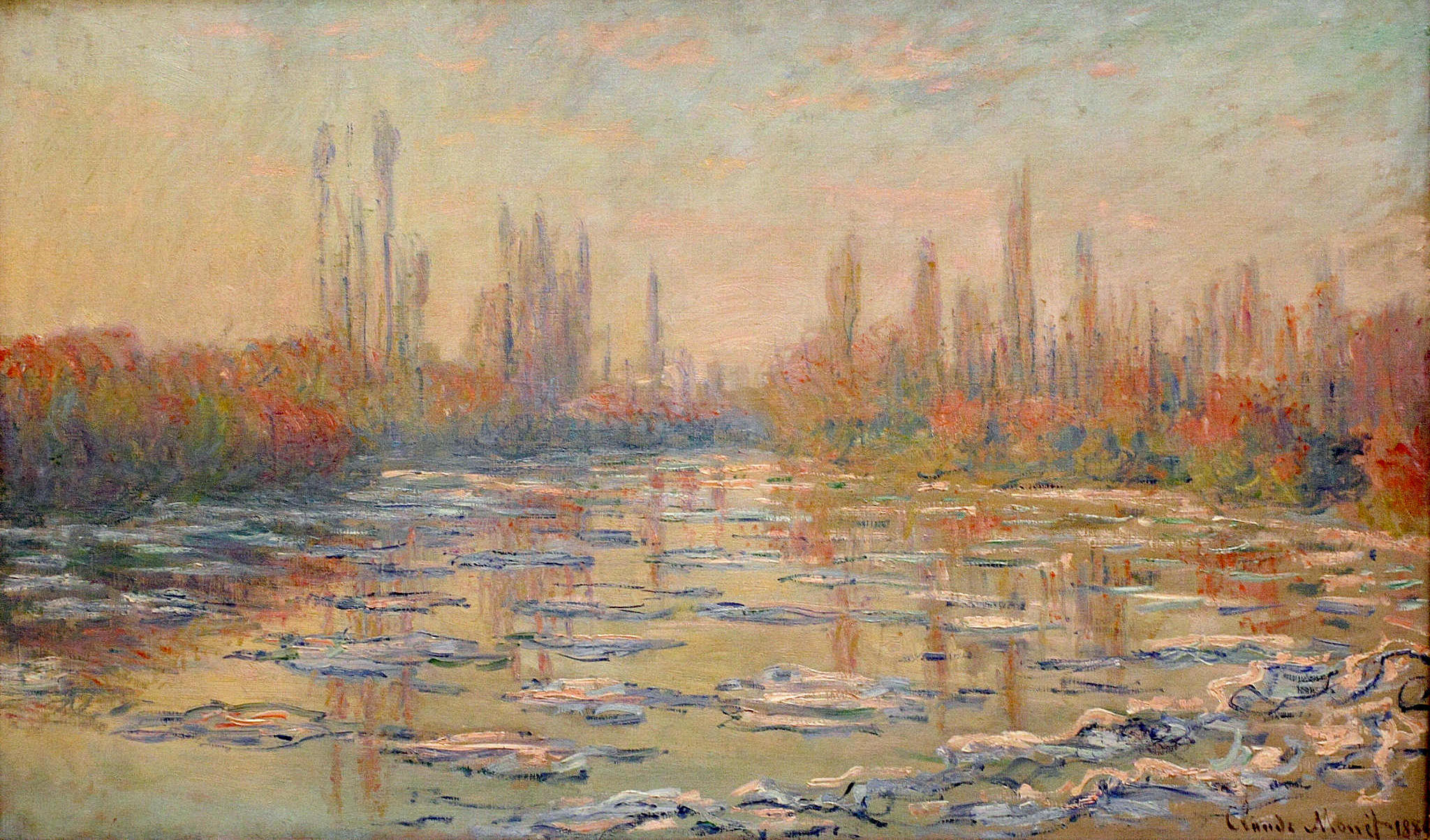}
  \hspace{-4pt} 
  \includegraphics[height=0.11\textwidth,width=0.16\textwidth]{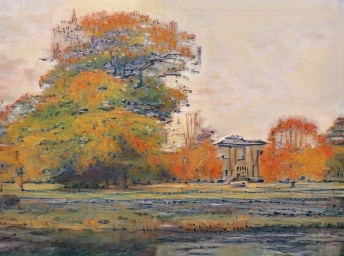}  
  \hspace{-4pt}
  \includegraphics[height=0.11\textwidth,width=0.16\textwidth]{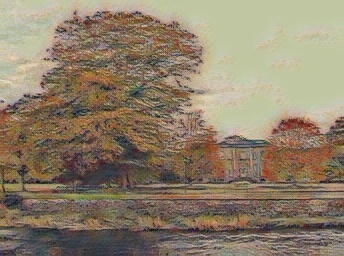}  
  \hspace{-4pt}  
  \includegraphics[height=0.11\textwidth,width=0.16\textwidth]{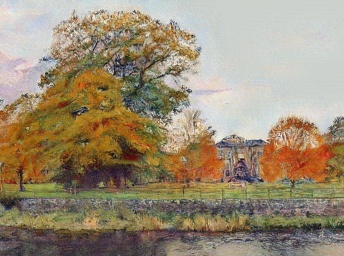}  
  \hspace{-4pt}  
    \includegraphics[height=0.11\textwidth,width=0.16\textwidth]{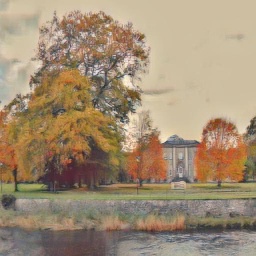}  \\

  \includegraphics[height=0.11\textwidth,width=0.16\textwidth]{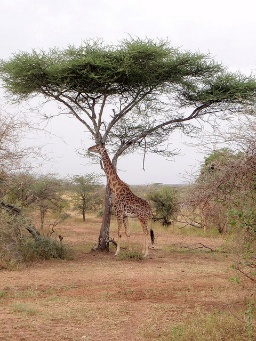}  
  \hspace{-4pt}
  \includegraphics[height=0.11\textwidth,width=0.14\textwidth]{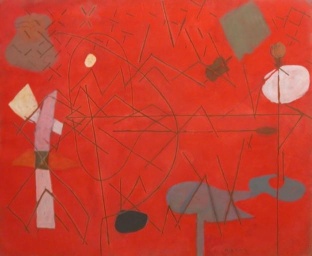}  
  \hspace{-4pt}
  \includegraphics[height=0.11\textwidth,width=0.16\textwidth]{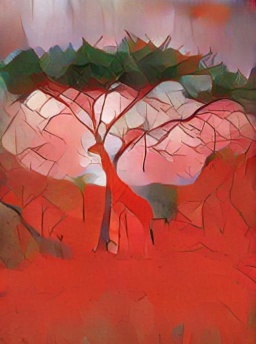}
  \hspace{-4pt} 
  \includegraphics[height=0.11\textwidth,width=0.16\textwidth]{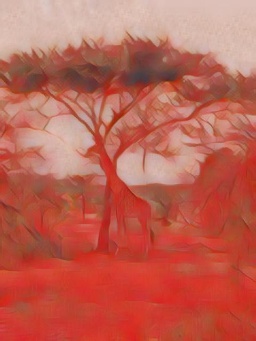}
  \hspace{-4pt}  
  \includegraphics[height=0.11\textwidth,width=0.16\textwidth]{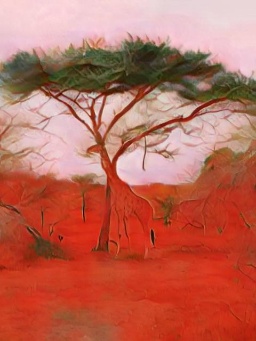}
  \hspace{-4pt}  
    \includegraphics[height=0.11\textwidth,width=0.16\textwidth]{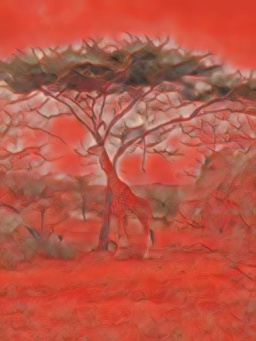}\\

 \includegraphics[height=0.11\textwidth,width=0.16\textwidth]{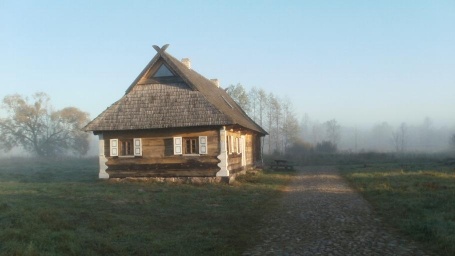}
  \hspace{-4pt}
 \includegraphics[height=0.11\textwidth,width=0.14\textwidth]{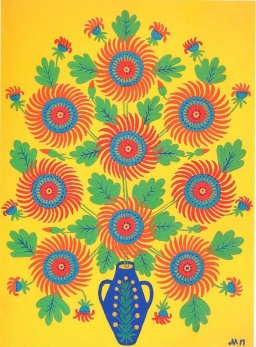}
  \hspace{-4pt}
    \includegraphics[height=0.11\textwidth,width=0.16\textwidth]{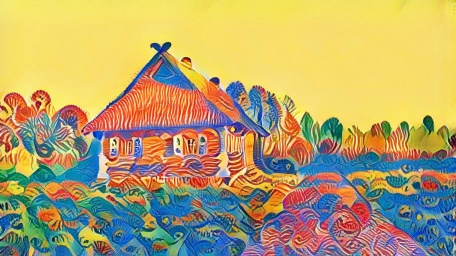}  
  \hspace{-4pt}
  \includegraphics[height=0.11\textwidth,width=0.16\textwidth]{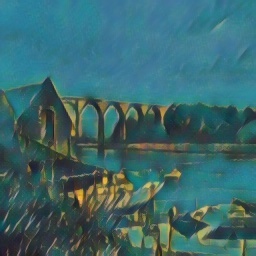}  
  \hspace{-4pt}
  \includegraphics[height=0.11\textwidth,width=0.16\textwidth]{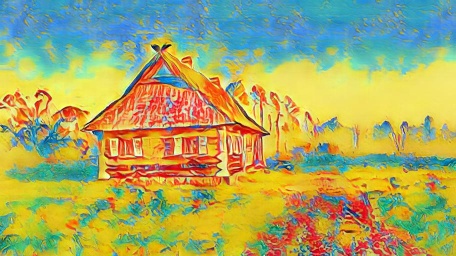}  
  \hspace{-4pt} 
    \includegraphics[height=0.11\textwidth,width=0.16\textwidth]{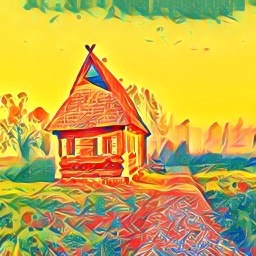}  \\

 \includegraphics[height=0.11\textwidth,width=0.16\textwidth]{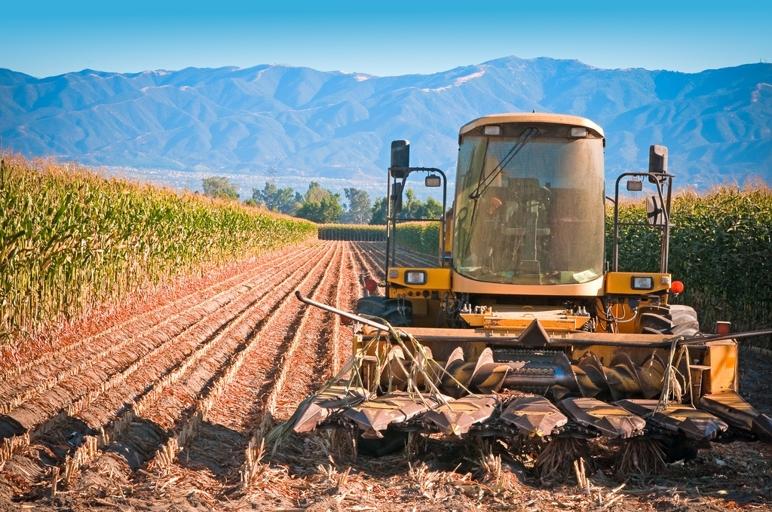}
  \hspace{-4pt}
 \includegraphics[height=0.11\textwidth,width=0.14\textwidth]{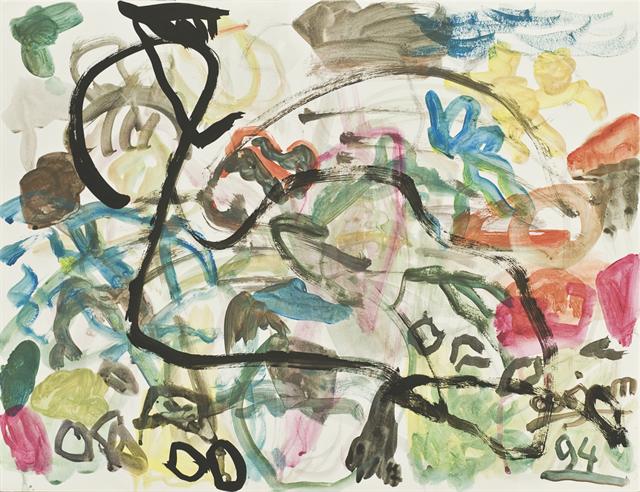}
  \hspace{-4pt}
    \includegraphics[height=0.11\textwidth,width=0.16\textwidth]{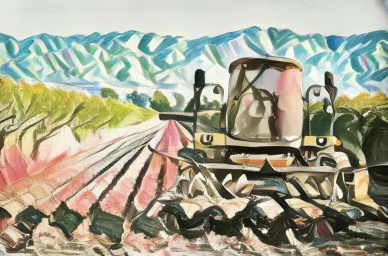}  
  \hspace{-4pt}
  \includegraphics[height=0.11\textwidth,width=0.16\textwidth]{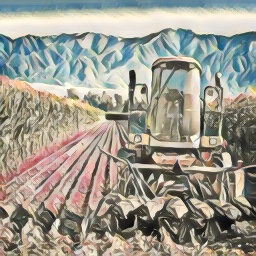}  
  \hspace{-4pt}
  \includegraphics[height=0.11\textwidth,width=0.16\textwidth]{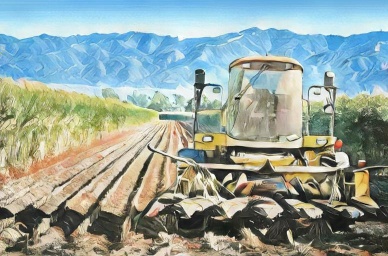}  
  \hspace{-4pt} 
    \includegraphics[height=0.11\textwidth,width=0.16\textwidth]{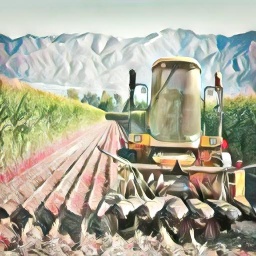}  \\

 \includegraphics[height=0.11\textwidth,width=0.16\textwidth]{SOTAComp/cc/2}
  \hspace{-4pt}
 \includegraphics[height=0.11\textwidth,width=0.14\textwidth]{SOTAComp/ss/2}
  \hspace{-4pt}
    \includegraphics[height=0.11\textwidth,width=0.16\textwidth]{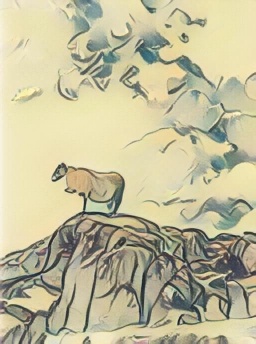}  
  \hspace{-4pt}
  \includegraphics[height=0.11\textwidth,width=0.16\textwidth]{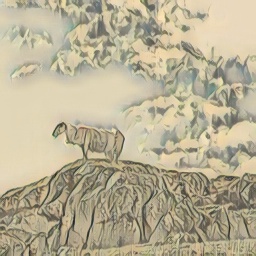}  
  \hspace{-4pt}
  \includegraphics[height=0.11\textwidth,width=0.16\textwidth]{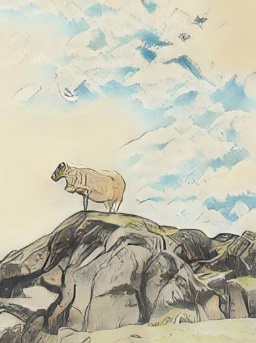}  
  \hspace{-4pt} 
    \includegraphics[height=0.11\textwidth,width=0.16\textwidth]{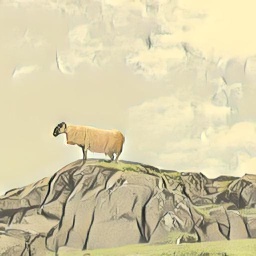}  \\

 \includegraphics[height=0.11\textwidth,width=0.16\textwidth]{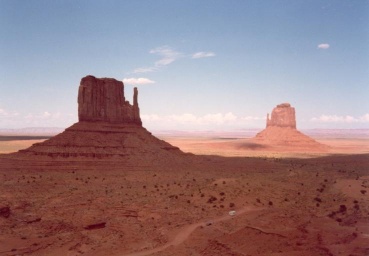}
  \hspace{-4pt}
 \includegraphics[height=0.11\textwidth,width=0.14\textwidth]{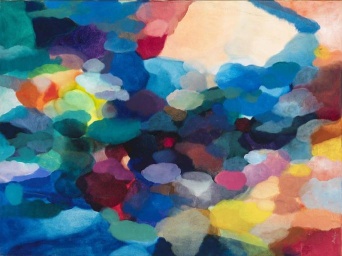}
  \hspace{-4pt}
    \includegraphics[height=0.11\textwidth,width=0.16\textwidth]{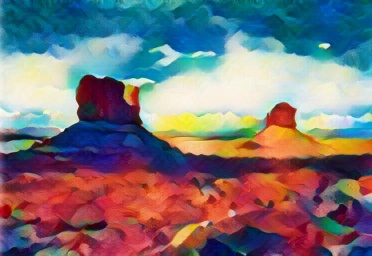}  
  \hspace{-4pt}
  \includegraphics[height=0.11\textwidth,width=0.16\textwidth]{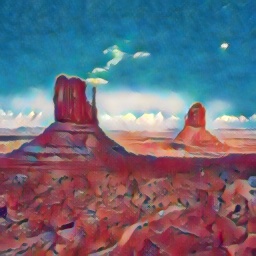}  
  \hspace{-4pt}
  \includegraphics[height=0.11\textwidth,width=0.16\textwidth]{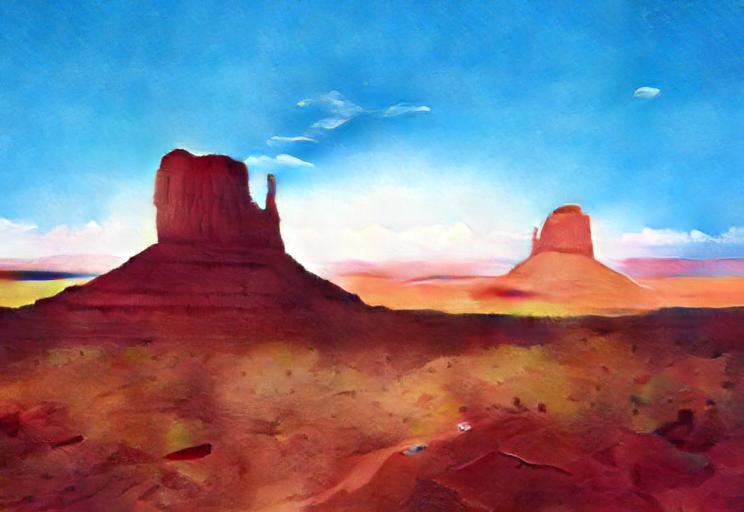}  
  \hspace{-4pt} 
    \includegraphics[height=0.11\textwidth,width=0.16\textwidth]{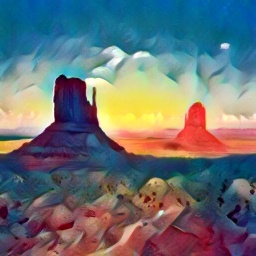}  \\

 \includegraphics[height=0.11\textwidth,width=0.16\textwidth]{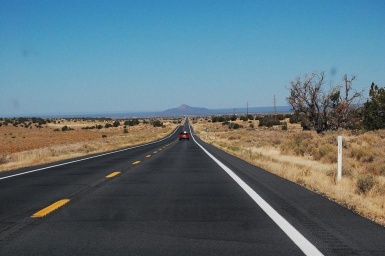}
  \hspace{-4pt}
 \includegraphics[height=0.11\textwidth,width=0.14\textwidth]{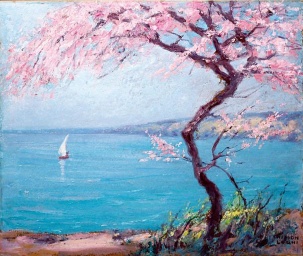}
  \hspace{-4pt}
    \includegraphics[height=0.11\textwidth,width=0.16\textwidth]{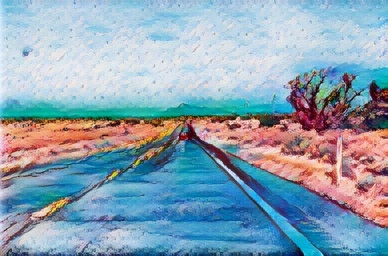}  
  \hspace{-4pt}
  \includegraphics[height=0.11\textwidth,width=0.16\textwidth]{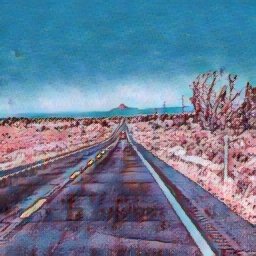}  
  \hspace{-4pt}
  \includegraphics[height=0.11\textwidth,width=0.16\textwidth]{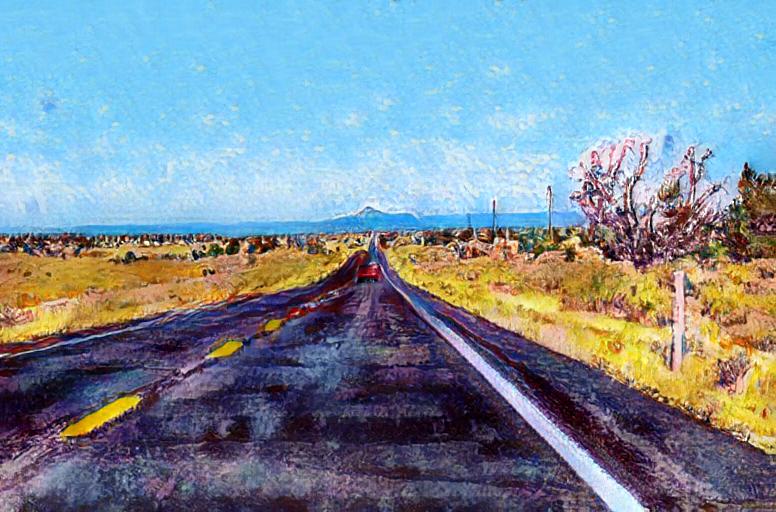}  
  \hspace{-4pt} 
    \includegraphics[height=0.11\textwidth,width=0.16\textwidth]{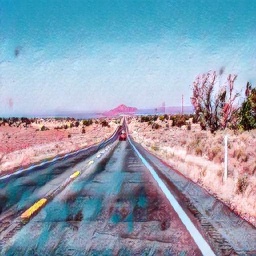}  \\

 \includegraphics[height=0.11\textwidth,width=0.16\textwidth]{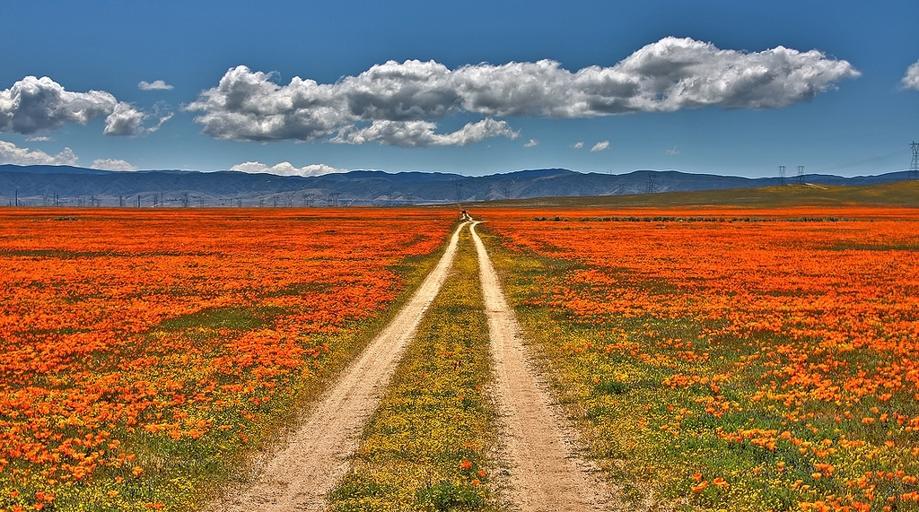}
  \hspace{-4pt}
 \includegraphics[height=0.11\textwidth,width=0.14\textwidth]{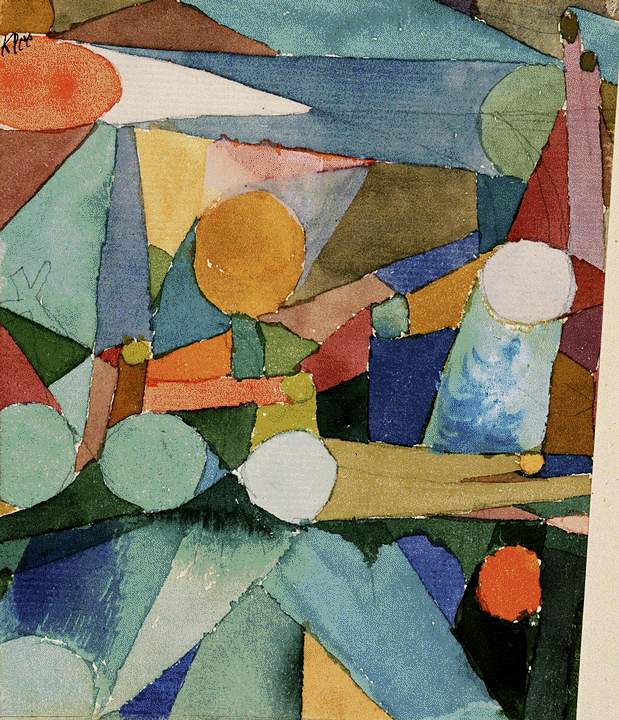}
  \hspace{-4pt}
    \includegraphics[height=0.11\textwidth,width=0.16\textwidth]{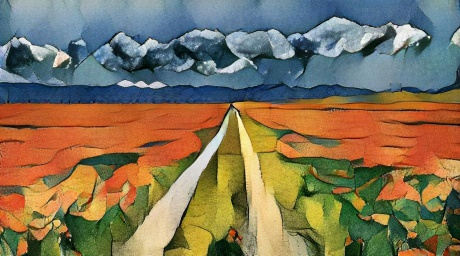}  
  \hspace{-4pt}
  \includegraphics[height=0.11\textwidth,width=0.16\textwidth]{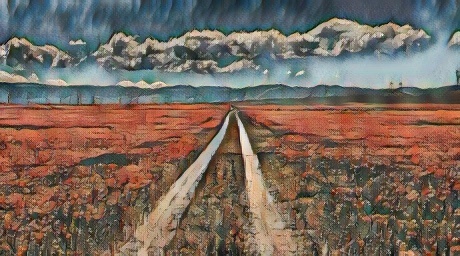}  
  \hspace{-4pt}
  \includegraphics[height=0.11\textwidth,width=0.16\textwidth]{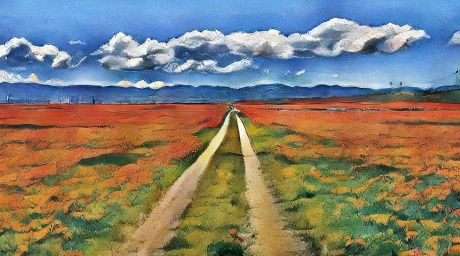}  
  \hspace{-4pt} 
    \includegraphics[height=0.11\textwidth,width=0.16\textwidth]{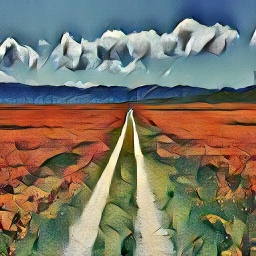}  \\

 \raisebox{0.005\textwidth }{\makebox[0.03\textwidth]{\parbox{0.95\linewidth}{\scriptsize \hspace{ 30pt} Content\hspace{50pt} Style \hspace{ 60pt} Ours  \hspace{50pt} AdaAttN'21  \hspace{ 50pt} IEC'21 \hspace{54pt} StrTr$^2$'22 }}}
 
\vspace{-0.3cm}
\caption{Qualitative performance comparison on stylized results.}
\label{fig:comp_sota_style_transfer}
\vspace{-0.3cm} 
\end{figure*} 


\begin{figure*}[ht!]
\vspace{-0.2cm}
\centering
\captionsetup[subfigure]{justification=raggedright, singlelinecheck=false, labelformat=empty}   
 \centering
  \raisebox{0.036\textwidth }{\makebox[0.11\textwidth]{}}
    \includegraphics[height=0.1\textwidth, width=0.16\textwidth]{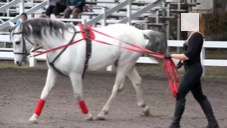}  
    \includegraphics[height=0.1\textwidth, width=0.16\textwidth]{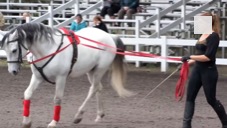}
    \includegraphics[height=0.1\textwidth, width=0.16\textwidth]{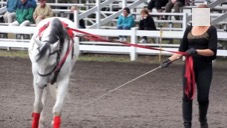}
    \includegraphics[height=0.1\textwidth, width=0.16\textwidth]{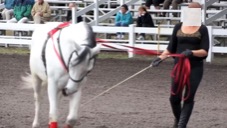} 
    \includegraphics[height=0.1\textwidth, width=0.16\textwidth]{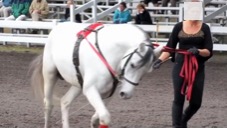} 
    \raisebox{0.036\textwidth }{\makebox[0.05\textwidth]{\scriptsize  Content}} \\
     \includegraphics[height=0.1\textwidth, width=0.11\textwidth]{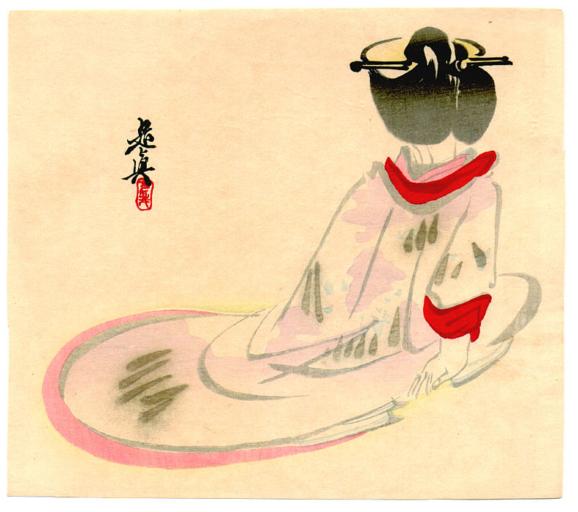} 
     \includegraphics[height=0.1\textwidth, width=0.16\textwidth]{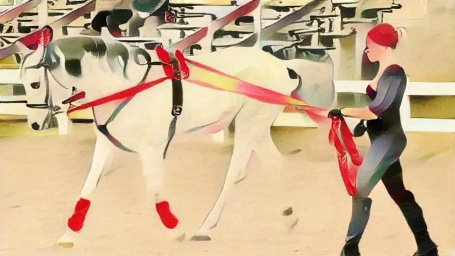}    
     \includegraphics[height=0.1\textwidth, width=0.16\textwidth]{Video_horse/ours/37}
      \includegraphics[height=0.1\textwidth, width=0.16\textwidth]{Video_horse/ours/69}
       \includegraphics[height=0.1\textwidth, width=0.16\textwidth]{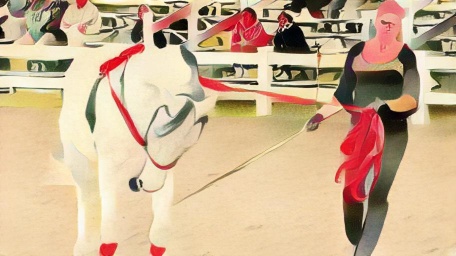}
       \includegraphics[height=0.1\textwidth, width=0.16\textwidth]{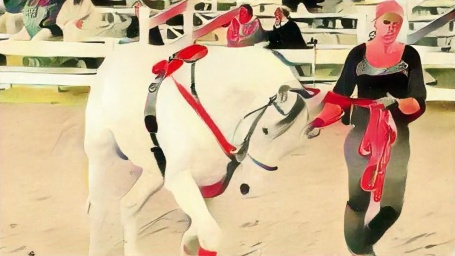} 
        \raisebox{0.036\textwidth }{\makebox[0.05\textwidth]{\scriptsize  Ours}}  \\
             \includegraphics[height=0.1\textwidth, width=0.11\textwidth]{Video_horse/ss/34} 
         \includegraphics[height=0.1\textwidth, width=0.16\textwidth]{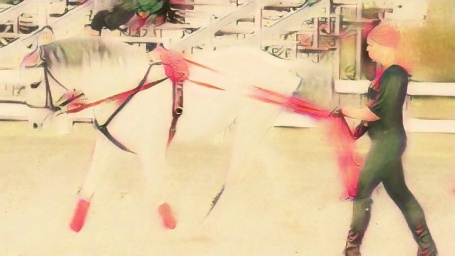}   
        \includegraphics[height=0.1\textwidth, width=0.16\textwidth]{Video_horse/DRB-GAN/37}
         \includegraphics[height=0.1\textwidth, width=0.16\textwidth]{Video_horse/DRB-GAN/69}
          \includegraphics[height=0.1\textwidth, width=0.16\textwidth]{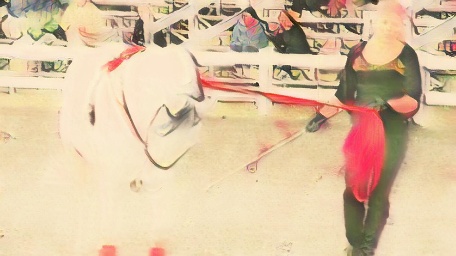}
    \includegraphics[height=0.1\textwidth, width=0.16\textwidth]{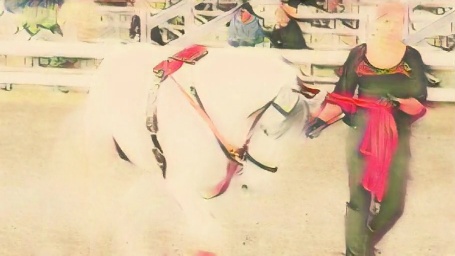}
            \raisebox{0.036\textwidth }{\makebox[0.05\textwidth]{\scriptsize  DRB-GAN}} \\
        \includegraphics[height=0.1\textwidth, width=0.11\textwidth]{Video_horse/ss/34}
           \includegraphics[height=0.1\textwidth, width=0.16\textwidth]{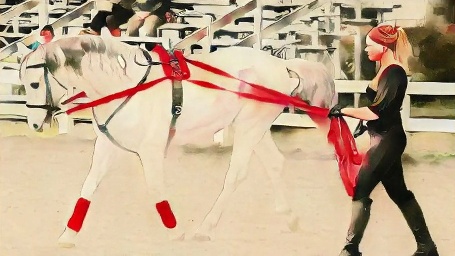}     
           \includegraphics[height=0.1\textwidth, width=0.16\textwidth]{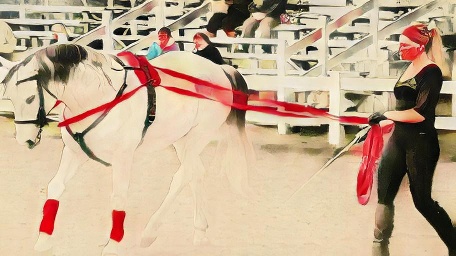}
            \includegraphics[height=0.1\textwidth, width=0.16\textwidth]{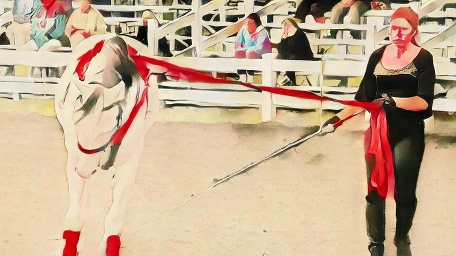}
             \includegraphics[height=0.1\textwidth, width=0.16\textwidth]{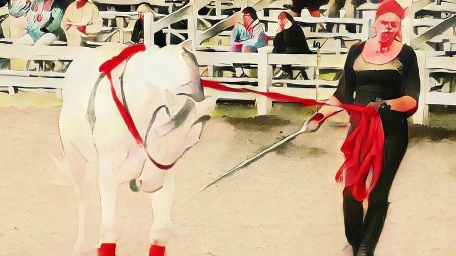}
        \includegraphics[height=0.1\textwidth, width=0.16\textwidth]{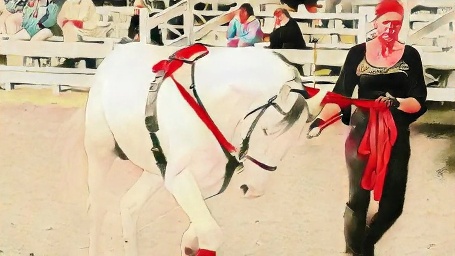}
        \raisebox{0.036\textwidth }{\makebox[0.05\textwidth]{\scriptsize  IEC}} \\
    \includegraphics[height=0.1\textwidth, width=0.11\textwidth]{Video_horse/ss/34}    
 \includegraphics[height=0.1\textwidth, width=0.16\textwidth]{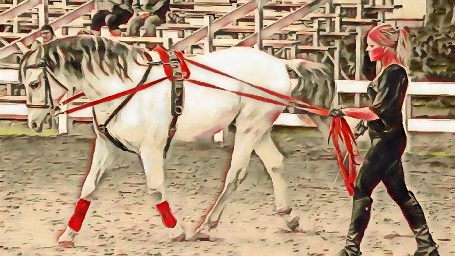}
 \includegraphics[height=0.1\textwidth, width=0.16\textwidth]{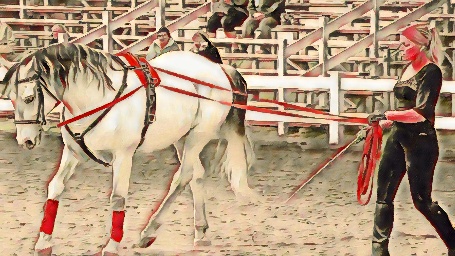}
  \includegraphics[height=0.1\textwidth, width=0.16\textwidth]{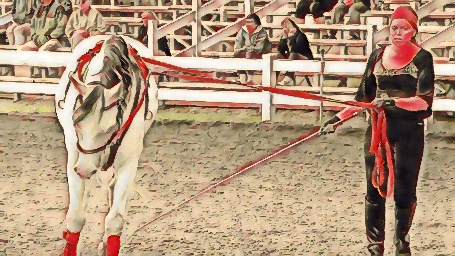}
   \includegraphics[height=0.1\textwidth, width=0.16\textwidth]{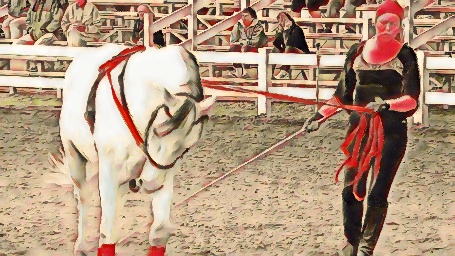}
      \includegraphics[height=0.1\textwidth, width=0.16\textwidth]{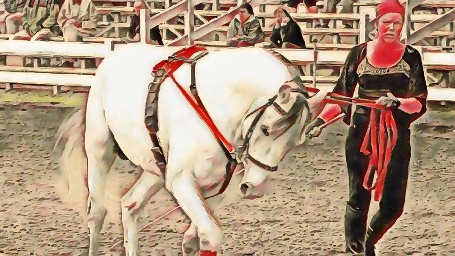}
        \raisebox{0.036\textwidth }{\makebox[0.05\textwidth]{\scriptsize  AdaAttN}} \\
             \includegraphics[height=0.1\textwidth, width=0.11\textwidth]{Video_horse/ss/34} 
        \includegraphics[height=0.1\textwidth, width=0.16\textwidth]{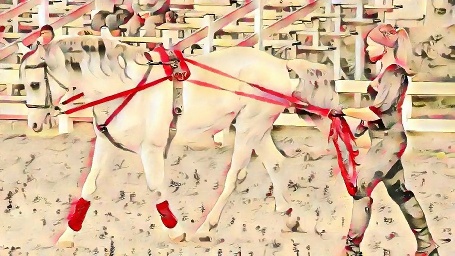}    
        \includegraphics[height=0.1\textwidth, width=0.16\textwidth]{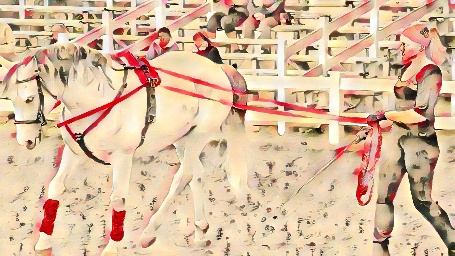}  
            \includegraphics[height=0.1\textwidth, width=0.16\textwidth]{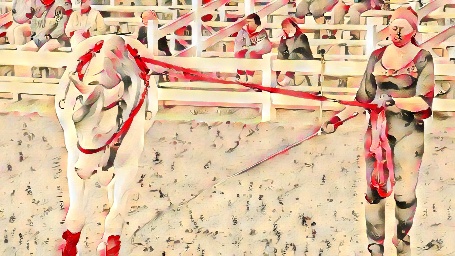} 
            \includegraphics[height=0.1\textwidth, width=0.16\textwidth]{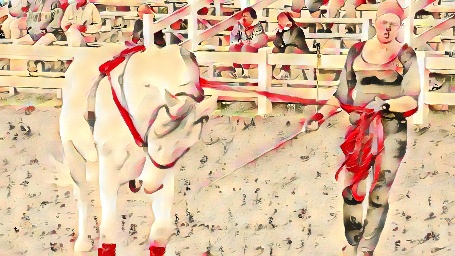}     \includegraphics[height=0.1\textwidth, width=0.16\textwidth]{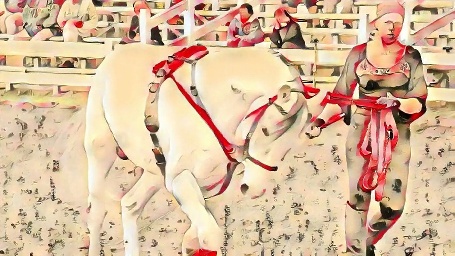}
            \raisebox{0.036\textwidth }{\makebox[0.05\textwidth]{\scriptsize  SANet}} \\          
             \includegraphics[height=0.1\textwidth, width=0.11\textwidth]{Video_horse/ss/34} 
        \includegraphics[height=0.1\textwidth, width=0.16\textwidth]{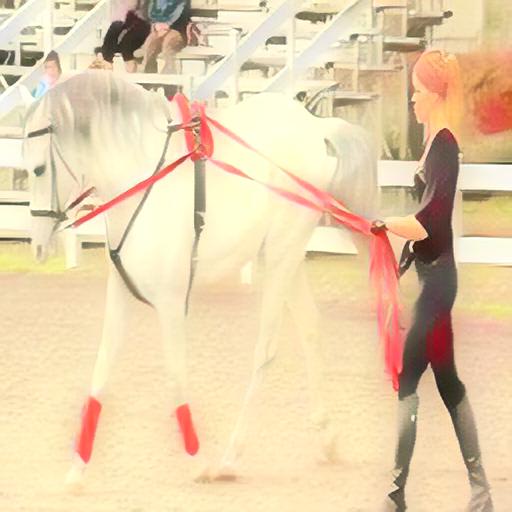}    
        \includegraphics[height=0.1\textwidth, width=0.16\textwidth]{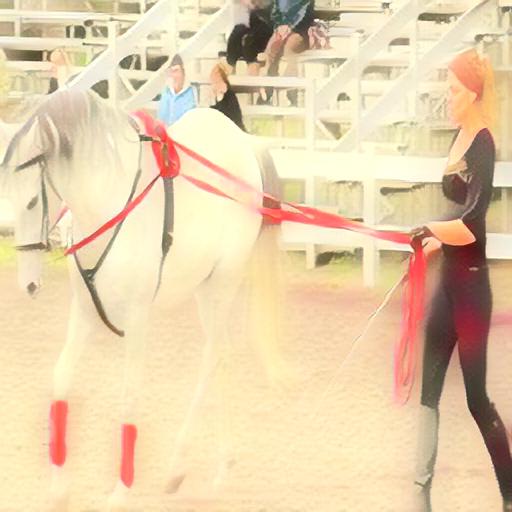}  
            \includegraphics[height=0.1\textwidth, width=0.16\textwidth]{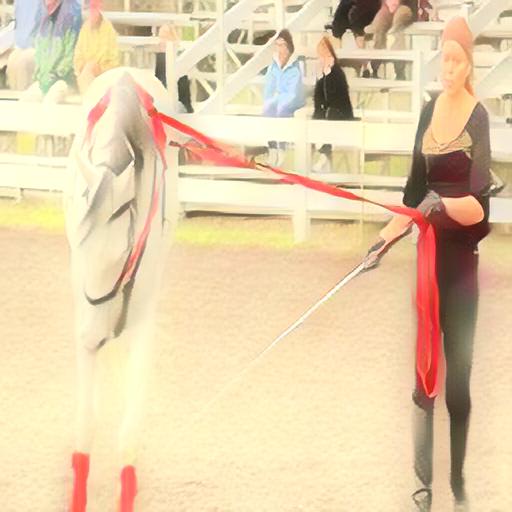} 
            \includegraphics[height=0.1\textwidth, width=0.16\textwidth]{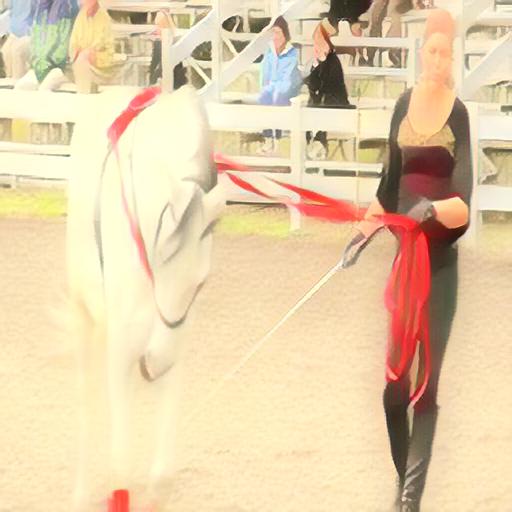}     \includegraphics[height=0.1\textwidth, width=0.16\textwidth]{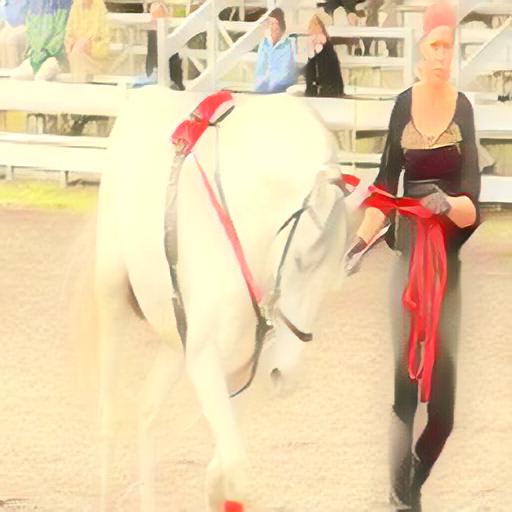}
            \raisebox{0.036\textwidth }{\makebox[0.05\textwidth]{\scriptsize  StrTR$^2$}} \\   
            
   \vspace{-0.2cm}   
\caption{Comparisons on video style transfer. The 1st row lists 5 frames from a video clip as content images. The style images are listed in the first column.} 
\label{fig:videox}
\vspace{-0.7cm} 
\end{figure*}

\section{More Results}
\bfsection{Style Interpolation} To make smooth transitions between different style images, we operate linear interpolation on feature map $Z_{cs}$. The interpolated weights $\tilde{Z}_{cs}$ are obtained as
$\tilde{Z}_{cs}= \alpha Z_{cs}^{s_1} + (1-\alpha) Z_{cs}^{s_2}$
 \noindent where $\alpha$ is the interpolation factor between 0 and 1, $s_1$ and $s_2$ represent two different style images used to produce the feature map. Based on the $\tilde{Z}_{cs}$, we generate the dynamic kernels used to modulate the content image for style transfer. As shown in Figure \ref{fig:inte}, our model captures subtle variations between the two styles.
 
\bfsection{User Study} We perform a user study to further compare our method against other baseline methods in terms of content preference (\ie, the structural similarity, artifacts, and texture details) and style preference (\ie, the style consistency and distortion). 100 content images and 100 style images are selected to synthesize 100 stylizations. For each user, we sample 20 content-style pairs and present the stylized images by all methods in random order. The users are asked to select their favorite one from three aspects: content preservation, stylization degree, and overall preference.  The style preference and content preference in Figure \ref{fig:User_Study}, which demonstrate that our stylized results are more appealing than competitors.

\begin{table}[t!]
\centering
\setlength{\tabcolsep}{2.2pt}{
\caption{ Ablation study on model variants with different settings. SK represents Style Kernels. CGM stands for Content-based Gating Module. GS is the Grouped Shuffling. }\label{fig:score}
\vspace{-0.3cm} 
    \begin{tabular}{l|cccc c c c c c c c}
\toprule
\small Method&\small Base &\small +$\mathcal{L}_{adv}$ &\small + $\mathcal{L}_{REMD}$ &\small + SK   &\small + CGM & \small + GS (Ours) \\
    \midrule
Style Loss$\downarrow$& 2.92&2.14&1.48&1.11&1.08&0.98\\
    \midrule
LPIPS $\downarrow$ & 0.49&0.39&0.34&0.31&0.30&0.30\\
 \bottomrule
	\end{tabular}}
 \vspace{-0.3cm}	
\end{table}

{\noindent\bf Ablation Study.} The main contribution of this paper is the proposed network taking style kernels to generate stylized image maintaining structure similarity to the content image and reflecting the style characteristics of the style reference. To further demonstrate the effectiveness of different components in our network, we start with a simple Base method without involving the CGM, style kernels (SK) and GS (namely, it only use the attention mechanism to warp the style feature), and train this model without $\mathcal{L}_{adv}$  and $\mathcal{L}_{REMD}$. Then we keep adding different modulates into it until it is the same as our full model. We compare different model variants and report the quantitative performances in Table \ref{fig:score}. One can see (1) The loss functions improve the performance of the base model in terms of both style and content preference scores; (2) \textbf{Style Kernels (SK)} works to inject the style characteristics into the output images, as shown in the fifth columns of Table \ref{fig:score}; (3) \textbf{Grouped Shuffling (GS)} improves the style consistency to the style reference at a limited cost of distorting content structure; (4) \textbf{Content-Based Gating Module (CGM)} removes misaligned style patterns and preserve the completeness of semantic regions as shown in Table \ref{fig:score}.

\begin{figure*}[ht!]
\centering
  \includegraphics[width=0.99\textwidth]{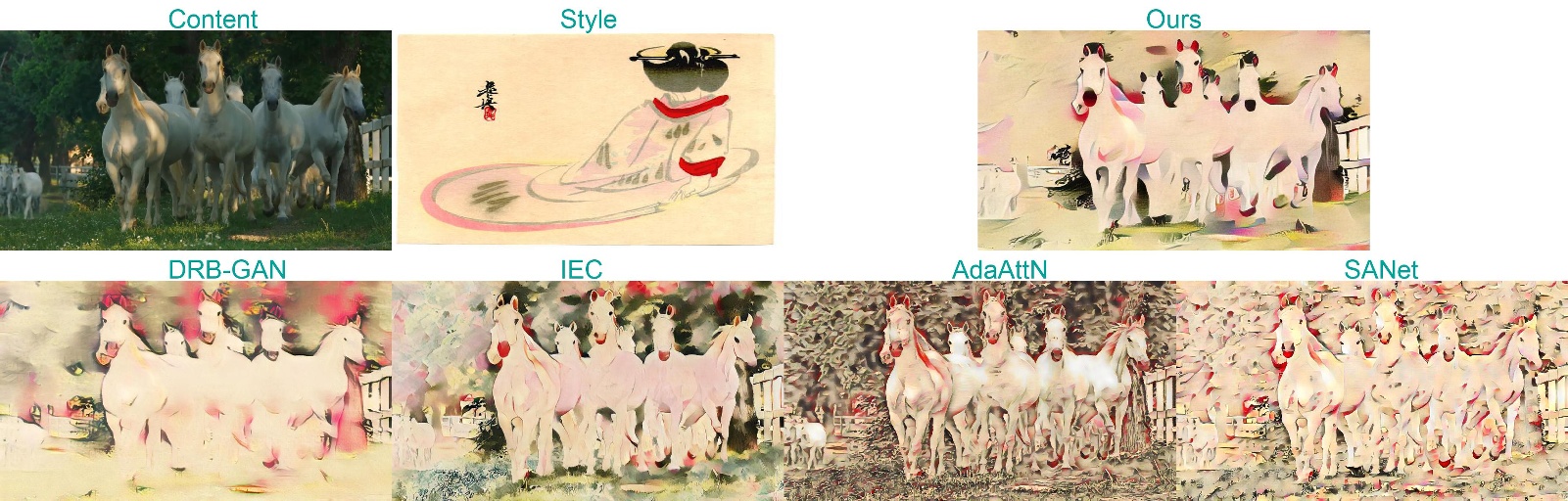}  
  \vspace{-0.2cm}
\caption{Snapshot of video style transfer task. Full video with more detainls are in the supplemental video.}
\label{fig:comp_video}
\vspace{-0.5cm} 
\end{figure*}

\bfsection{Multiple Style Transfer} In Figure \ref{fig:multi}, we show the result of applying our model on multiple style transfer, where the style reference is a concatenation of two different style images. We can see our model can flexibly create a new stylization reflecting style characteristics in different style references.

{\noindent\bf  Visualizations on Focused Region.}
In Figure \ref{fig:focusing_region}, we demonstrate more visualizations on focused regions. Given a query point in the content image (yellow and Cyan rectangles), the corresponding focused region is learned to filter out points with low correlation. The focused region contains a complete semantic area, enabling our method to create style-consistent stylizations with well-preserved content structure. This is because our CGM aims to select focusing regions, where query points can aggregate style information from fairly correlated nodes.

{\noindent\bf Arbitrary Style Transfer.}
In Figure \ref{fig:comp_arbitrary_style_transfer} we demonstrate stylization results of all the combinations between 7 content images and 9 style images. The first row and column are the content images and style images, respectively. Others are the stylization results created by our method. One can see from the stylization results that, in each row, our method can robustly produce stylizations with consistent style characteristics sharing by the style image, and the structure similarity is well maintained as seen in each column.

{\noindent\bf More Qualitative Comparisons with  State-of-the-art methods.} In Figure \ref{fig:comp_sota_style_transfer}, we demonstrate additional qualitative comparisons with state-of-the-art methods. The main advantage as we can observe is that our method is able to maintain the completeness of semantic regions in the stylized images. While other methods manipulate the content structure and introduce artifacts in the generated images.

{\noindent\bf Video Style Transfer.} We compare the performances on image sequences style transfer in Figure \ref{fig:videox}, provide the real-world video style transfer result in Figure \ref{fig:comp_video}, 
which proves the effectiveness of video style transfer and demonstrates that our model outperforms the current SOTA methods in terms of style consistency and structure similarity. 

\bfsection{High Resolution Style Transfer} Our model is also robust to perform style transfer on images of high resolutions. To illustrate the influence of image resolutions, we show the qualitative comparison in Figure \ref{fig:size1}, Figure \ref{fig:size2} and Figure \ref{fig:size3}. It demonstrates that our model creates consistent stylizations on different content images with slight variations. A lot of fine details and brushstrokes are visible.

\begin{figure*}[ht!]
\vspace{-0.5cm}
\centering
\captionsetup[subfigure]{justification=raggedright, singlelinecheck=false, labelformat=empty}   
 \centering
    \includegraphics[width=0.9\textwidth]{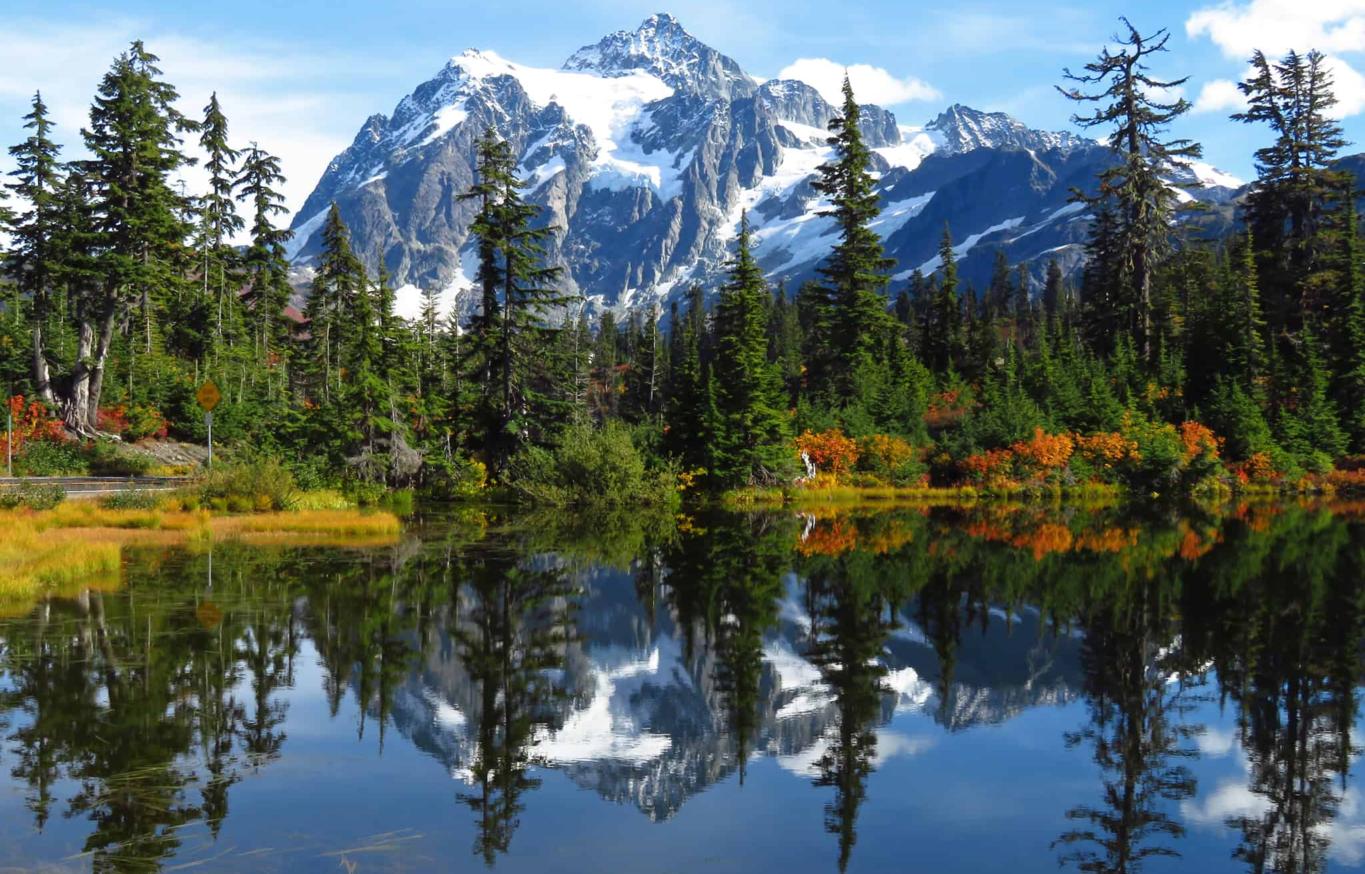}  \\  
\caption{ Content image of high resolution .}\label{fig:size} 
\vspace{-1.0cm}
\end{figure*}

\begin{figure*}[ht!]
\vspace{-0.5cm}
\centering
\captionsetup[subfigure]{justification=raggedright, singlelinecheck=false, labelformat=empty}   
 \centering
    \includegraphics[width=0.9\textwidth]{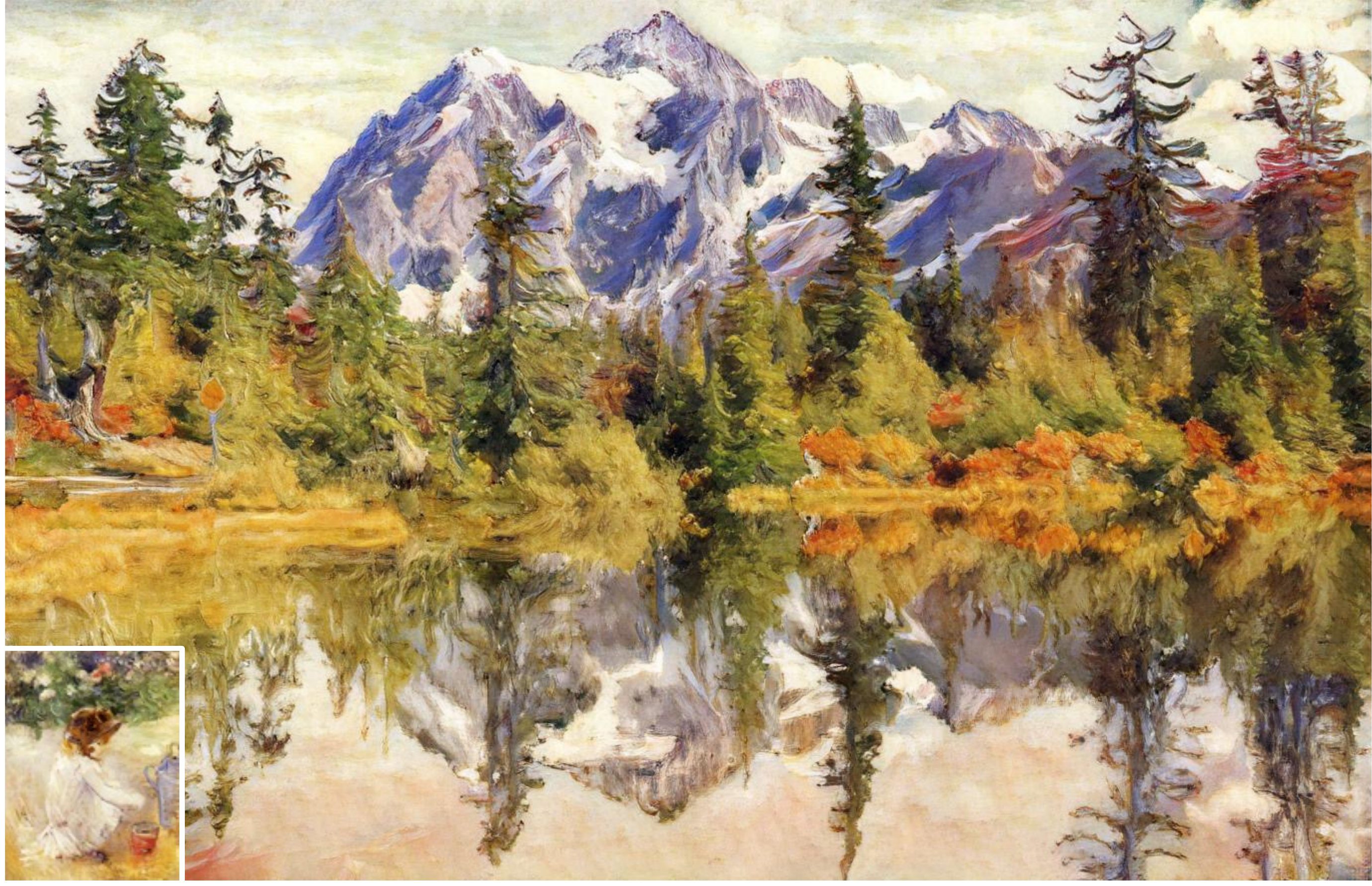}  \\  
\caption{ Stylized image. The style image is at the bottom left.}\label{fig:size1} 
\vspace{-1.0cm}
\end{figure*}

\begin{figure*}[ht!]
\vspace{-0.5cm}
\centering
\captionsetup[subfigure]{justification=raggedright, singlelinecheck=false, labelformat=empty}   
 \centering
    \includegraphics[width=0.9\textwidth]{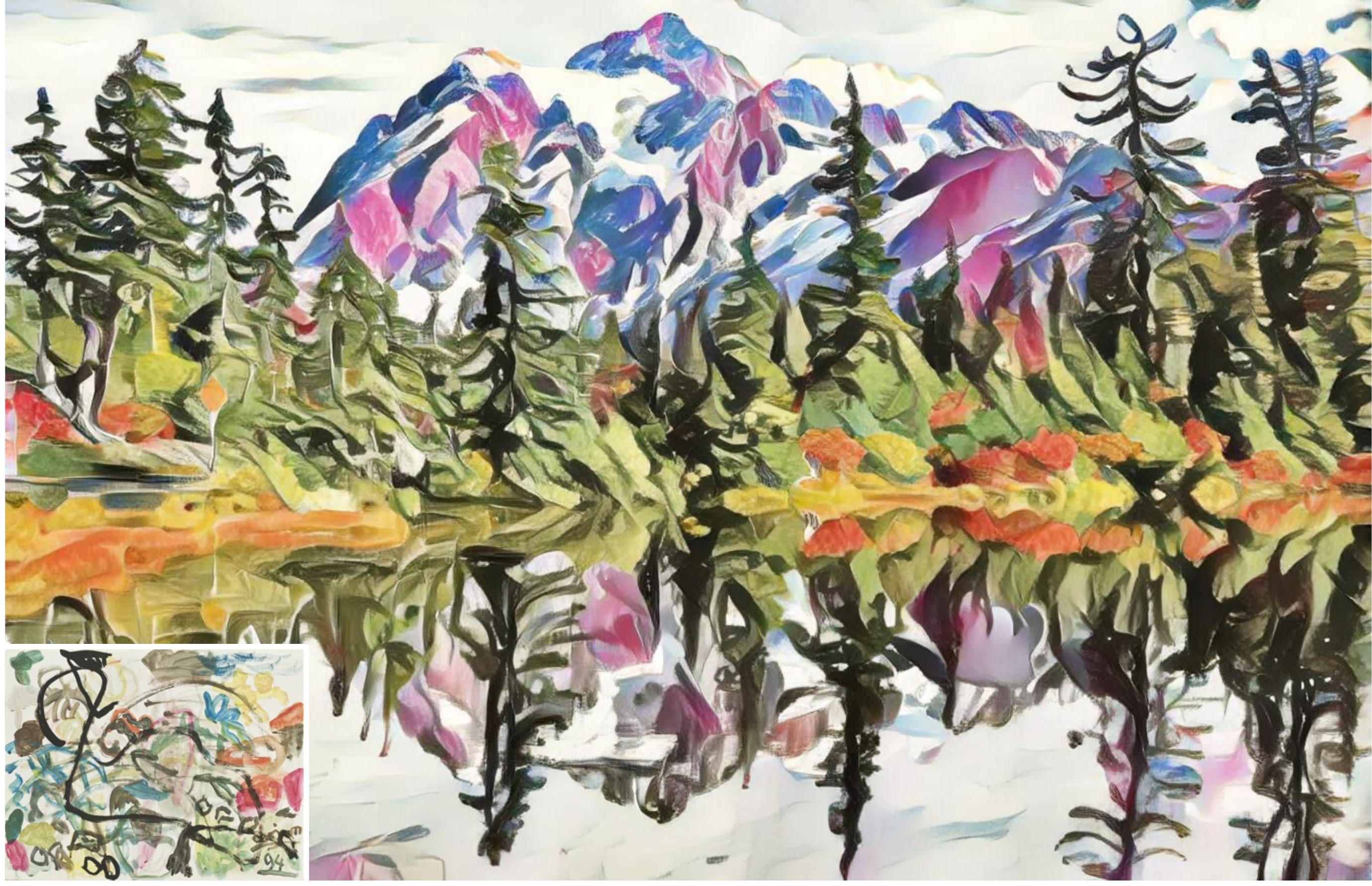}  \\   
\caption{Stylized image. The style image is at the bottom left.}\label{fig:size2} 
\vspace{-1.0cm}
\end{figure*}

\begin{figure*}[ht!]
\vspace{-0.5cm}
\centering
\captionsetup[subfigure]{justification=raggedright, singlelinecheck=false, labelformat=empty}   
 \centering
    \includegraphics[width=0.9\textwidth]{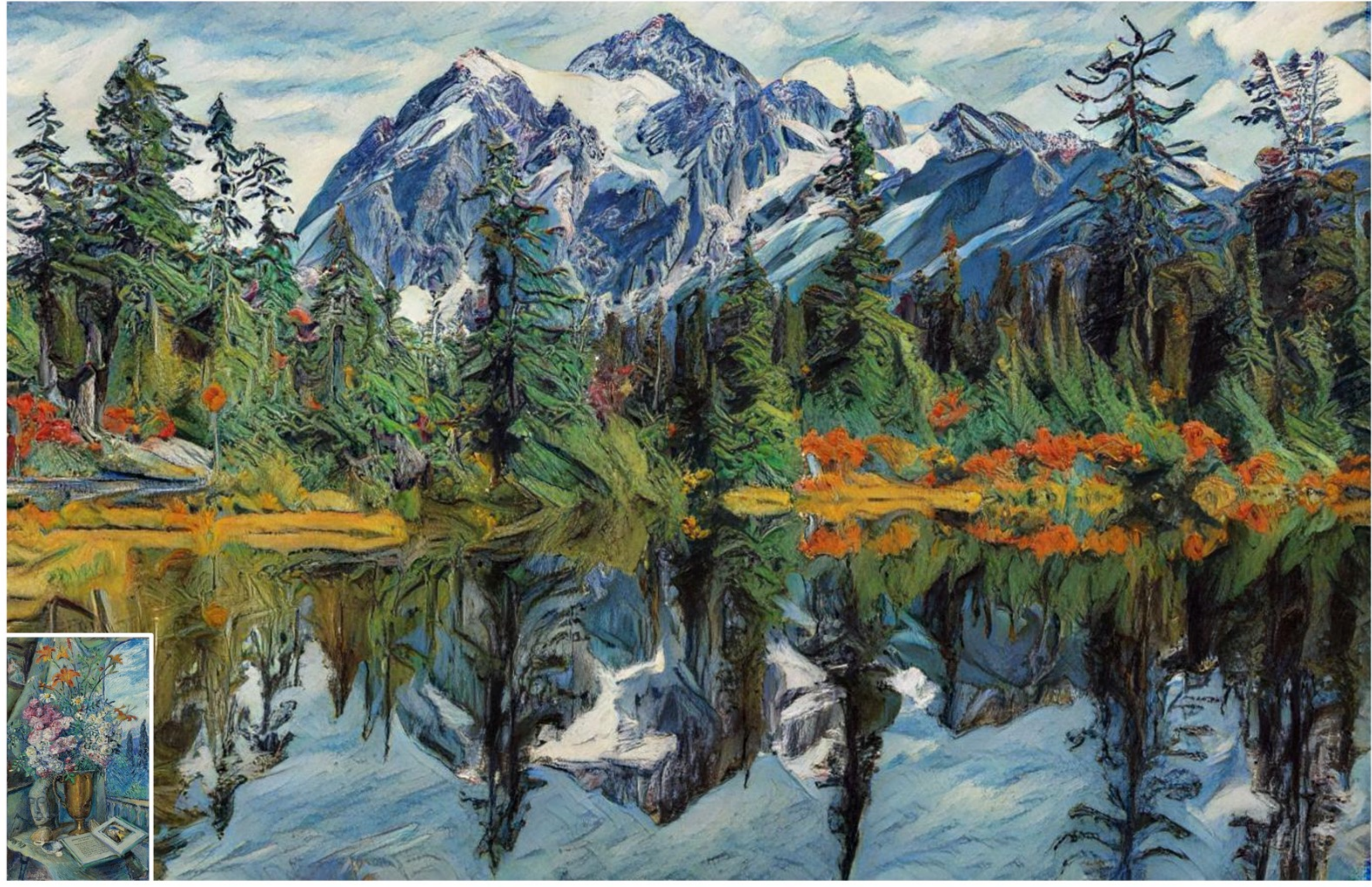}  \\  
\caption{ Stylized image. The style image is at the bottom left.}\label{fig:size3} 
\vspace{-1.0cm}
\end{figure*}

\end{document}